\documentclass{article}

\PassOptionsToPackage{numbers,sort&compress}{natbib}

\usepackage[eandd,final]{neurips_2026}
\makeatletter
\renewcommand{\@noticestring}{%
  Accepted for poster at NeurIPS 2026.%
}
\makeatother

\usepackage[utf8]{inputenc} 
\usepackage[T1]{fontenc}    
\usepackage{hyperref}       
\hypersetup{
	colorlinks=true,       
	linkcolor=blue,        
	citecolor=blue,        
	filecolor=magenta,     
	urlcolor=blue         
}
\usepackage{url}            
\usepackage{booktabs}       
\usepackage{amsfonts}       
\usepackage{nicefrac}       
\usepackage{microtype}      
\usepackage{xcolor}         
\usepackage{xspace}  
\usepackage{graphicx}
\usepackage{amsmath}
\usepackage{amssymb}
\usepackage{mathtools}
\usepackage{amsthm}
\usepackage{listings}
\usepackage{enumitem}
\usepackage{silence}
\usepackage[capitalize,noabbrev]{cleveref}
\usepackage[textsize=tiny]{todonotes}
\usepackage{wrapfig}
\usepackage{multirow}
\usepackage{makecell}
\usepackage{subcaption}
\usepackage{pifont}
\usepackage{colortbl}
\usepackage{xcolor}
\usepackage{tcolorbox}
\tcbuselibrary{skins} 
\usepackage{arydshln}
\newcommand{\cmark}{\textcolor[rgb]{0.13,0.55,0.13}{\ding{51}}}
\newcommand{\xmark}{\textcolor[rgb]{0.7,0.7,0.7}{\ding{55}}}
\newcommand{\pmark}{\textcolor[rgb]{0.85,0.65,0.13}{\kern0.1em$\boldsymbol{\sim}$\kern0.1em}}
\definecolor{oursrow}{rgb}{0.93,0.96,1.0}

\title{ColdDDI: Evaluating Knowledge Utilization in Cold-Start Drug-Drug Interaction Prediction}

\author{%
  Jiheng Liang \\
  Data Science, William \& Mary \\
  \texttt{jliang09@wm.edu} \\
  \And
  Chen Zhao \\
  Computer Science, Baylor University \\
  \texttt{chen\_zhao@baylor.edu} \\
  \And
  Di Wu \\
  Southwest University \\
  \texttt{wudi.cigit@gmail.com} \\
  \And
  Chenyang Bu \\
  Hefei University of Technology \\
  \texttt{chenyangbu@hfut.edu.cn} \\
  \AND
  Yunpeng Hong \\
  Hefei University of Technology \\
  \texttt{yunpenghong@hfut.edu.cn} \\
  \And
  Xingquan Zhu \\
  Florida Atlantic University \\
  \texttt{xzhu3@fau.edu} \\
  \And
  Yi He\thanks{Corresponding author.} \\
  Data Science, William \& Mary \\
  \texttt{yihe@wm.edu} \\
}

\begin{document}
\maketitle
\addtocontents{toc}{\protect\setcounter{tocdepth}{-1}}
\begin{abstract}
Cold-start drug--drug interaction (DDI) prediction tests whether models can identify clinically significant interactions for drugs {\em without} training-time interaction history. 
Existing benchmarks mostly report aggregate edge-prediction scores, leaving a key evaluation question unanswered: when models receive molecular, textual, or knowledge-graph (KG) evidence, {\em do they actually use the evidence that pharmacologically supports the interaction}? 
We introduce \textsc{ColdDDI}, a reconstructible diagnostic benchmark built from DrugBank~5.1.13, with 1{,}900 approved small-molecule drugs and 565{,}731 positive DDI pairs.
\textsc{ColdDDI} evaluates three levels of drug novelty, namely 
test pairs whose two drugs were both seen during training, 
pairs with one unseen drug, and pairs with two unseen drugs.
It also annotates each interaction by whether it changes drug exposure or drug effect, and by whether the biomedical knowledge graph contains shared enzymes, transporters, or targets that can plausibly mediate the interaction. 
These annotations allow \textsc{ColdDDI} to separate two factors that aggregate metrics conflate, namely {\em whether mechanistic evidence is available}, and {\em whether a model prediction depends on that evidence}. 
We evaluate eight conventional DDI methods and 13 LLMs; for open-weight LLMs, we test five prompt patterns and use masking, drug replacement, and channel-sensitivity metrics to probe knowledge utilization. 
\textsc{ColdDDI} exposes that, in the hardest split where both drugs are unseen, the main performance divide is mediator availability.
A fine-tuned 1B LLM recovers 89--93\% of interactions with a shared enzyme, transporter, or target, but only 40--62\% without such a mediator. 
More importantly, KG-provided evidence is not always used; several KG-augmented baselines change little when the shared mediator is masked or disrupted, whereas fine-tuned LLMs respond strongly to this intervention. 
Thus, \textsc{ColdDDI} evaluates knowledge utilization rather than knowledge access alone, showing where cold-start DDI models rely on mechanistic evidence and where they fail despite receiving it. 
We release derivative annotations, splits, prompts, diagnostics, evaluation code, and a DrugBank reconstruction pipeline at \url{https://github.com/0217ljh/ColdDDI-NeurIPS2026}.

\end{abstract}

\section{Introduction}
\label{sec:intro}
A performance number in high-stakes biomedical machine learning 
is useful {\em only} when the evaluation protocol matches the deployment claim it is used to support. 
Drug--drug interaction (DDI) prediction illustrates this problem sharply. 
Polypharmacy makes unintended DDIs a persistent source of adverse events, treatment failures, and preventable hospitalizations~\cite{DeepDDI_2018}, yet the combinatorial space of possible drug pairs makes exhaustive experimental screening infeasible~\cite{drugbank5_dataset, ddi_review_1}. 
Computational models are therefore used to prioritize potentially adverse pairs, but the most clinically relevant question is, 
for totally new or poorly characterized drugs, whether a model can still infer adverse DDIs when interaction-history evidence is {\em unknown} or {\em unavailable}.


Cold-start DDI prediction formalizes this deployment mismatch, where the train--test split must be defined over drugs, not over drug pairs.
A pair-level split can hold out an interaction while exposing both endpoint drugs during training, allowing models to exploit interaction-history features unavailable for new drugs. 
Drug-wise splitting instead separates drugs observed during training from drugs held out for evaluation, producing three settings of increasing difficulty, namely \textbf{S0} (both drugs seen), \textbf{S1} (one drug unseen), and \textbf{S2} (both drugs unseen)~\cite{setting-for-cold-start,GMPNN-CS-2022}.
S1 reflects a common case of pairing a new or poorly characterized drug with an established medication, whereas S2 stress-tests fully inductive prediction when neither endpoint has training-time interaction records. 
In S1 and S2, models must rely on drug-level evidence such as molecular structure, biomedical knowledge graph (KG) associations, and pharmacological text~\cite{cold-start-with_modality}. 
A cold-start benchmark should therefore evaluate whether a model uses the evidence available for unseen-drug reasoning and generalization.

%


Recent DDI methods have expanded the evidence available for such reasoning. 
Molecular structure models capture chemical features through graphs, motifs, or images~\cite{SSI-DDI-2021,DSN-DDI-2023,HDN-DDI-2025,he2025imageddi}; KG-based and KG-integrated models expose biological associations among drugs, enzymes, transporters, targets, and pathways~\cite{SumGNN-2021,MKG-FENN-2024,TIGER-2024,EmerGNN-2023}; and pretrained large language models (LLMs) can incorporate pharmacological text or prompt-supplied biomedical context~\cite{TextDDI-2023, DDI-GPT-2024,DDI-LLM-2025,review-2, RADDI-2026}. 
Benchmarks have also been improved.
GMPNN-CS formalized drug-wise evaluation~\cite{GMPNN-CS-2022} in the S1/S2 split, DDI-Ben studied robustness under distribution shift~\cite{Distribution-shift-2025}, and large-scale resources aggregate multiple DDI datasets and baselines~\cite{openddi-2026}. 
However, these resources mostly support aggregate comparison 
by telling which model scores higher under a chosen split. 
They rarely specify {\em which pharmacological regimes those scores represent}, {\em which evidence sources support the prediction}, or {\em whether a model actually uses the evidence it receives}. 
As \cref{tab:benchmark-comparison} summarizes, no existing benchmark provides formal cold-start splits, mechanism-level stratification, modality-level diagnostics, and systematic LLM comparison all at once.

\begin{table}[!t]
\caption{Comparison of DDI evaluation resources. \cmark\ fully / \pmark\ partial / \xmark\ not supported. \textbf{D1 Setting}: drug-wise S0/S1/S2 splits. \textbf{D2 Mechanism}: pharmacological mechanism and KG-mediated stratification. \textbf{D3 Modality}: input/prompt comparison with per-modality diagnostics. \textbf{D4 Model}: diverse traditional models and LLM families.}
\label{tab:benchmark-comparison}
\centering
\footnotesize
\setlength{\tabcolsep}{3.5pt}
\renewcommand{\arraystretch}{1}
\begin{tabular}{@{}l l l r cccc@{}}
\toprule
\textbf{Benchmark}
  & \textbf{Year}
  & \textbf{Source}
  & \textbf{Scale}
  & \textbf{D1}
  & \textbf{D2}
  & \textbf{D3}
  & \textbf{D4}\\
  &
  &
  & \footnotesize{(drugs\,/\,pairs)}
  & \footnotesize{Setting}
  & \footnotesize{Mechanism}
  & \footnotesize{Modality}
  & \footnotesize{Model}\\
\midrule
TWOSIDES~\cite{TWOSIDES-2012,Decagon-2018}
  & 2012 & FAERS\,/\,SIDER & 645\,/\,63K
  & \xmark & \xmark & \xmark & \xmark \\
DeepDDI~\cite{DeepDDI_2018}
  & 2018 & DrugBank~5.0 & 1,710\,/\,192K
  & \xmark & \xmark & \xmark & \xmark \\
OGBL-DDI~\cite{ogb-2020}
  & 2020 & DrugBank~5.0 & 4,267\,/\,1.3M$^{\dagger}$
  & \xmark & \xmark & \xmark & \xmark \\
TDC-DDI~\cite{tdc-2021}
  & 2021 & DrugBank~5.0 & 1,706\,/\,192K
  & \xmark & \xmark & \xmark & \xmark \\
DDInter~\cite{DDInter-2022}
  & 2022 & FDA\,/\,Lit. & 1,972\,/\,237K
  & \xmark & \pmark & \xmark & \xmark \\
DDI-Ben~\cite{Distribution-shift-2025}
  & 2025 & DrugBank~5.0 & 1,710\,/\,188K
  & \cmark & \xmark & \xmark & \pmark \\
LLM-DDI~\cite{review-2}
    & 2026 & DrugBank\,5.1.12 & 6,089\,/\,1.0M$^{\dagger}$
  & \xmark & \xmark & \pmark & \cmark \\
OpenDDI~\cite{openddi-2026}
  & 2026 & Multi-source & 14,090\,/\,2.5M$^{\dagger}$
  & \pmark & \xmark & \pmark & \xmark \\
\midrule
\rowcolor{oursrow}
\textbf{ColdDDI (ours)}
  & \textbf{2026} & \textbf{DrugBank~5.1.13} & \textbf{1,900\,/\,565K}
  & \cmark & \cmark & \cmark & \cmark \\
\bottomrule
\addlinespace[2pt]
\multicolumn{8}{@{}p{\linewidth}@{}}{\footnotesize
$^{\dagger}$ Scale inflated by loose filtering (OGBL-DDI, LLM-DDI: experimental compounds and un-curated edges) or by LLM-augmented multi-source aggregation with duplicates (OpenDDI). ColdDDI applies a seven-step filter (Appendix~\ref{app-A.1:filtering-pipeline}).}
\end{tabular}
\end{table}


Mechanism is the missing layer between 
cold-start setting and knowledge utilization.
When interaction histories are unavailable, the model must infer risk from pharmacological evidence, but the relevant evidence differs across DDI mechanisms. 
Pharmacokinetic (PK) interactions~\cite{Stockley-book-2018,G-and-G-book-2018} alter absorption, distribution, metabolism, or excretion, and often have explicit mediators such as shared enzymes or transporters.
For example, two drugs may interact because one affects the enzyme that metabolizes the other, such as CYP3A4. 
Pharmacodynamic (PD) interactions~\cite{Stockley-book-2018,G-and-G-book-2018} act through receptors, targets, or downstream pathways, and are typically more heterogeneous.
Even within PK or PD, the knowledge graph (KG) may or may not contain a shared mediating entity, a bridge linking the two drugs.
We therefore distinguish \textbf{Type A} pairs, where the KG contains a shared mediator compatible with the interaction mechanism, from \textbf{Type B} pairs, where no such direct mediator is available. 
Specifically, 
Type A tests whether models can exploit explicit mediating evidence, whereas Type B tests whether they can generalize when direct KG bridges are absent. 
Combining these two axes yields four instance-level regimes, namely
\textbf{PK-A}, \textbf{PK-B}, \textbf{PD-A}, and \textbf{PD-B}. 
This taxonomy turns mechanism into an evaluable object.
Without this stratification, aggregate metrics can make a model look strong because it succeeds on direct-mediator cases while hiding failures on interactions that require pharmacological reasoning and explicit knowledge utilization.

We propose \textsc{ColdDDI}, a mechanism-stratified diagnostic benchmark for cold-start DDI prediction.
\textsc{ColdDDI} is built from DrugBank~5.1.13 through a filtering pipeline, yielding 1{,}900 approved small-molecule drugs and 565{,}731 positive DDI pairs. 
It evaluates methods under drug-wise S0/S1/S2 splits and annotates each positive pair by pharmacological mechanism ({\em i.e.,} PK or PD) and KG-mediated evidence availability ({\em i.e.,} Type~A or Type~B). 
We use this benchmark to evaluate eight conventional DDI methods and 13 LLMs (open-weight LLMs across five prompt patterns), and we add controlled masking and sensitivity analyses to measure how predictions change when drug names or KG entities are removed. 
The benchmark release includes mechanism annotations, split files, prompt templates, evaluation code, and a reconstruction pipeline from a user-provided DrugBank XML file.


Specifically, \textsc{ColdDDI} makes three evaluation contributions:

\begin{enumerate}[leftmargin=*,itemsep=1pt,topsep=2pt,parsep=0pt]

     \item \textbf{Mechanism-readable cold-start evaluation.} 
    \textsc{ColdDDI} couples drug-wise S0/S1/S2 splits with an instance-level PK/PD $\times$ Type A/B taxonomy, so each cold-start error can be interpreted by both pharmacological mechanism and the presence or absence of a direct KG mediator. 
    We find that the most challenging cold-start cases are not simply unseen drugs, but unseen-drug pairs without an explicit mediating entity in the KG. The fine-tuned LLM recalls 89\% and 93\% of PK-A and PD-A positives, but only 40\% and 62\% of PK-B and PD-B positives.

    \item \textbf{Diagnostics for knowledge utilization rather than knowledge access.} 
    We introduce controlled name/entity masking, drug-replacement sensitivity, Knowledge Prediction Sensitivity (KPS), and the channel-interaction asymmetry (KSAI) to measure whether a model prediction changes when a specific evidence channel is perturbed.
    We find that KG access does not imply KG utilization, as only EmerGNN and the fine-tuned LLM show strong bridge sensitivity, while fusion baselines such as TIGER and MKG-FENN largely wash out the mediating-path signal.

    \item \textbf{A controlled audit of LLMs under true drug-wise cold start.} 
    We evaluate 13 LLMs against eight conventional DDI methods under the same cold-start protocol. 
    Fine-tuning makes LLMs competitive on S2, while under direct inference model scale or prompt are less relevant. Even then, LLM success remains concentrated on interactions with explicit KG bridges.

\end{enumerate}

\section{Existing DDI Evaluation Benchmarks}
\label{sec:existing-benchmarks}

\cref{tab:benchmark-comparison} summarizes how existing DDI benchmarks cover the four dimensions (D1--D4) as follows.
\begin{enumerate}[leftmargin=*,itemsep=1pt,topsep=1pt,parsep=0pt]
    \item \textbf{D1 Setting. Which benchmarks support drug-wise cold start?} 
    Widely used platforms (TWOSIDES~\cite{TWOSIDES-2012, Decagon-2018}, DeepDDI~\cite{DeepDDI_2018}, OGBL-DDI~\cite{ogb-2020}, TDC-DDI~\cite{tdc-2021}, DDInter~\cite{DDInter-2022}) use transductive or pair-level splits where test drugs may appear in training.
    Only DDI-Ben~\cite{Distribution-shift-2025} fully adopts the drug-wise S1/S2 protocol~\cite{GMPNN-CS-2022}, and OpenDDI~\cite{openddi-2026} introduces an unseen-drug holdout but does not separate S1 from S2.
    \textsc{ColdDDI} reports S0/S1/S2 with explicit S1 vs.\ S2 separation, so the contribution of one-cold and both-cold regimes is decomposable.

    \item \textbf{D2 Mechanism. Which benchmarks explain PK/PD or evidence availability?} 
    Nearly all existing benchmarks pool PK and PD interactions into a single aggregate score. 
    Although OpenDDI reports per-type accuracy, DDI-Ben stratifies by event frequency, and DDInter~\cite{DDInter-2022} provides structured PK/PD labels, these resources do not jointly stratify positive pairs by PK/PD mechanism and KG mediator availability. 
    \textsc{ColdDDI} stratifies every positive pair along PK/PD~$\times$~Type~A/B, so aggregate scores can be decomposed by pharmacological regime and by whether a usable mediator exists.

    \item \textbf{D3 Modality. Which benchmarks diagnose channel use and function?} 
    Few benchmarks diagnose how models actually use input channels under cold start, leaving the actual source of useful signal hard to attribute.
    The only two benchmarks are OpenDDI, which runs dataset-level modality ablations (removing an entire modality), and LLM-DDI~\cite{review-2}, which runs prompt-perturbation robustness checks (e.g., SMILES-only or genes-only).
    But both operate at the channel level and report aggregate accuracy changes, without isolating which entity within a channel drives the gain.
    \textsc{ColdDDI} combines controlled name/entity masking (R0--R7) with drug-replacement sensitivity (KPS-F), channel-mask sensitivity (KPS-Channel), and channel-interaction asymmetry (KSAI).
    These diagnostics quantify how predictions depend on specific evidence channels within each mechanism subtype (Appendix~\ref{app-E.1:diagnostic-metrics}).
    
    \item \textbf{D4 Model. Has any benchmark evaluated LLMs under cold start?} 
    Existing benchmarks evaluate LLMs at scale or under cold start, but not both.
    LLM-DDI, the broadest LLM evaluation to date (18 LLMs), ignores pharmacological mechanism and is limited to aggregate metrics rather than mechanism-level interpretation. DDI-Ben extends to cold-start S1/S2 but covers only two LLM-based methods (TextDDI~\cite{TextDDI-2023}, DDI-GPT~\cite{DDI-GPT-2024}).
    \textsc{ColdDDI} evaluates 13 LLMs alongside eight conventional methods under the same drug-wise cold-start protocol with PK/PD~$\times$~Type~A/B stratification, exposing whether each method's cold-start gains are mechanism-driven or scalar.
\end{enumerate}

Together, these four dimensions enable \textsc{ColdDDI} to distinguish evidence availability from predictive dependence under cold start, complementing large-scale resources such as OpenDDI~\cite{openddi-2026}.
Task formulations also vary. Most prior work~\cite{SSI-DDI-2021, DSN-DDI-2023, MKG-FENN-2024, Distribution-shift-2025} predicts multi-class event types, while OGBL-DDI~\cite{ogb-2020} and OpenDDI's emerging-drug split~\cite{openddi-2026} use binary detection. 
\textsc{ColdDDI} adopts binary detection to surface S0/S1/S2 differences without long-tail noise (multi-class variant in Appendix~\ref{app-B.4:multiclass}).

\section{The ColdDDI Benchmark}
\label{sec:coldddi-benchmark}

\subsection{Data Source and Task Formulation}
\label{sec:data-source}
We build ColdDDI from DrugBank~5.1.13~\cite{drugbank5_dataset} (released Jan 2025) via a seven-step pipeline (following~\cite{DeepDDI_2018}).
The seven steps are selecting small molecules, keeping RDKit-parseable SMILES, retaining approved drugs, extracting and normalizing DDI mechanisms, dropping types with $<$10 occurrences, removing inorganic and metal-containing molecules~\cite{dataset-construction-for-metal}, and discarding drugs with $<$10 interaction edges, yielding 1{,}900 drugs, 565{,}731 positive pairs, and 215 types (Appendix~\ref{app-A.1:filtering-pipeline}).
DrugBank's license prohibits redistribution, so we release only our derivatives (annotations, splits, prompts, evaluation code) plus a reconstruction pipeline. Academic access to DrugBank is free and granted within 1--3 days\footnote{\url{https://go.drugbank.com/releases/5-1-13}}, after which our pipeline reconstructs the benchmark in a single command on the user-provided DrugBank XML (Appendix~\ref{app-A.6:release}).

We formulate DDI prediction as a binary task~\cite{ogb-2020,openddi-2026}.
Given a drug pair $(d_i,d_j)$, we predict whether an interaction exists. 
We sample unrecorded drug pairs at a 1:1 ratio to recorded positive pairs~\cite{ddi_review_1,GMPNN-CS-2022} and assign them negative labels for binary classification.
These pairs are unlabeled rather than verified non-interactions.
Appendix~\ref{app-B.2:negative-sampling} reports sensitivity to sampling ratios and strategies, together with a temporal backfill analysis of subsequently recorded interactions.

\subsection{Cold-Start Settings}
\label{sec:cold-start-settings}

Following the drug-wise disjoint splitting protocol~\cite{setting-for-cold-start, GMPNN-CS-2022}, we partition all drugs $\mathcal{V}_{\text{all}}$ into seen ($G_1$) and unseen ($G_2$) sets, with $G_1 \cap G_2 = \emptyset$ and $G_1 \cup G_2 = \mathcal{V}_{\text{all}}$.
Let $\mathcal{E}_{\text{all}}$ denote all positive DDI pairs.
The positive training set $\mathcal{E}_{\text{train}} \subseteq \mathcal{E}_{\text{all}}^{(G_1)}$ consists of DDI pairs within $G_1$, where $\mathcal{E}_{\text{all}}^{(G_1)} = \{(u,v) \mid u,v \in G_1\} \cap \mathcal{E}_{\text{all}}$ denotes all positive DDI pairs within $G_1$.
Based on whether each drug in a test pair has been observed during training, we define three evaluation settings of increasing difficulty:
\begin{itemize}[leftmargin=*,itemsep=1pt,topsep=1pt,parsep=0pt]
  \item \textbf{S0 (Transductive).} Both drugs appear in $G_1$. Test pairs are held-out edges from $\mathcal{E}_{\text{all}}^{(G_1)} \setminus \mathcal{E}_{\text{train}}$. This setting serves as a performance upper bound.
  \item \textbf{S1 (Semi-Inductive).} One drug is seen ($\in G_1$) and the other is unseen ($\in G_2$). S1 simulates prescribing a newly approved compound alongside an established medication.
  \item \textbf{S2 (Fully Inductive).} Both endpoint drugs are in $G_2$ with no training-time interaction history. Throughout the paper, S2 is interpreted as an evaluation of interaction-history cold start rather than a literal investigational-drug trial.
\end{itemize}
Under binary classification, we define the negative sampling spaces for each setting as:
\begin{align}
    \overline{\mathcal{E}}^{(G_1)} &= \{(u,v) \mid u,v \in G_1\} 
    \setminus \mathcal{E}_{\text{all}}, \\
    \overline{\mathcal{E}}^{(G_1, G_2)} &= \{(u,v) \mid u \in G_1, v \in G_2\} 
    \setminus \mathcal{E}_{\text{all}}, \\
    \overline{\mathcal{E}}^{(G_2)} &= \{(u,v) \mid u,v \in G_2\} 
    \setminus \mathcal{E}_{\text{all}}.
\end{align}
For each training epoch, a fresh negative set $\mathcal{N}_{\text{train}}^{(t)}$ is sampled from $\overline{\mathcal{E}}^{(G_1)}$ at a 1:1 ratio, whereas validation and test negatives are only sampled once and fixed across evaluations.
All three settings share the same training set, isolating drug novelty as the sole variable.
The formal split composition is summarized in \cref{tab:split_composition}. 
Split statistics and structural-overlap checks are provided in Appendix~\ref{app-B.1:split-stats}.
%
\begin{table}[t]
\caption{Split composition for the three evaluation settings. $G_1$: seen drugs; $G_2$: unseen drugs; $G_1 \cap G_2 = \emptyset$. Negatives are sampled from the corresponding $\overline{\mathcal{E}}$ spaces.}
\label{tab:split_composition}
\centering
\footnotesize
\setlength{\tabcolsep}{4pt}
\renewcommand{\arraystretch}{1}
\begin{tabular}{@{}l ccc@{}}
\toprule
  & \textbf{S0 (Transductive)} & \textbf{S1 (Semi-Inductive)} & \textbf{S2 (Fully Inductive)} \\
\midrule
Train (+) & \multicolumn{3}{c}{$\mathcal{E}_{\text{train}} \subseteq \mathcal{E}_{\text{all}}^{(G_1)}$} \\
Train ($-$) & \multicolumn{3}{c}{$\mathcal{N}_{\text{train}}^{(t)} \sim \overline{\mathcal{E}}^{(G_1)}$, resampled each epoch} \\
\midrule
Val/Test (+) & $\mathcal{E}_{\text{all}}^{(G_1)} \setminus \mathcal{E}_{\text{train}}$ & $\{(u,v) \mid u \in G_1, v \in G_2\} \cap \mathcal{E}_{\text{all}}$ & $\mathcal{E}_{\text{all}}^{(G_2)}$ \\
Val/Test ($-$) & fixed from $\overline{\mathcal{E}}^{(G_1)}$ & fixed from $\overline{\mathcal{E}}^{(G_1, G_2)}$ & fixed from $\overline{\mathcal{E}}^{(G_2)}$ \\
\bottomrule
\end{tabular}
\end{table}

\subsection{DDI Mechanism Taxonomy}
\label{sec:taxonomy}
DDIs fall into two mechanistically distinct categories~\cite{Stockley-book-2018, G-and-G-book-2018}.
\emph{Pharmacokinetic} (PK) interactions alter absorption, distribution, metabolism, or excretion (changing drug concentration).
\emph{Pharmacodynamic} (PD) interactions modify the target/pathway effect (synergistic or additive).
We assign each of 215 DDI types a PK or PD label via keyword matching on its DrugBank description template.
When both keyword sets match, we assign the type to PK (Appendix~\ref{app-A.2:pkpd-keywords}).
Within each class, we distinguish \textbf{Type~A} (DrugBank KG contains shared mediating entities with compatible action roles) from \textbf{Type~B} (no such mediating chain in the KG, even though one may exist biologically), yielding four subtypes: \textbf{PK-A} (29.6\%), \textbf{PK-B} (22.6\%), \textbf{PD-A} (3.8\%), \textbf{PD-B} (43.9\%).
The automatic stratification for both axes is detailed in Appendix~\ref{app-A.3:ab-subdivision}.

\textbf{Taxonomy Validation.} 
To assess the reliability of the automated taxonomy, we randomly sampled
50 DDI types (covering $\mathbf{60.2\%}$ of all DDI edges) with 10 drug pairs per type.
Two graduate-level pharmacology annotators independently labeled PK/PD and A/B against information from DrugBank.
PK/PD agreement reaches $\kappa = \mathbf{0.842}$, with per-pair F1 of $\mathbf{0.88}$ for PK and $\mathbf{0.90}$ for PD against the keyword matching of Appendix~\ref{app-A.2:pkpd-keywords}.
For A/B subclassification, annotators agree with the automated labels on $\mathbf{93.8\%}$ of pairs (see Appendix~\ref{app-A.4:taxonomy-validation}).

\subsection{LLM Inference Patterns and Release}
\label{sec:llm-patterns}
We design five inference patterns varying along two dimensions, information type and prompt format.
The patterns are \textbf{P1} Zero-Shot (drug names only), \textbf{P2} Few-Shot SMILES (based on Tanimoto-similar drugs), \textbf{P3} One-Hop KG triplets ($(h,r,t)$ triplets, $\leq$3 entities/relation), \textbf{P4} One-Hop KG sequence (same KG content rendered as sentences), and \textbf{P5} Few-Shot 2-hop (based on two-hop shared entities).
Full templates are listed in Appendix~\ref{app-D.1:prompts}.

\textbf{Memorization Controls.}
Four controls assess drug-name dependence and potential pretraining memorization.
(1)~Name masking with \texttt{[DRUG\_A]}/\texttt{[DRUG\_B]} (\cref{sec:modality-results}).
(2)~No DDI description or interaction hint in any P1--P5 template (Appendix~\ref{app-D.1:prompts}).
(3)~Temporal separation, with approved-year stratification and four temporal splits showing no Recall advantage for older drugs (Appendix~\ref{app-D.7:temporal-control}).
(4)~Synonym robustness with canonical-vs-synonym substitutions on S2 showing modest indicator change (Appendix~\ref{app-D.6:synonym}).
Additionally, open-weight LLMs without fine-tuning produce near-random S2 predictions (\cref{sec:model-diversity}), indicating that pretrained DDI knowledge is not reliably expressed in direct inference.

\textbf{Release.}
ColdDDI ships as a prediction and evaluation platform with automated annotation tools.
All released artifacts (mechanism annotations, drug-wise splits, prompt templates, evaluation code, and diagnostic scripts) are available through our code repository for reproduction. 
The full repository layout, reproduction, hardware requirements, and licensing terms are in Appendix~\ref{app-A.6:release}.

\section{Experiments}
\label{sec:experimental-setup}

\subsection{Models}
\label{sec:baselines}
\textbf{Conventional Baselines.} We select eight approaches spanning the three input modalities (molecular structure, KG, and text) for DDI prediction.
\textbf{Matrix-Based:} DeepDDI~\cite{DeepDDI_2018} encodes each drug as a PCA-reduced Tanimoto similarity profile against a reference set.
\textbf{Molecular Graph-Based:} SSI-DDI~\cite{SSI-DDI-2021}, DSN-DDI~\cite{DSN-DDI-2023}, and HDN-DDI~\cite{HDN-DDI-2025} encode each drug from its SMILES-derived molecular graph.
\textbf{KG-Only:} EmerGNN~\cite{EmerGNN-2023} performs flow-based bidirectional message passing on the DrugBank biomedical KG.
\textbf{Hybrid:} TIGER~\cite{TIGER-2024} uses multi-relation KG embeddings and MKG-FENN~\cite{MKG-FENN-2024} fuses multi-modal KG features with chemical fingerprints.
\textbf{Pretrained LM:} TextDDI~\cite{TextDDI-2023} encodes textual drug descriptions from DrugBank and PubChem.
\textbf{LLMs.} We evaluate 11 open-source LLMs from three families spanning 0.5B--14B (Llama-3.2 1B/3B~\cite{LLAMA3-2024}, Llama-2 7B/13B~\cite{LLAMA2-2023}, Qwen~2.5 0.5B/3B/7B/14B~\cite{QWEN2_5-2024}, Gemma~3 1B/4B/12B~\cite{GEMMA3-2025}), plus two proprietary LLMs (GPT-4o~\cite{GPT4-2023}, Claude Sonnet~4.6~\cite{CLAUDE-2024}) for ceiling reference.
Each open-source LLM is run under direct inference and LoRA fine-tuning~\cite{lora-2022} across all five P1--P5 patterns.
Two proprietary LLMs are run under direct inference only.
For output scoring we use the softmax of the \texttt{Yes}/\texttt{No} token logits for open-source models and elicit an explicit $[0,1]$ confidence for proprietary ones.
The hyperparameters of conventional baseline and LoRA configurations are shown in Appendix~\ref{app-C.1:baselines} and Appendix~\ref{app-D.2:lora}.

\subsection{Evaluation Protocols and Modality Analysis}
\label{sec:modality-analysis}

\textbf{Evaluation Protocol.}
D1 evaluates on S0/S1/S2 cold-start splits.
We report mean AUC-ROC and Recall over three random seeds (full results in Appendices~\ref{app-C.3:baseline-full-metrics} and~\ref{app-D.4:full-results}).
The 1{,}900-drug set is the primary evaluation set for the eight baselines and the best LLM by combined AUC-ROC and Recall.
Other LLM and modality analyses use a stratified 800-drug subset, justified by modest prediction gap relative to the full dataset (Appendix~\ref{app-B.3:subset-stability}).
D2, D3, and D4 inherit D1's predictions and apply post-hoc stratifications, sharing identical splits and per-epoch negatives across all methods.

\textbf{Modality Decomposition.}
A $2{\times}2$ factorial on S2 crosses drug name (visible vs.\ \texttt{[DRUG\_A/B]}) with shared mediating entities (preserved vs.\ \texttt{[ENTITY]}), yielding R0/R1/R2/R3 (formal definition in \cref{tab:masking-conditions}) whose pairwise comparisons isolate each channel's controlled contribution.
To further quantify per-modality contribution for both conventional baselines and LLMs, we additionally adopt a drug-swap protocol: given $(u,v)$ with label $y$, identify $u'\in G_2$ with $y(u',v)\neq y(u,v)$.
\textbf{KPS-F}\,$=\mathbb{E}|f(u,v)-f(u',v)|$ measures drug-replacement sensitivity; channel-specific \textbf{KPS-Name} / \textbf{KPS-KG} / \textbf{KPS-mol} use symmetric channel masking on both drugs; \textbf{KSAI}\,$=\text{KPS-KG}_{\text{name-masked}}-\text{KPS-KG}_{\text{name-present}}$ measures how much the drug name compensates for KG perturbation (LLMs only). For any per-bucket metric $M$ (Recall, KPS-F, etc.), we report an \textbf{A--B gap}\,$=\tfrac{1}{2}(M_{\text{PK-A}}+M_{\text{PD-A}})-\tfrac{1}{2}(M_{\text{PK-B}}+M_{\text{PD-B}})$, where positive values indicate stronger Type A sensitivity.

\section{Results and Analysis}
\label{sec:results}

\subsection{Cold-Start Degradation Is Real and Model-Independent}
\label{sec:cold-start-results}

\cref{tab:overall-performance} reports S0/S1/S2 AUC-ROC and Recall. Full indicators are in Appendices~\ref{app-C.3:baseline-full-metrics} and~\ref{app-D.4:full-results}.

\begin{table}[t]
\centering
\caption{AUC-ROC and Recall on the 1{,}900-drug set (3-seed mean$\pm$std). Best in \textbf{bold}.}
\label{tab:overall-performance}
\setlength{\tabcolsep}{3pt}
\renewcommand{\arraystretch}{1}
\resizebox{\textwidth}{!}{
\footnotesize
\begin{tabular}{@{}ll ccc ccc@{}}
\toprule
& & \multicolumn{3}{c}{\textbf{AUC-ROC}} & \multicolumn{3}{c}{\textbf{Recall}} \\
\cmidrule(lr){3-5}\cmidrule(lr){6-8}
\textbf{Group} & \textbf{Method} & \textbf{S0} & \textbf{S1} & \textbf{S2} & \textbf{S0} & \textbf{S1} & \textbf{S2} \\
\midrule
Matrix
& DeepDDI~\cite{DeepDDI_2018}
& $0.997_{\pm0.000}$ & $0.819_{\pm0.003}$ & $0.672_{\pm0.009}$
& $0.964_{\pm0.001}$ & $0.682_{\pm0.012}$ & $0.509_{\pm0.030}$ \\
\hdashline
\multirow{3}{*}{Mol-Graph}
& SSI-DDI~\cite{SSI-DDI-2021}
& $0.829_{\pm0.010}$ & $0.706_{\pm0.006}$ & $0.627_{\pm0.003}$
& $0.771_{\pm0.007}$ & $0.628_{\pm0.015}$ & $0.580_{\pm0.040}$ \\
& DSN-DDI~\cite{DSN-DDI-2023}$^{\dagger}$
& $0.950_{\pm0.038}$ & $\mathbf{0.827}_{\pm0.056}$ & $0.711_{\pm0.063}$
& $0.936_{\pm0.028}$ & $\mathbf{0.731}_{\pm0.038}$ & $0.696_{\pm0.095}$ \\
& HDN-DDI~\cite{HDN-DDI-2025}
& $0.853_{\pm0.006}$ & $0.736_{\pm0.003}$ & $0.645_{\pm0.002}$
& $0.789_{\pm0.026}$ & $0.659_{\pm0.004}$ & $0.589_{\pm0.036}$ \\
\hdashline
KG-only
& EmerGNN~\cite{EmerGNN-2023}
& $0.980_{\pm0.001}$ & $0.799_{\pm0.000}$ & $0.707_{\pm0.006}$
& $0.939_{\pm0.009}$ & $0.712_{\pm0.003}$ & $0.619_{\pm0.014}$ \\
\hdashline
\multirow{2}{*}{KG-integrated}
& TIGER~\cite{TIGER-2024}
& $0.976_{\pm0.002}$ & $0.771_{\pm0.027}$ & $0.630_{\pm0.015}$
& $0.918_{\pm0.008}$ & $0.595_{\pm0.065}$ & $0.426_{\pm0.101}$ \\
& MKG-FENN~\cite{MKG-FENN-2024}
& $\mathbf{0.999}_{\pm0.000}$ & $0.703_{\pm0.005}$ & $0.563_{\pm0.007}$
& $\mathbf{0.987}_{\pm0.002}$ & $0.566_{\pm0.051}$ & $\mathbf{1.000}^{\ddagger}_{\pm0.001}$ \\
\hdashline
Pretrained LM
& TextDDI~\cite{TextDDI-2023}
& $0.996_{\pm0.002}$ & $0.786_{\pm0.005}$ & $0.658_{\pm0.004}$
& $0.971_{\pm0.007}$ & $0.680_{\pm0.023}$ & $0.497_{\pm0.023}$ \\
\hdashline
\rowcolor{gray!20}
LLM (FT)
& Llama-3.2-1B (P4)
& $0.916_{\pm0.026}$ & $0.806_{\pm0.038}$ & $\mathbf{0.764}_{\pm0.006}$
& $0.839_{\pm0.021}$ & $0.710_{\pm0.034}$ & $0.652_{\pm0.004}$ \\
\bottomrule
\end{tabular}
}
\par\vspace{2pt}
{\footnotesize $^{\dagger}$~DSN-DDI exhibits large cross-seed variance, indicating limited robustness.}\\
{\footnotesize $^{\ddagger}$~MKG-FENN predicts $99.96\%$ of S2 pairs as positive, with precision $0.500$.}
\end{table}

\textbf{Finding \#1: Cold-start degradation cuts across architectures.} 
Every method, from matrix-based DeepDDI to pretrained TextDDI, and fine-tuned Llama-3.2-1B, declines by 15--44 AUC points from S0 to S2 (\cref{tab:overall-performance}).
The drop is therefore not a property of any single design choice but a property of evaluating against unseen drugs.
In the hardest S2 split, fine-tuned Llama-3.2-1B (P4) and EmerGNN achieve the best and second-best results (AUC $0.764$ / Recall $0.652$ vs.\ $0.707$ / $0.619$), excluding the high-variance DSN-DDI.
DSN-DDI's S2 mean appears competitive but its large cross-seed variance ($\pm0.063$ AUC, $\pm0.095$ Recall, around $10\%$ relative change) indicates limited robustness.

These aggregate scores do not explain why models fail.
The S2 column collapses heterogeneous failure modes into a single number, hiding which pharmacological regime each method fails on.

\subsection{KG Mediator Availability, Not PK/PD, Drives Cold-Start Failure}
\label{sec:stratified-analysis}

To localize where aggregate S2 breaks down, we stratify each test pair by pharmacological mechanism (PK vs.\ PD) and by KG mediator availability (Type~A vs.\ Type~B, \cref{sec:taxonomy}).
%
\cref{tab:stratified-recall} reports per-subtype S2 Recall and AUC-ROC on the 800-drug subset.
Within each seed, AUC-ROC compares each positive subtype against the same complete S2 sampled-unrecorded pool.
Per-method breakdowns by Tanimoto-similarity tier are in Appendix~\ref{app-C.4:mech-sim-baselines}.
Additional stratified metrics and their sensitivity to the sampled-unrecorded comparison pool are reported in Appendix~\ref{app-C.5:negative-pool-sensitivity}.

\begin{table}[!t]
\caption{S2 Recall and AUC-ROC by mechanism subtype on the
800-drug subset (3-seed mean$\pm$std). Best in \textbf{bold}.}
\label{tab:stratified-recall}
\centering
\footnotesize
\setlength{\tabcolsep}{3pt}
\renewcommand{\arraystretch}{1.05}
\resizebox{\textwidth}{!}{%
\begin{tabular}{@{}l cc cc cc cc cc@{}}
\toprule
\multirow{2}{*}{\textbf{Method}}
& \multicolumn{2}{c}{\textbf{PK-A}}
& \multicolumn{2}{c}{\textbf{PK-B}}
& \multicolumn{2}{c}{\textbf{PD-A}}
& \multicolumn{2}{c}{\textbf{PD-B}}
& \multicolumn{2}{c}{\textbf{A--B gap}} \\
\cmidrule(lr){2-3}
\cmidrule(lr){4-5}
\cmidrule(lr){6-7}
\cmidrule(lr){8-9}
\cmidrule(lr){10-11}
& Recall & AUC-ROC
& Recall & AUC-ROC
& Recall & AUC-ROC
& Recall & AUC-ROC
& Recall & AUC-ROC \\
\midrule
DeepDDI
& $0.515_{\pm0.038}$ & $0.678_{\pm0.034}$
& $0.370_{\pm0.012}$ & $0.578_{\pm0.055}$
& $0.614_{\pm0.021}$ & $0.739_{\pm0.024}$
& $0.508_{\pm0.092}$ & $0.675_{\pm0.036}$
& $+0.125_{\pm0.053}$ & $+0.082_{\pm0.024}$ \\

SSI-DDI
& $0.558_{\pm0.201}$ & $0.642_{\pm0.081}$
& $0.349_{\pm0.129}$ & $0.501_{\pm0.052}$
& $0.692_{\pm0.069}$ & $0.742_{\pm0.073}$
& $0.538_{\pm0.047}$ & $0.634_{\pm0.038}$
& $+0.182_{\pm0.031}$ & $+0.125_{\pm0.033}$ \\

DSN-DDI
& $0.533_{\pm0.014}$ & $0.767_{\pm0.054}$
& $0.465_{\pm0.041}$ & $\boldsymbol{0.719_{\pm0.073}}$
& $0.608_{\pm0.111}$ & $0.789_{\pm0.039}$
& $0.541_{\pm0.022}$ & $\boldsymbol{0.768_{\pm0.042}}$
& $+0.067_{\pm0.063}$ & $+0.034_{\pm0.030}$ \\

HDN-DDI
& $0.734_{\pm0.052}$ & $0.698_{\pm0.049}$
& $0.440_{\pm0.147}$ & $0.507_{\pm0.055}$
& $0.762_{\pm0.042}$ & $0.741_{\pm0.035}$
& $0.626_{\pm0.087}$ & $0.630_{\pm0.022}$
& $+0.215_{\pm0.077}$ & $+0.151_{\pm0.012}$ \\

EmerGNN
& $0.777_{\pm0.037}$ & $0.851_{\pm0.022}$
& $0.308_{\pm0.039}$ & $0.565_{\pm0.047}$
& $0.743_{\pm0.086}$ & $0.828_{\pm0.053}$
& $0.436_{\pm0.109}$ & $0.660_{\pm0.046}$
& $+0.388_{\pm0.078}$ & $+0.227_{\pm0.031}$ \\

TIGER
& $0.457_{\pm0.054}$ & $0.546_{\pm0.010}$
& $0.443_{\pm0.025}$ & $0.541_{\pm0.032}$
& $0.617_{\pm0.067}$ & $0.668_{\pm0.062}$
& $0.489_{\pm0.067}$ & $0.584_{\pm0.018}$
& $+0.071_{\pm0.052}$ & $+0.044_{\pm0.019}$ \\

MKG-FENN
& $0.989^{\dagger}_{\pm0.011}$ & $0.572_{\pm0.014}$
& $0.989^{\dagger}_{\pm0.012}$ & $0.543_{\pm0.042}$
& $1.000^{\dagger}_{\pm0.000}$ & $0.626_{\pm0.060}$
& $0.989^{\dagger}_{\pm0.012}$ & $0.566_{\pm0.043}$
& $+0.006^{\dagger}_{\pm0.006}$ & $+0.045_{\pm0.024}$ \\

TextDDI
& $0.573_{\pm0.117}$ & $0.660_{\pm0.060}$
& $\boldsymbol{0.474_{\pm0.061}}$ & $0.587_{\pm0.033}$
& $0.810_{\pm0.039}$ & $0.831_{\pm0.023}$
& $\boldsymbol{0.651_{\pm0.012}}$ & $0.724_{\pm0.017}$
& $+0.129_{\pm0.048}$ & $+0.090_{\pm0.016}$ \\
\midrule
\rowcolor{gray!20}
LLM-FT
& $\boldsymbol{0.890_{\pm0.089}}$
& $\boldsymbol{0.881_{\pm0.068}}$
& $0.397_{\pm0.113}$ & $0.598_{\pm0.083}$
& $\boldsymbol{0.926_{\pm0.043}}$
& $\boldsymbol{0.928_{\pm0.013}}$
& $0.622_{\pm0.129}$ & $0.752_{\pm0.050}$
& $\boldsymbol{+0.398_{\pm0.105}}$
& $\boldsymbol{+0.229_{\pm0.051}}$ \\
\bottomrule
\end{tabular}%
}
\par\vspace{2pt}
{\footnotesize
$^{\dagger}$~MKG-FENN's uniformly high Recall and near-zero Recall gap are saturation artifacts (see Table~\ref{tab:overall-performance}). On the 800-drug subset, it predicts $98.29\%$ of S2 pairs as positive, with precision $0.503$. These Recall entries are excluded from bolding.\par}
\end{table}

\begin{table}[!t]
\centering
\caption{Masking effects and cross-channel asymmetry on LLM-FT ($\times 10^{2}$, 800-drug subset, 3-seed mean$\pm$std). Columns 2--4 are $\Delta$AUC-ROC vs.\ R0 per mask condition. KSAI is the per-pair averaged $|P_{R_1}{-}P_{R_3}|{-}|P_{R_0}{-}P_{R_2}|$ defined in Appendix~\ref{app-E.1:diagnostic-metrics}. Bold marks the dominant channel per row.}
\label{tab:masking-effects}
\footnotesize
\setlength{\tabcolsep}{4pt}
\renewcommand{\arraystretch}{1}
\begin{tabular}{@{}lccccc@{}}
\toprule
\textbf{Group} & \textbf{Entity mask} & \textbf{Name mask}
& \textbf{Both mask} & \textbf{KSAI} & \textbf{Dominant} \\
 & (R0$\to$R2) & (R0$\to$R1) & (R0$\to$R3) &  & \textbf{channel} \\
\midrule
PK-A & $\boldsymbol{-8.0_{\pm2.7}}$ & $+1.5_{\pm1.5}$ & $-7.0_{\pm1.9}$ & $+1.2_{\pm1.0}$ & Entity \\
PK-B & $+0.4_{\pm1.8}$ & $\boldsymbol{-4.4_{\pm0.4}}$ & $-4.9_{\pm2.4}$ & $+3.3_{\pm0.5}$ & Name \\
PD-A & $\boldsymbol{-2.9_{\pm2.5}}$ & $-0.6_{\pm1.0}$ & $-5.5_{\pm3.9}$ & $+3.8_{\pm0.8}$ & Entity \\
PD-B & $-1.6_{\pm0.4}$ & $\boldsymbol{-3.4_{\pm1.2}}$ & $-7.5_{\pm2.6}$ & $+4.6_{\pm0.4}$ & Name \\
\midrule
ALL  & $-3.6_{\pm1.0}$ & $-1.7_{\pm0.8}$ & $-6.8_{\pm0.4}$ & $+3.1_{\pm0.4}$ &  \\
\bottomrule
\end{tabular}
\end{table}

\textbf{Finding \#2: Difficulty separates along Type A vs.\ Type B, not along PK vs.\ PD.}
On the strongest methods, Type A regimes are near-solved while Type B regimes remain hard.
LLM-FT recalls $89$--$93\%$ of PK-A and PD-A pairs but only $40$--$62\%$ of PK-B and PD-B pairs, an A--B gap of $+0.398$, mirrored by EmerGNN ($+0.388$).
The paired $95\%$ confidence intervals lie entirely above zero for five of the nine methods, including LLM-FT and EmerGNN (Appendix~\ref{app-C.6:recall-gap-ci}).
The Type~A advantage also persists under PharmGKB~\cite{PharmGKB_2021}, ChEMBL~\cite{CHeMBL_2024}, and KEGG~\cite{kegg_2025} enrichment, as shown in Appendix~\ref{app-C.7:external-kg}.
Higher training frequencies of shared mediators or DDI types do not consistently yield higher Recall (Appendix~\ref{app-C.8:frequency-analysis}).
By comparison, PK/PD differences are several-fold smaller.
The dividing axis between easy and hard cold-start cases is therefore whether the KG annotates shared mediating entities, not which pharmacological class the interaction belongs to.

The magnitude of the A--B gap varies sharply across methods.
Mol-graph baselines show moderate gaps ($+0.067$ to $+0.215$); TIGER and MKG-FENN stay nearly flat ($+0.071$ and $+0.006^{\dagger}$ respectively) despite identical KG access.
EmerGNN and LLM-FT show larger and comparable gaps. 
The former's gap aligns with its bidirectional flow-propagation design, which expands the set of shared entities that act as mediating bridges by routing them in both directions.
Baselines without such routing (TIGER, MKG-FENN) cannot trace mediating paths even with the same KG access.
Two methods with the same KG inputs can therefore differ substantially in A--B sensitivity, suggesting that having mediating evidence and using it are different things.

\subsection{KG Access Does Not Imply Knowledge Utilization}
\label{sec:modality-results}

In \textsc{ColdDDI}, \emph{knowledge utilization} denotes measurable predictive dependence on a provided information channel under controlled perturbation.
To test whether KG access translates into KG utilization, we apply three perturbations: 
(i)~prompt-level masking, replacing drug names or KG-entity strings with placeholders on LLM-FT, measuring $\Delta$AUC-ROC vs.\ the unmasked condition (Appendix~\ref{app-E.2:masking-full}); 
(ii)~cross-method drug-replacement, swapping one drug in each S2 positive pair with an opposite-label cold alternate and quantifying how this swap shifts the prediction (Appendix~\ref{app-E.1:diagnostic-metrics}); 
and (iii)~attention-level intervention, zeroing attention weights from the answer token to a target token class inside LLM-FT (Appendix~\ref{app-E.4:attention}).

\textbf{Finding \#3 (prompt-level): Type A is entity-dominated, and Type B is name-dominated with cross-channel compensation.}
\cref{tab:masking-effects} shows entity masking dominates Type A (PK-A $-8.0$, PD-A $-2.9$ vs.\ name $+1.5$, $-0.6$) and name masking dominates Type B (PK-B $-4.4$, PD-B $-3.4$ vs.\ entity $+0.4$, $-1.6$).
KSAI (cross-channel asymmetry) is modest on Type A ($+0.025$) and larger on Type B ($+0.040$, \cref{fig:channel-interaction}), so compensation is concentrated where no mediator anchors the prediction.
Temporal analysis, direct inference, and input masking jointly inform the assessment of pretraining memorization in Appendix~\ref{app-D.8:pretraining-memorization}.


\begin{table}[!t]
\centering
\caption{KPS-F and A--B gap by method class (\cref{sec:experimental-setup}), 3-seed mean$\pm$std on 800-drug subset, S2.}
\label{tab:kps-summary}
\footnotesize
\setlength{\tabcolsep}{5pt}
\renewcommand{\arraystretch}{1}
\begin{tabular}{@{}ll cc@{}}
\toprule
\textbf{Group} & \textbf{Method} & \textbf{Absolute KPS-F} & \textbf{A--B gap} \\
\midrule
Matrix
 & DeepDDI                   & $0.317_{\pm0.066}$ & $+0.052_{\pm0.030}$ \\
\hdashline
\multirow{3}{*}{Mol-Graph}
 & SSI-DDI                   & $0.164_{\pm0.056}$ & $+0.022_{\pm0.014}$ \\
 & DSN-DDI                   & $0.412_{\pm0.065}$ & $+0.035_{\pm0.021}$ \\
 & HDN-DDI                   & $0.154_{\pm0.043}$ & $+0.032_{\pm0.010}$ \\
\hdashline
KG-only
 & EmerGNN                   & $0.388_{\pm0.016}$ & \textbf{$+0.170_{\pm0.057}$} \\
\hdashline
\multirow{2}{*}{KG-integrated}
 & TIGER                     & $0.216_{\pm0.074}$ & $+0.008_{\pm0.018}$ \\
 & MKG-FENN                  & $0.075_{\pm0.030}$ & $-0.007_{\pm0.003}$ \\
\hdashline
Pretrained LM
 & TextDDI                   & $0.310_{\pm0.058}$ & $+0.042_{\pm0.007}$ \\
\hdashline
\rowcolor{gray!20}
LLM (fine-tuned)
 & Llama-3.2-1B (P4)         & $0.336_{\pm0.030}$ & \textbf{$+0.145_{\pm0.039}$} \\
\bottomrule
\end{tabular}
\end{table}

\begin{table}[!t]
\caption{Attention intervention on LLM-FT (Llama-3.2-1B, S2, single random seed). Attention from the answer token to KG-entity (resp.\ name) tokens is zeroed across every layer and head. $\Delta$AUROC vs.\ unintervened baseline.}
\label{tab:attn_intervention}
\centering
\footnotesize
\setlength{\tabcolsep}{6pt}
\renewcommand{\arraystretch}{1}
\begin{tabular}{@{}l cccc@{}}
\toprule
\textbf{Intervention} & \textbf{PK-A} & \textbf{PK-B} & \textbf{PD-A} & \textbf{PD-B} \\
\midrule
Zero KG attention   & $\boldsymbol{-0.219}$ & $-0.061$ & $-0.191$ & $-0.081$ \\
Zero name attention & $+0.000$ & \textbf{$-0.040$} & $-0.002$ & $\boldsymbol{-0.036}$ \\
\bottomrule
\end{tabular}
\end{table}

\textbf{Finding \#4 (cross-method): Only LLM-FT and EmerGNN show A--B sensitivity, and other methods remain bridge-blind.}
\cref{tab:kps-summary} reports KPS-F and its A--B gap across nine methods.
EmerGNN ($+0.170$) and LLM-FT ($+0.145$) stand out with large A--B gaps. Type A drug swaps shift their predictions more than Type B swaps, indicating active use of mediating entities.
The remaining seven methods (DeepDDI, SSI-DDI, DSN-DDI, HDN-DDI, TIGER, MKG-FENN, TextDDI) show gap $\le 0.052$ despite having molecular, KG, or text access. 
Their predictions are insensitive to whether a mediating entity exists.
KG access alone therefore does not imply bridge utilization. EmerGNN gains it through bidirectional KG flow (\cref{sec:stratified-analysis}), while LLM-FT learns to attend to KG tokens during fine-tuning.
Per-method per-bucket and per-channel KPS values are in Appendix~\ref{app-E.3:kps-full}.

\textbf{Finding \#5 (attention-level): LLM-FT internally routes Type A through KG attention, with weaker channel separation on Type B.}
\cref{tab:attn_intervention} shows a Type-A dissociation.
Zeroing KG attention collapses PK-A ($-0.219$) and PD-A ($-0.191$), while zeroing name attention leaves Type A intact.
This finding matches the masking dissociation in \cref{tab:masking-effects}, ruling out a prompt-format artifact.
The stronger PK-A than PD-B response to KG-attention suppression also holds across three seeds and two LLM architectures (Appendix~\ref{app-E.4.4:attn-stability}).
Relative to Type A, Type B is less affected by KG-attention suppression but more affected by name-attention suppression, consistent with the subtype-dependent masking pattern.
The two interventions differ in scope. Masking destroys channel content globally, whereas attention zeroing blocks the answer token's direct attention to the target channel at every layer.
Target content can reach the answer indirectly via intermediate tokens across layers, and zeroing forces softmax to redistribute attention mass elsewhere. The Type B effect does not isolate active channel use.

\begin{tcolorbox}[colback=gray!5!white, boxsep=0.3pt, title=\textbf{Takeaway}, fonttitle=\small\bfseries]
Cross-method KPS reveals a clear gap. Across nine methods with molecular, KG, or text inputs, only EmerGNN and LLM-FT show large A--B gaps in drug-replacement sensitivity, while the remaining seven remain bridge-blind. LLM-FT can therefore use mediating entities. Prompt-level masking and attention zeroing confirm this. Type A predictions are entity-dominated under masking, and KG attention is the necessary read-out path under intervention. KG access therefore does not imply KG utilization. EmerGNN gains this through bidirectional KG flow by design, while LLM-FT gains it through fine-tuned attention to KG tokens.

\end{tcolorbox}


\subsection{Fine-Tuned LLMs Improve S2 Performance but Remain Bridge-Dependent}
\label{sec:model-diversity}


We audit whether scale, prompting, or fine-tuning produces cold-start DDI capability within the LLM family. 
\cref{tab:llm-family} reports S2 AUC on the 800-drug subset across three open-weight families (Llama, Qwen 2.5, Gemma 3) with multiple sizes and two closed-source models (GPT-4o, Claude Sonnet 4.6), under both direct inference and LoRA fine-tuning on prompt~P4.

\begin{table}[!t]
\centering
\caption{S2 AUC for LLMs (P4, 800-drug subset). FT is 3-seed mean, Inf single-seed. Bold marks per-family best (combined AUC-ROC and Recall, Appendix~\ref{app-D.4:full-results}). Closed-source metrics in \cref{tab:app-e1-closed}.}
\label{tab:llm-family}
\setlength{\tabcolsep}{4pt}
\renewcommand{\arraystretch}{1}
\footnotesize
\begin{tabular}{@{}lcc@{\hspace{10pt}}lcc@{\hspace{10pt}}lcc@{\hspace{10pt}}lc@{}}
\toprule
\multicolumn{3}{c}{\textbf{Llama}} & \multicolumn{3}{c}{\textbf{Qwen 2.5}} & \multicolumn{3}{c}{\textbf{Gemma 3}} & \multicolumn{2}{c}{\textbf{Closed-source}} \\
\cmidrule(lr){1-3} \cmidrule(lr){4-6} \cmidrule(lr){7-9} \cmidrule(lr){10-11}
\textbf{Size} & \textbf{Inf} & \textbf{FT} & \textbf{Size} & \textbf{Inf} & \textbf{FT} & \textbf{Size} & \textbf{Inf} & \textbf{FT} & \textbf{Model} & \textbf{Inf} \\
\midrule
3.2-1B & 0.525 & \textbf{0.776}          & 0.5B & 0.489 & \textbf{0.766} & 1B  & 0.520          & 0.756          & GPT-4o            & 0.710  \\
3.2-3B & 0.459 & 0.785 & 3B   & 0.517 & 0.769 & 4B  & 0.496          & 0.781          & Claude Sonnet 4.6 & 0.751 \\
2-7B   & 0.474 & 0.731          & 7B   & 0.486 & 0.740 & 12B & 0.548          & \textbf{0.798} &                   &           \\
2-13B  & 0.513 & 0.781          & 14B  & 0.510 & 0.750 &     &                &                &                   &           \\
\bottomrule
\end{tabular}%
\end{table}

\textbf{Finding \#6: Direct prompting fails as a cold-start DDI solution, while LoRA fine-tuning under P4 succeeds without a monotonic scale advantage.}
Across Llama, Qwen 2.5, and Gemma 3 from $0.5$B to $14$B, direct inference stays near random and collapses to a single answer, and closed-source GPT-4o and Claude Sonnet 4.6 fall short of the best fine-tuned open-weight LLMs.
LoRA fine-tuning raises S2 AUC to $0.73$--$0.80$ across small and mid-size models, while larger same-family variants do not yield monotonic improvement.
A bridge here is a KG mediating entity linking the two drugs, and \cref{sec:modality-results} shows LLM-FT underperforms on Type B pairs lacking a usable bridge, leaving the fine-tuning gain bridge-dependent. 
Per-model and per-prompt details are in Appendices~\ref{app-D.3:direct-inference} and~\ref{app-D.4:full-results}.
Additional comparisons using clinical drug descriptions, alone or alongside KG context, are reported in Appendix~\ref{app-D.4.1:description-ablation}.

\section{Conclusion}
\label{sec:conclusion}

A key evaluation question for new drugs is whether a DDI predictor can use their drug-level evidence available when interaction history is absent.
This paper proposed \textsc{ColdDDI} to address this question,
which turns cold-start DDI evaluation from an aggregate ranking task into a diagnostic test of knowledge utilization.
The drug-wise splits, mechanism-aware annotations, and controlled perturbations of \textsc{ColdDDI} separate three factors that prior evaluations often collapsed, namely, drug novelty, the availability of biomedical mediators, and the model dependence on them.
Our empirical findings are clear and consistent across model families.
All methods lose accuracy when test drugs lack training-time interaction history; yet, the stratified results show that drug novelty alone does not explain the errors.
The main performance divide is whether
the knowledge graph provides a direct pharmacological bridge between the two drugs.
Drug pairs linked by a shared enzyme, transporter, or target are much easier; pairs without such a mediator require indirect reasoning over weaker or multi-hop evidence, where current models fail more often.
This stratification separates knowledge access from knowledge use.
Several KG-augmented or multimodal baselines receive relevant KG inputs but remain weakly sensitive to mediator removal or drug replacement.
In contrast, the fine-tuned LLM predictions change more substantially when the shared mediators are masked or disrupted. 
When an explicit KG bridge is available for those LLMs, predictions are routed mainly through KG-entity evidence; without such bridge, they rely more heavily on auxiliary information such as drug names. 
The LLM audit further shows that scale and prompting alone do not change this pattern. We conclude that fine-tuned LLMs can be useful cold-start DDI predictors when explicit biomedical mediators are available, but they should not yet be interpreted as general-purpose pharmacological reasoners.
%

%


\paragraph{Limitations and Future Work.}
\textsc{ColdDDI} has several scope limitations. First, the main task is binary DDI prediction; although the same drug-wise protocol extends to multi-class event-type prediction (see results in Appendix~\ref{app-B.4:multiclass}), fine-grained prediction remains affected by the long-tailed distribution of DrugBank interaction types. Second, the PK/PD taxonomy is keyword-based, albeit human-validated (Appendix~\ref{app-A.4:taxonomy-validation}), and may simplify borderline or mixed mechanisms. Third, \textsc{ColdDDI} is derived from DrugBank~5.1.13 alone, so its labels, KG facts, and mediator annotations inherit DrugBank's curation biases and open-world incompleteness. In particular, the absence of an annotated enzyme, transporter, or target bridge should be interpreted as the absence of a direct DrugBank-recorded bridge, not as evidence that no biological mediator exists. 

These limitations also define the next research direction. The hardest cold-start cases are interactions without explicit KG bridges, where models must reason over distributed, indirect, or multi-hop pharmacological evidence rather than retrieve a shared mediator. Promising directions include multi-hop relational attention and pathway-level evidence aggregation, recovery of missing mediator evidence through KG completion or external retrieval, and adaptive integration of molecular, textual, and KG information. The diagnostic indicators introduced here, including KPS and KSAI, can extend naturally to other inductive biomedical prediction tasks where knowledge access and knowledge utilization must be separated.
We release annotations, splits, prompts, evaluation code, and a reconstruction pipeline to support such follow-up work (Appendix~\ref{app-A.6:release}).

\paragraph{Broader Impact.}
\textsc{ColdDDI} is intended to make cold-start DDI evaluation more rigorous and interpretable. By separating drug novelty, mediator availability, and perturbation-sensitive knowledge use, the benchmark helps identify when a model exploits pharmacological evidence and when it instead relies on shortcut signals such as drug-name recall or aggregate co-occurrence patterns. This can support safer model development for pharmacovigilance, post-marketing surveillance, drug repurposing, and early safety screening of newly approved or poorly characterized drugs.
The intended role of \textsc{ColdDDI} is diagnostic and comparative, {\em i.e.,} to expose where current methods use mechanistic evidence, where they fail despite receiving it, and which failure modes must be addressed before cold-start DDI prediction can responsibly inform clinical or regulatory workflows.

\clearpage
\bibliographystyle{unsrtnat}
\bibliography{references}

@article{DeepDDI_2018,
  title={Deep learning improves prediction of drug--drug and drug--food interactions},
  author={Ryu, Jae Yong and Kim, Hyun Uk and Lee, Sang Yup},
  journal={Proceedings of the national academy of sciences},
  volume={115},
  number={18},
  pages={E4304--E4311},
  year={2018},
  publisher={National Academy of Sciences}
}

@article{drugbank5_dataset,
  title={DrugBank 5.0: a major update to the DrugBank database for 2018},
  author={Wishart, David S and Feunang, Yannick D and Guo, An C and Lo, Elvis J and Marcu, Ana and Grant, Jason R and Sajed, Tanvir and Johnson, Daniel and Li, Carin and Sayeeda, Zinat and others},
  journal={Nucleic acids research},
  volume={46},
  number={D1},
  pages={D1074--D1082},
  year={2018},
  publisher={Oxford University Press}
}

@article{ddi_review_1,
  title={Comprehensive evaluation of deep and graph learning on drug--drug interactions prediction},
  author={Lin, Xuan and Dai, Lichang and Zhou, Yafang and Yu, Zu-Guo and Zhang, Wen and Shi, Jian-Yu and Cao, Dong-Sheng and Zeng, Li and Chen, Haowen and Song, Bosheng and others},
  journal={Briefings in bioinformatics},
  volume={24},
  number={4},
  pages={bbad235},
  year={2023},
  publisher={Oxford University Press}
}

@article{cold-start-with_modality,
  title={Evaluation of knowledge graph embedding approaches for drug-drug interaction prediction in realistic settings},
  author={Celebi, Remzi and Uyar, Huseyin and Yasar, Erkan and Gumus, Ozgur and Dikenelli, Oguz and Dumontier, Michel},
  journal={BMC bioinformatics},
  volume={20},
  number={1},
  pages={726},
  year={2019},
  publisher={Springer}
}

@article{setting-for-cold-start,
  title={Cold-start problems in data-driven prediction of drug--drug interaction effects},
  author={Dewulf, Pieter and Stock, Michiel and De Baets, Bernard},
  journal={Pharmaceuticals},
  volume={14},
  number={5},
  pages={429},
  year={2021},
  publisher={MDPI}
}

@article{SSI-DDI-2021,
  title={SSI--DDI: substructure--substructure interactions for drug--drug interaction prediction},
  author={Nyamabo, Arnold K and Yu, Hui and Shi, Jian-Yu},
  journal={Briefings in Bioinformatics},
  volume={22},
  number={6},
  pages={bbab133},
  year={2021},
  publisher={Oxford University Press}
}

@article{GMPNN-CS-2022,
  title={Drug--drug interaction prediction with learnable size-adaptive molecular substructures},
  author={Nyamabo, Arnold K and Yu, Hui and Liu, Zun and Shi, Jian-Yu},
  journal={Briefings in Bioinformatics},
  volume={23},
  number={1},
  pages={bbab441},
  year={2022},
  publisher={Oxford University Press}
}

@article{DSN-DDI-2023,
  title={DSN-DDI: an accurate and generalized framework for drug--drug interaction prediction by dual-view representation learning},
  author={Li, Zimeng and Zhu, Shichao and Shao, Bin and Zeng, Xiangxiang and Wang, Tong and Liu, Tie-Yan},
  journal={Briefings in Bioinformatics},
  volume={24},
  number={1},
  pages={bbac597},
  year={2023},
  publisher={Oxford University Press}
}

@article{HDN-DDI-2025,
  title={HDN-DDI: a novel framework for predicting drug-drug interactions using hierarchical molecular graphs and enhanced dual-view representation learning},
  author={Sun, Jinchen and Zheng, Haoran},
  journal={BMC bioinformatics},
  volume={26},
  number={1},
  pages={28},
  year={2025},
  publisher={Springer}
}

@inproceedings{TextDDI-2023,
  title={Learning to describe for predicting zero-shot drug-drug interactions},
  author={Zhu, Fangqi and Zhang, Yongqi and Chen, Lei and Qin, Bing and Xu, Ruifeng},
  booktitle={Proceedings of the 2023 conference on empirical methods in natural language processing},
  pages={14855--14870},
  year={2023}
}

@article{RADDI-2026,
  title={RADDI: A retrieval augmented framework for drug-drug interaction prediction},
  author={Zhang, Hengbo and Wang, Yingying and Gao, Xinyu and Xiong, Yun},
  journal={Big Data Mining and Analytics},
  volume={9},
  number={2},
  pages={360--375},
  year={2026},
  publisher={TUP}
}

@article{DDI-LLM-2025,
  title={DDI-LLM: Predicting Unseen Drug--Drug Interactions Using Large Language Models and Molecular Graphs},
  author={Tanhaei, Mohammad},
  journal={Intelligent Pharmacy},
  year={2025},
  publisher={Elsevier}
}

@article{Distribution-shift-2025,
  title={Benchmarking drug--drug interaction prediction methods: a perspective of distribution changes},
  author={Shen, Zhenqian and Zhou, Mingyang and Zhang, Yongqi and Yao, Quanming},
  journal={Bioinformatics},
  volume={41},
  number={11},
  pages={btaf569},
  year={2025},
  publisher={Oxford University Press}
}

@article{review-2,
  title={LLMs for Drug-Drug Interaction Prediction Using Textual Drug Descriptors},
  author={De Vito, Gabriele and Ferrucci, Filomena and Angelakis, Athanasios},
  journal={Knowledge-Based Systems},
  pages={115486},
  year={2026},
  publisher={Elsevier}
}

@article{ogb-2020,
  title={Open Graph Benchmark: Datasets for Machine Learning on Graphs},
  author={Hu, Weihua and Fey, Matthias and Zitnik, Marinka and Dong, Yuxiao and Ren, Hongyu and Liu, Bowen and Catasta, Michele and Leskovec, Jure},
  journal={Advances in Neural Information Processing Systems},
  volume={33},
  pages={22118--22133},
  year={2020}
}

@inproceedings{tdc-2021,
  title={Therapeutics Data Commons: Machine Learning Datasets and Tasks for Drug Discovery and Development},
  author={Huang, Kexin and Fu, Tianfan and Gao, Wenhao and Zhao, Yue and Roohani, Yusuf and Leskovec, Jure and Coley, Connor W. and Xiao, Cao and Sun, Jimeng and Zitnik, Marinka},
  booktitle={Advances in Neural Information Processing Systems 34, Datasets and Benchmarks Track},
  year={2021},
  url={https://arxiv.org/abs/2102.09548}
}

@article{dataset-construction-for-metal,
  title={Machine learning-based prediction of drug--drug interactions by integrating drug phenotypic, therapeutic, chemical, and genomic properties},
  author={Cheng, Feixiong and Zhao, Zhongming},
  journal={Journal of the American Medical Informatics Association},
  volume={21},
  number={e2},
  pages={e278--e286},
  year={2014},
  publisher={BMJ Publishing Group BMA House, Tavistock Square, London, WC1H 9JR}
}

@article{SumGNN-2021,
  title={SumGNN: multi-typed drug interaction prediction via efficient knowledge graph summarization},
  author={Yu, Yue and Huang, Kexin and Zhang, Chao and Glass, Lucas M and Sun, Jimeng and Xiao, Cao},
  journal={Bioinformatics},
  volume={37},
  number={18},
  pages={2988--2995},
  year={2021},
  publisher={Oxford University Press}
}

@inproceedings{MKG-FENN-2024,
  title={Mkg-fenn: A multimodal knowledge graph fused end-to-end neural network for accurate drug--drug interaction prediction},
  author={Wu, Di and Sun, Wu and He, Yi and Chen, Zhong and Luo, Xin},
  booktitle={Proceedings of the AAAI Conference on Artificial Intelligence},
  volume={38},
  number={9},
  pages={10216--10224},
  year={2024}
}

@inproceedings{TIGER-2024,
  title={Dual-channel learning framework for drug-drug interaction prediction via relation-aware heterogeneous graph transformer},
  author={Su, Xiaorui and Hu, Pengwei and You, Zhu-Hong and Yu, Philip S and Hu, Lun},
  booktitle={Proceedings of the AAAI conference on artificial intelligence},
  volume={38},
  number={1},
  pages={249--256},
  year={2024}
}

@inproceedings{lora-2022,
  title={{LoRA}: Low-Rank Adaptation of Large Language Models},
  author={Hu, Edward J. and Shen, Yelong and Wallis, Phillip and
               Allen-Zhu, Zeyuan and Li, Yuanzhi and Wang, Shean and
               Wang, Lu and Chen, Weizhu},
  booktitle={International Conference on Learning Representations},
  year={2022},
  url={https://openreview.net/forum?id=nZeVKeeFYf9}
}

@article{EmerGNN-2023,
  title={Emerging drug interaction prediction enabled by a flow-based graph neural network with biomedical network},
  author={Zhang, Yongqi and Yao, Quanming and Yue, Ling and Wu, Xian and Zhang, Ziheng and Lin, Zhenxi and Zheng, Yefeng},
  journal={Nature Computational Science},
  volume={3},
  number={12},
  pages={1023--1033},
  year={2023},
  publisher={Nature Publishing Group US New York}
}

@article{DDInter-2022,
  title={DDInter: an online drug--drug interaction database towards improving clinical decision-making and patient safety},
  author={Xiong, Guoli and Yang, Zhijiang and Yi, Jiacai and Wang, Ningning and Wang, Lei and Zhu, Huimin and Wu, Chengkun and Lu, Aiping and Chen, Xiang and Liu, Shao and others},
  journal={Nucleic acids research},
  volume={50},
  number={D1},
  pages={D1200--D1207},
  year={2022},
  publisher={Oxford University Press}
}

@article{Decagon-2018,
  title={Modeling polypharmacy side effects with graph convolutional networks},
  author={Zitnik, Marinka and Agrawal, Monica and Leskovec, Jure},
  journal={Bioinformatics},
  volume={34},
  number={13},
  pages={i457--i466},
  year={2018},
  publisher={Oxford University Press}
}

@article{DDI-GPT-2024,
  title={Ddi-gpt: Explainable prediction of drug-drug interactions using large language models enhanced with knowledge graphs},
  author={Xu, Chengqi and Bulusu, Krishna C and Pan, Heng and Elemento, Olivier},
  journal={BioRxiv},
  pages={2024--12},
  year={2024},
  publisher={Cold Spring Harbor Laboratory}
}

@article{openddi-2026,
  title={OpenDDI: A Comprehensive Benchmark for DDI Prediction},
  author={Jin, Xinmo and Fan, Bowen and Li, Xunkai and Sun, Henan and Zeng, YuXin and Chen, Zekai and Sun, Yuxuan and Li, Jia and Dai, Qiangqiang and Qin, Hongchao and others},
  journal={arXiv preprint arXiv:2602.00539},
  year={2026}
}

@book{Stockley-book-2018,
  title={Stockley's Drug Interactions},
  editor={Preston, Claire L.},
  edition={11},
  year={2016},
  publisher={Pharmaceutical Press}
}

@book{G-and-G-book-2018,
  title={Goodman \& Gilman's the pharmacological basis of therapeutics},
  author={Brunton, Laurence L and Knollmann, Bj{\"o}rn C and Hilal-Dandan, Randa and others},
  volume={13},
  year={2018},
  publisher={McGraw-Hill Education New York}
}

@article{TWOSIDES-2012,
  title={Data-driven prediction of drug effects and interactions},
  author={Tatonetti, Nicholas P and Ye, Patrick P and Daneshjou, Roxana and Altman, Russ B},
  journal={Science translational medicine},
  volume={4},
  number={125},
  pages={125ra31--125ra31},
  year={2012},
  publisher={American Association for the Advancement of Science}
}

@article{LLAMA2-2023,
  title={Llama 2: Open foundation and fine-tuned chat models},
  author={Touvron, Hugo and Martin, Louis and Stone, Kevin and Albert, Peter and Almahairi, Amjad and Babaei, Yasmine and Bashlykov, Nikolay and Batra, Soumya and Bhargava, Prajjwal and Bhosale, Shruti and others},
  journal={arXiv preprint arXiv:2307.09288},
  year={2023}
}

@article{QWEN2_5-2024,
  title={Qwen2. 5 Technical Report},
  author={Yang, An and Yang, Baosong and Zhang, Beichen and Hui, Binyuan and Zheng, Bo and Yu, Bowen and Li, Chengyuan and Liu, Dayiheng and Huang, Fei and Wei, Haoran and others},
  journal={arXiv preprint arXiv:2412.15115},
  year={2024}
}

@article{GEMMA3-2025,
  title={Gemma 3 Technical Report},
  author={Gemma Team Aishwarya Kamath and Johan Ferret and others},
  journal={ArXiv},
  year={2025},
  volume={abs/2503.19786},
  url={https://api.semanticscholar.org/CorpusID:277313563}
}

@article{GPT4-2023,
  title={Gpt-4 technical report},
  author={Achiam, Josh and Adler, Steven and Agarwal, Sandhini and Ahmad, Lama and Akkaya, Ilge and Aleman, Florencia Leoni and Almeida, Diogo and Altenschmidt, Janko and Altman, Sam and Anadkat, Shyamal and others},
  journal={arXiv preprint arXiv:2303.08774},
  year={2023}
}

@inproceedings{CLAUDE-2024,
  title={The Claude 3 Model Family: Opus, Sonnet, Haiku},
  author={{Anthropic}},
  url={https://api.semanticscholar.org/CorpusID:268232499},
  year={2024}
}

@article{he2025imageddi,
  title={ImageDDI: Image-enhanced molecular motif sequence representation for drug-drug interaction prediction},
  author={He, Yuqin and Ma, Tengfei and Li, Chaoyi and Ma, Pengsen and Xiang, Hongxin and Wang, Jianmin and Liu, Yiping and Song, Bosheng and Zeng, Xiangxiang},
  journal={Information Fusion},
  pages={103574},
  year={2025},
  publisher={Elsevier}
}

@article{LLAMA3-2024,
  title={The llama 3 herd of models},
  author={Grattafiori, Aaron and Dubey, Abhimanyu and Jauhri, Abhinav and Pandey, Abhinav and Kadian, Abhishek and Al-Dahle, Ahmad and Letman, Aiesha and Mathur, Akhil and Schelten, Alan and Vaughan, Alex and others},
  journal={arXiv preprint arXiv:2407.21783},
  year={2024}
}

@article{PharmGKB_2021,
  title={PharmGKB, an integrated resource of pharmacogenomic knowledge},
  author={Gong, Li and Whirl-Carrillo, Michelle and Klein, Teri E},
  journal={Current protocols},
  volume={1},
  number={8},
  pages={e226},
  year={2021},
  publisher={Wiley Online Library}
}

@article{CHeMBL_2024,
  title={The ChEMBL Database in 2023: a drug discovery platform spanning multiple bioactivity data types and time periods},
  author={Zdrazil, Barbara and Felix, Eloy and Hunter, Fiona and Manners, Emma J and Blackshaw, James and Corbett, Sybilla and De Veij, Marleen and Ioannidis, Harris and Lopez, David Mendez and Mosquera, Juan F and others},
  journal={Nucleic acids research},
  volume={52},
  number={D1},
  pages={D1180--D1192},
  year={2024},
  publisher={Oxford University Press}
}

@article{kegg_2025,
  title={KEGG: biological systems database as a model of the real world},
  author={Kanehisa, Minoru and Furumichi, Miho and Sato, Yoko and Matsuura, Yuriko and Ishiguro-Watanabe, Mari},
  journal={Nucleic acids research},
  volume={53},
  number={D1},
  pages={D672--D677},
  year={2025},
  publisher={Oxford University Press}
}

\clearpage
\section*{NeurIPS Paper Checklist}

\begin{enumerate}

\item {\bf Claims}
    \item[] Question: Do the main claims made in the abstract and introduction accurately reflect the paper's contributions and scope?
    \item[] Answer: \answerYes{} 
    \item[] Justification: Yes. \cref{sec:intro} enumerates three contributions, namely (i) a mechanism-readable cold-start evaluation that couples drug-wise S0/S1/S2 splits with PK/PD~$\times$~Type-A/B annotations, (ii) knowledge-utilization diagnostics (controlled masking, drug-replacement sensitivity, KPS, and KSAI), and (iii) a controlled audit of eight conventional baselines and 13 LLMs under the same cold-start protocol. Each is substantiated by experiments in \cref{sec:results} and the appendix tables.

\item {\bf Limitations}
    \item[] Question: Does the paper discuss the limitations of the work performed by the authors?
    \item[] Answer: \answerYes{} 
    \item[] Justification: \cref{sec:conclusion} contains a ``Limitations and Future Work'' paragraph discussing three limitations of ColdDDI: the main paper adopts binary detection (a 215-class extension is reported in \cref{app-B.4:multiclass}), the PK/PD taxonomy is keyword-based with two-annotator human verification (\cref{app-A.4:taxonomy-validation}), and the data source is DrugBank 5.1.13 alone.

\item {\bf Theory Assumptions and Proofs}
    \item[] Question: For each theoretical result, does the paper provide the full set of assumptions and a complete (and correct) proof?
    \item[] Answer: \answerNA{} 
    \item[] Justification: ColdDDI is an empirical benchmark and evaluation study, which does not present formal theorems or proofs.

    \item {\bf Experimental Result Reproducibility}
    \item[] Question: Does the paper fully disclose all the information needed to reproduce the main experimental results of the paper to the extent that it affects the main claims and/or conclusions of the paper (regardless of whether the code and data are provided or not)?
    \item[] Answer: \answerYes{} 
    \item[] Justification: \cref{sec:experimental-setup} states the splits, models, and evaluation protocol.
    \cref{app-C.1:baselines} and \cref{tab:baseline-hparams} list per-baseline hyperparameters.
    \cref{app-D.2:lora} and \cref{tab:lora-hparams} list the LoRA configuration.
    \cref{app-C.2:training-protocol} pins the unified training protocol shared across all methods.
    \cref{app-A.6.2:walkthrough} provides a five-stage reproduction walkthrough from DrugBank XML to evaluation.

\item {\bf Open access to data and code}
    \item[] Question: Does the paper provide open access to the data and code, with sufficient instructions to faithfully reproduce the main experimental results, as described in supplemental material?
    \item[] Answer: \answerYes{} 
    \item[] Justification: All derivative artifacts (mechanism annotations, drug-wise split indices, prompt templates, baseline and LLM evaluation code, and the reconstruction pipeline) are released through the code repository \url{https://github.com/0217ljh/ColdDDI-NeurIPS2026}.
    DrugBank's academic license prohibits redistribution of the underlying records, so users should supply their own DrugBank 5.1.13 XML and run \texttt{reconstruct.py} to rebuild the benchmark.
    The five-stage walkthrough is in \cref{app-A.6.2:walkthrough} and license terms are in \cref{app-A.6.4:license}.

\item {\bf Experimental Setting/Details}
    \item[] Question: Does the paper specify all the training and test details (e.g., data splits, hyperparameters, how they were chosen, type of optimizer, etc.) necessary to understand the results?
    \item[] Answer: \answerYes{} 
    \item[] Justification: Drug-wise splits and per-split edge counts are specified in \cref{sec:cold-start-settings} and \cref{app-B.1:split-stats}.
    The negative-sampling strategy and per-epoch re-sampling protocol are in \cref{app-B.2:negative-sampling}.
    For per-baseline optimizers, learning rates, and architecture dimensions, they are listed in \cref{tab:baseline-hparams}.
    The LoRA adapter configuration, including how it was selected from an Optuna search, is in \cref{tab:lora-hpo-space} and \cref{tab:lora-hparams}.

\item {\bf Experiment Statistical Significance}
    \item[] Question: Does the paper report error bars suitably and correctly defined or other appropriate information about the statistical significance of the experiments?
    \item[] Answer: \answerYes{} 
    \item[] Justification: All cross-method headline comparisons are 3-seed mean over three random seeds (\cref{tab:overall-performance,tab:stratified-recall,tab:masking-effects}).
    \cref{tab:kps-summary} additionally reports mean$\pm$std, and every appendix table that reports per-method or per-bucket numbers (\cref{app-C.3:baseline-full-metrics,app-D.4:full-results,app-D.5:mech-sim-llms,app-E.3:kps-full}) gives mean$\pm$std. 
     The attention-level experiments in \cref{app-E.4:attention} are run on a single random seed.

\item {\bf Experiments Compute Resources}
    \item[] Question: For each experiment, does the paper provide sufficient information on the computer resources (type of compute workers, memory, time of execution) needed to reproduce the experiments?
    \item[] Answer: \answerYes{} 
    \item[] Justification: \cref{tab:baseline-hparams} reports per-baseline parameter counts and wall-clock on a single RTX 5090 for both the 800-drug subset and the 1{,}900-drug full benchmark.
    \cref{app-A.6.3:hardware-software} lists the general hardware and software requirements (Python 3.10, single 5090 or 2$\times$A40 GPU).
    \cref{app-D.2:lora} documents the LLM fine-tuning compute (2$\times$A40 96 GB, $\sim$1{,}400 single-GPU hours total, per-cell 5--20 hours by model size).
    
\item {\bf Code Of Ethics}
    \item[] Question: Does the research conducted in the paper conform, in every respect, with the NeurIPS Code of Ethics \url{https://neurips.cc/public/EthicsGuidelines}?
    \item[] Answer: \answerYes{} 
    \item[] Justification: The dataset is derived from DrugBank under their academic license, with redistribution restrictions respected by releasing only derivative artifacts plus a reconstruction pipeline (\cref{app-A.6.4:license}).
    All prior baselines and benchmarks are cited with their original publications.
    The human-verification annotation protocol uses externally recruited graduate-level annotators with a written rubric, inter-annotator agreement reporting, and third-party conflict resolution (\cref{app-A.4:taxonomy-validation}).

\item {\bf Broader Impacts}
    \item[] Question: Does the paper discuss both potential positive societal impacts and negative societal impacts of the work performed?
    \item[] Answer: \answerYes{} 
    \item[] Justification: \cref{sec:conclusion} contains a ``Broader Impact'' paragraph discussing both the positive impact (more rigorous and interpretable cold-start DDI evaluation) and the cautionary note that \textsc{ColdDDI}'s role is diagnostic and predictions must be validated before responsibly informing clinical or regulatory workflows.
    
\item {\bf Safeguards}
    \item[] Question: Does the paper describe safeguards that have been put in place for responsible release of data or models that have a high risk for misuse (e.g., pretrained language models, image generators, or scraped datasets)?
    \item[] Answer: \answerNA{} 
    \item[] Justification: ColdDDI releases only annotations, splits, prompts, and evaluation code.
    The model artifacts to be released are task-specific LoRA adapters for binary DDI prediction (not generation), and the dataset is derived from the peer-curated DrugBank rather than scraped from the web. 
    Clinical-deployment considerations are addressed in the Broader Impact paragraph of \cref{sec:conclusion}.

\item {\bf Licenses for existing assets}
    \item[] Question: Are the creators or original owners of assets (e.g., code, data, models), used in the paper, properly credited and are the license and terms of use explicitly mentioned and properly respected?
    \item[] Answer: \answerYes{} 
    \item[] Justification: DrugBank 5.1.13 is cited and used under its academic license (\cref{app-A.6.4:license}).
    The eight conventional baselines and 13 LLMs are each cited with their original publications in \cref{sec:baselines} and \cref{app-C.1:baselines}.
    Software dependencies, including RDKit, PEFT, and Transformers, are specified in \texttt{requirements.txt}.

\item {\bf New Assets}
    \item[] Question: Are new assets introduced in the paper well documented and is the documentation provided alongside the assets?
    \item[] Answer: \answerYes{} 
    \item[] Justification: All new artifacts (annotations, splits, prompts, code, diagnostic indicators) are released with the repository layout in \cref{app-A.6.1:repo-structure}, a reproduction walkthrough in \cref{app-A.6.2:walkthrough}, and licensing terms in \cref{app-A.6.4:license}.

\item {\bf Crowdsourcing and Research with Human Subjects}
    \item[] Question: For crowdsourcing experiments and research with human subjects, does the paper include the full text of instructions given to participants and screenshots, if applicable, as well as details about compensation (if any)? 
    \item[] Answer: \answerYes{} 
    \item[] Justification: \cref{app-A.4:taxonomy-validation} documents the two-annotator taxonomy validation protocol, including sampling design, recruitment of graduate-level annotators, the annotation rubric and reference materials, and the conflict-resolution procedure. Annotators received no monetary compensation and participated voluntarily as part of an academic collaboration.

\item {\bf Institutional Review Board (IRB) Approvals or Equivalent for Research with Human Subjects}
    \item[] Question: Does the paper describe potential risks incurred by study participants, whether such risks were disclosed to the subjects, and whether Institutional Review Board (IRB) approvals (or an equivalent approval/review based on the requirements of your country or institution) were obtained?
    \item[] Answer: \answerNA{} 
    \item[] Justification: The annotation task only labels public DrugBank pharmacology descriptions and involves no personal data, identifiable subjects, or interventions, so it does not constitute human-subjects research requiring IRB approval.

\end{enumerate}   

\clearpage
\appendix
\addtocontents{toc}{\protect\setcounter{tocdepth}{2}}

\begingroup
\renewcommand{\contentsname}{Appendix Contents}
\setlength{\parskip}{0pt}
\tableofcontents
\endgroup

\clearpage
\section{Dataset Construction}
\label{app-A:dataset-construction}

\subsection{Filtering Pipeline}
\label{app-A.1:filtering-pipeline}


\textsc{ColdDDI} is constructed from DrugBank~5.1.13~\cite{drugbank5_dataset} (released January 2025), which contains 17{,}430 drug entries spanning small molecules, biologics, and investigational compounds.
We apply a seven-step construction pipeline to retain only clinically relevant small-molecule drugs with machine-readable structural representations.
\cref{tab:filtering} summarizes the per-step statistics.

\textbf{Step 1: Small-Molecule Selection.}
We retain only entries whose DrugBank type is \texttt{small molecule}, discarding biologics (e.g., monoclonal antibodies, peptides) and other non-small-molecule entries. This reduces the set from 17{,}430 to 13{,}166 drugs.

\textbf{Step 2: SMILES Validation.}
Each remaining drug is required to have a SMILES string that can be parsed by RDKit (v2023.09) without errors. Entries with missing, malformed, or unparseable SMILES are removed. Combined with Step~1, 12{,}303 drugs remain.

\textbf{Step 3: Approved-Drug Restriction.}
We restrict to compounds whose DrugBank \texttt{groups} field includes \texttt{approved}, filtering out investigational, experimental, and withdrawn drugs. The intersection of Steps~1--3 yields 2{,}643 drugs.

\textbf{Step 4: DDI Edge Extraction and Description Normalization.}
The 2{,}643 retained drugs participate in 613{,}514 pairwise DDI edges. Each DDI entry carries a natural-language template (e.g., \textit{``Lovastatin may decrease the excretion rate of Metformin, which could result in a higher serum level''}); we strip the two drug names and normalize whitespace to obtain a canonical type string (e.g., \textit{``may decrease the excretion rate of which could result in a higher serum level''}). This normalization yields 451 unique types across the 613{,}514 edges.

\textbf{Step 5: Low-Frequency Type Removal.}
DDI types with fewer than 10 occurrences are discarded to suppress noise from rare descriptions. This removes 230 types (51\%) and 459 edges (0.07\%). Drugs left without any remaining DDI edges are dropped at the same step (drop 492 drugs), leaving 221 types and 613{,}055 edges across 2{,}151 drugs.

\textbf{Step 6: Inorganic and Metal-Containing Molecule Removal.}
Two chemical filters are applied via RDKit atom-level inspection. First, drugs whose molecular structure lacks any carbon atom are excluded (76 drugs), as they represent inorganic compounds. Second, drugs containing common metal elements (alkali, alkaline earth, transition, and post-transition metals) are removed (142 drugs, with some overlap). After also dropping drugs that no longer appear in any remaining DDI edge, 1{,}994 drugs and 565{,}978 pairs remain.

\textbf{Step 7: Low-Degree Drug Removal.}
Drugs with fewer than 10 DDI edges are removed to ensure sufficient interaction context. This removes 94 drugs (predominantly amino acids, vitamin derivatives, and topical agents with sparse documented interactions). The final dataset contains \textbf{1{,}900 drugs}, \textbf{565{,}731 positive DDI pairs}, and \textbf{215 interaction types}.

\begin{table}[h]
\centering
\caption{Filtering pipeline statistics. Each step is applied cumulatively.}
\label{tab:filtering}
\begin{tabular}{clrrr}
\toprule
\textbf{Step} & \textbf{Operation} & \textbf{Drugs} & \textbf{DDI Edges} & \textbf{DDI Types} \\
\midrule
0 & DrugBank 5.1.13 (raw) & 17{,}430 & 1{,}427{,}655 & n/a \\
1 & Retain small molecules & 13{,}166 & 1{,}205{,}013 & n/a \\
2 & Require valid RDKit SMILES & 12{,}303 & 1{,}161{,}857 & n/a \\
3 & Restrict to approved compounds & 2{,}643 & 613{,}514 & n/a \\
4 & Extract DDI edges \& normalize descriptions & 2{,}643 & 613{,}514 & 451 \\
5 & Remove DDI types with $<$10 occurrences & 2{,}151 & 613{,}055 & 221 \\
6 & Remove inorganic \& metal-containing drugs & 1{,}994 & 565{,}978 & 215 \\
7 & Remove low-degree drugs (degree $<$ 10) & \textbf{1{,}900} & \textbf{565{,}731} & \textbf{215} \\
\bottomrule
\end{tabular}
\end{table}

\subsection{PK/PD Keyword Matching}
\label{app-A.2:pkpd-keywords}


Each DDI type in DrugBank is associated with a canonical mechanism description template that implicitly encodes whether the interaction operates through a PK or PD pathway~\cite{Stockley-book-2018,G-and-G-book-2018}.
PK interactions concern the absorption, distribution, metabolism, or excretion (ADME) of one or both drugs (for example, inhibition of a shared CYP enzyme leading to elevated serum concentration).
PD interactions modify the pharmacological effect at the target site (such as additive CNS depression when two central-nervous-system depressants are co-administered).

We classify the 215 DDI types into PK or PD categories by matching their description templates against two curated keyword lists:

\begin{itemize}[leftmargin=*,itemsep=1pt,topsep=1pt,parsep=0pt]
    \item \textbf{PK keywords} (ADME-related language): \textit{metabolism, excretion, absorption, serum concentration, bioavailability, clearance, CYP, transporter}, among other ADME-process descriptors.
    \item \textbf{PD keywords} (pharmacological-effect language): \textit{risk}, \textit{adverse effect}, \textit{activities} (e.g., antihypertensive, anticoagulant), \textit{therapeutic efficacy, effectiveness, bleeding, hemorrhage, QTc} (prolongation), \textit{sedation, hypotensive, hypertension, arrhythmia, tachycardia, nephrotoxicity, bradycardia, analgesic, myopathy, CNS depressant}, alongside additional adverse-effect descriptors enumerated in the released classifier.
\end{itemize}

A DDI type is assigned to PK if its template matches at least one PK keyword, and to PD if it matches at least one PD keyword.

\paragraph{PK-Precedence Tiebreak.}
When both keyword sets match (for example, the template \textit{``can cause an increase in the absorption of resulting in an increased serum concentration and potentially a worsening of adverse effects''} matches the PK keywords \textit{absorption} and \textit{serum concentration} alongside the PD keyword \textit{adverse effect}), we assign the label PK following the mechanism-based classification convention used in standard pharmacology reference texts~\cite{Stockley-book-2018,G-and-G-book-2018}. Interactions are classified by their initiating event rather than by their clinical manifestation.
In the example above, the ADME terms describe the initiating mechanism while the effect term describes the downstream manifestation, so the underlying mechanism is pharmacokinetic.
Under this rule, all 215 types receive an unambiguous PK/PD assignment, with one borderline type (84 edges) flagged as Mixed for analysis but treated as PK in the main evaluation.

The classification is performed at the DDI-type level; each of the 565{,}731 edges inherits the PK/PD label of its type. \cref{tab:pkpd-distribution} summarizes the resulting distribution.

\begin{table}[h]
\centering
\caption{PK/PD classification results. Mixed indicates types where both keyword sets match. In evaluation, Mixed is subsumed under PK via the precedence rule.}
\label{tab:pkpd-distribution}
\begin{tabular}{lrrr}
\toprule
\textbf{Label} & \textbf{\# DDI Types} & \textbf{\# DDI Edges} & \textbf{Edge \%} \\
\midrule
PK                         & 30  & 295{,}406 & 52.2\% \\
PD                         & 184 & 270{,}241 & 47.8\% \\
Mixed (resolved as PK)     & 1   & 84        & $<$0.1\% \\
\midrule
\textbf{Total}             & \textbf{215} & \textbf{565{,}731} & \textbf{100\%} \\
\bottomrule
\end{tabular}
\end{table}

Although PK accounts for only 30 of 215 types (14.0\%), it covers 52.2\% of all edges because the two highest-frequency types, \textit{metabolism} (99{,}273 edges) and \textit{excretion} (96{,}650 edges), are both PK.
PD spans 184 types but is distributed across a broader range of pharmacological effects: the single keyword \textit{risk} (templates such as \textit{``The risk or severity of CNS depression can be increased''}) covers 123 of these types (179{,}729 edges), while finer-grained categories (\textit{activities} 56 types, \textit{therapeutic efficacy} 4 types, \textit{QTc} 8 types, \textit{bleeding} 9 types) capture the remaining PD interactions. Types matching multiple keyword categories are counted in each, so the finer-grained subcounts overlap.
This asymmetry reflects a general pharmacological distinction: PK interactions converge on a limited set of ADME mechanisms (primarily hepatic enzyme inhibition/induction), whereas PD interactions arise from a much wider variety of receptor- and pathway-level effects.

\subsection{A/B Subclassification Details}
\label{app-A.3:ab-subdivision}


The PK/PD classification captures the \emph{type} of mechanism but does not indicate whether the knowledge graph (KG) contains sufficient information to support explicit reasoning about that mechanism.
We introduce a second axis (Type~A vs.\ Type~B) based on whether one or more shared biomedical entities that mediate the interaction are annotated in the KG for a given drug pair.

\paragraph{Shared-Entity Identification Protocol.}
For each positive DDI pair $(d_i, d_j)$, we query the DrugBank-derived KG for shared biomedical entities through which the interaction plausibly operates:
\begin{itemize}[leftmargin=*,itemsep=1pt,topsep=1pt,parsep=0pt]
   
   \item \textbf{PK Pairs}: we search for shared \textbf{enzymes} or \textbf{transporters}, checking whether $\mathrm{enzymes}(d_i) \cup \mathrm{transporters}(d_i)$ and $\mathrm{enzymes}(d_j) \cup \mathrm{transporters}(d_j)$ share at least one entity with a valid action-pair combination (e.g., substrate-inhibitor, substrate-inducer). If such an entity is found, the pair is classified as \textbf{PK-A}; otherwise \textbf{PK-B}.

    \item \textbf{PD Pairs}: we search for shared \textbf{pharmacological targets}, checking whether $\mathrm{targets}(d_i)$ and $\mathrm{targets}(d_j)$ share at least one entity with compatible actions (e.g., antagonist-antagonist, agonist-agonist, agonist-antagonist). If found, the pair is classified as \textbf{PD-A}; otherwise \textbf{PD-B}.
\end{itemize}

\paragraph{Mechanism Chain Representation.}
For each Type~A pair, we record a structured mechanism chain that traces the interaction pathway through the shared mediating entity or entities. Arrow directions in the chains below depict the biological causality of the interaction (who acts on what), not DrugBank's fixed annotation direction (drug $\rightarrow$ entity). Two chain patterns emerge:

\textbf{PK chains} are directed: the two drugs play asymmetric roles, with one drug modulating a mediating enzyme or transporter that in turn acts on the other drug. For clarity we depict the single-entity case:
\[
d_i \xrightarrow{\text{[inhibitor]}} \text{CYP3A4} \xrightarrow{\text{[substrate]}} d_j
\]
For example, \textit{Lumacaftor $\rightarrow$ [inhibitor] $\rightarrow$ ATP-dependent translocase ABCB1 $\rightarrow$ [substrate] $\rightarrow$ Daptomycin}. The interaction arises because one drug modifies the activity of the enzyme or transporter that processes the other (a \emph{shared-entity pattern matching} signal); 42\% of PK-A pairs share two or more such entities.

\textbf{PD chains} are convergent: both drugs adopt the same role and act independently on one or more shared targets, in a symmetric arrangement:
\[
d_i \xrightarrow{\text{[inhibitor]}} \text{Prothrombin} \xleftarrow{\text{[inhibitor]}} d_j
\]
For example, \textit{Bivalirudin $\rightarrow$ [inhibitor] $\rightarrow$ Prothrombin $\leftarrow$ [inhibitor] $\leftarrow$ Dabigatran etexilate}. PD interactions often involve multiple overlapping targets (45\% of PD-A pairs share two or more) rather than a single dominant entity.

\paragraph{Key-Entity Statistics.}
\cref{tab:ab-stats} summarizes the distribution. The 84 Mixed-type edges are resolved as PK via the precedence rule and included in the PK-A/PK-B split accordingly (21 edges as PK-A and 63 as PK-B).

\begin{table}[h]
\centering
\caption{A/B subdivision statistics across all 565{,}731 DDI edges. \texttt{n/a} marks columns inapplicable to Type~B by construction (no shared mediating entity).}
\label{tab:ab-stats}
\resizebox{\textwidth}{!}{%
\begin{tabular}{lrrrr}
\toprule
\textbf{Group} & \textbf{\# Edges} & \textbf{Edge \%} & \textbf{\# Unique Key Entities} & \textbf{Dominant Entity} \\
\midrule
PK-A & 167{,}718 & 29.6\% & 79 (enzyme: 38, transporter: 41) & CYP3A4 (48.0\% of PK-A) \\
PK-B & 127{,}772 & 22.6\% & n/a                              & n/a \\
PD-A & 21{,}686  & 3.8\%  & 282 (target: 282)                & Histamine H1 receptor (9.7\% of PD-A) \\
PD-B & 248{,}555 & 43.9\% & n/a                              & n/a \\
\bottomrule
\end{tabular}%
}
\end{table}

Four observations are noteworthy:

\textbf{(1)~PK-A is dominated by a small set of enzymes.}
Among the 79 unique key entities in PK-A, CYP3A4 alone accounts for 80{,}432 edges (48.0\%), followed by CYP2D6 (13.9\%) and CYP1A2 (8.8\%). The top-5 entities (CYP3A4, CYP2D6, CYP1A2, CYP3A5, and ABCB1 transporter) cover 80.8\% of all PK-A edges. This concentration reflects the central role of cytochrome P450 enzymes in drug metabolism.

\textbf{(2)~PD-A is distributed across many targets.}
In contrast, PD-A spans 282 unique targets with a much flatter distribution: Histamine H1 receptor (2{,}109 edges) accounts for only 9.7\% of PD-A. The set includes dopamine (D2), muscarinic (M1, M2), adrenergic ($\beta$-1, $\alpha$-1A), GABAergic, serotonergic, and opioid receptors. PD-A pairs typically expose multiple bridges rather than a single dominant one, which is the structural property the masking experiments exploit on Type A pairs (\cref{sec:modality-results}).

\textbf{(3)~PD-B is the largest group.}
Nearly half (43.9\%) of all DDI edges are PD-B, where no shared target can be identified in the KG.
These interactions operate through convergent downstream pathways. For example, Aspirin and Warfarin both elevate bleeding risk via distinct multi-hop chains:
\[
\text{Aspirin} \xrightarrow{\text{[inhibits]}} \text{COX-1/2} \to \downarrow\text{thromboxane A2} \to \downarrow\text{platelet aggregation (primary hemostasis)}
\]
\[
\text{Warfarin} \xrightarrow{\text{[inhibits]}} \text{VKORC1} \to \downarrow\text{active vitamin K} \to \downarrow\text{factors II/VII/IX/X (coagulation cascade)}
\]
The two chains share no entity at any single hop in the KG. They converge only at the system-level outcome (impaired hemostasis $\to$ bleeding). The mediating path is therefore multi-hop and reasoning has to go beyond shared-bridge utilization (\cref{sec:modality-results}).

\textbf{(4)~A/B is not reducible to PK/PD.}
PK has a relatively balanced A/B ratio (56.8\% A vs.\ 43.2\% B), whereas PD is heavily skewed toward B (92.0\% B vs.\ 8.0\% A). 
This asymmetry arises because metabolic enzymes and transporters are well-documented in DrugBank, while pharmacological targets are more diverse and less systematically annotated.

\paragraph{Action-Pair Patterns.}

The mechanism-chain narrative above predicts role-asymmetry in PK and role-symmetry in PD; the action-pair frequencies in Type A edges quantify exactly this contrast.
\cref{tab:pka-actions,tab:pda-actions} report the top-5 combinations under the shared entities convention (an edge with multiple shared entities contributes to multiple rows).

\begin{table}[h]
\centering
\begin{minipage}[t]{0.46\textwidth}
\centering
\captionof{table}{PK-A action pairs.}
\label{tab:pka-actions}
\begin{tabular}{lrr}
\toprule
\textbf{Action Pair} & \textbf{\# Edges} & \textbf{\%} \\
\midrule
substrate-inhibitor & 72{,}542 & 43.2\% \\
inhibitor-substrate & 55{,}031 & 32.8\% \\
substrate-inducer   & 20{,}248 & 12.1\% \\
inducer-substrate   & 15{,}027 & 9.0\%  \\
inhibitor-inhibitor & 4{,}889  & 2.9\%  \\
\bottomrule
\end{tabular}
\end{minipage}
\hfill
\begin{minipage}[t]{0.46\textwidth}
\centering
\captionof{table}{PD-A action pairs.}
\label{tab:pda-actions}
\begin{tabular}{lrr}
\toprule
\textbf{Action Pair} & \textbf{\# Edges} & \textbf{\%} \\
\midrule
antagonist-antagonist & 7{,}758 & 35.7\% \\
inhibitor-inhibitor   & 3{,}466 & 16.0\% \\
agonist-agonist       & 2{,}059 & 9.5\%  \\
antagonist-agonist    & 1{,}549 & 7.1\%  \\
agonist-antagonist    & 1{,}526 & 7.0\%  \\
\bottomrule
\end{tabular}
\end{minipage}
\end{table}

PK-A's four modulator--substrate rows (43.2 / 32.8 / 12.1 / 9.0\%) each outweigh symmetric inhibitor--inhibitor (2.9\%). 
PD-A is conversely dominated by same-role co-action (antagonist--antagonist 35.7\%, inhibitor--inhibitor 16.0\%, agonist--agonist 9.5\%), with the two mixed-direction rows at 7.1\% / 7.0\%. 
\emph{Inhibitor} refers to different entity types across the two tables (shared enzyme/transporter in PK-A, shared pharmacological target in PD-A), so the two patterns are not strict mirrors: PK's asymmetry sits at the mechanism level (substrate vs.\ modulator), PD's symmetry at the action level (both drugs modulate the target).
The asymmetric PK signature underlies the directed shared-entity pattern matching exploited in \cref{sec:stratified-analysis}. The symmetric PD signature underlies the convergent multi-bridge structure of finding~(2).

\subsection{Taxonomy Validation}
\label{app-A.4:taxonomy-validation}


This subsection details the human-verification protocol used to assess the reliability of the automated PK/PD and A/B taxonomy.

\subsubsection{Sampling Design}
From the 215 DDI types we uniformly sample 50 types, which collectively cover $\mathbf{60.2\%}$ of all 565{,}731 DDI edges.
For each sampled type we further uniformly sample $K{=}10$ drug pairs from its DDI edges, yielding $N{=}500$ pairs in total.
Unlike the automated keyword matching, which assigns one PK/PD label per type from the DrugBank mechanism description, annotators label both PK/PD and A/B per pair. 
This finer granularity captures within-type variation across pairs.

\subsubsection{Annotator Recruitment}
Two annotators are recruited externally (not co-authors of \textsc{ColdDDI}) through graduate-program mailing lists in pharmacy and medicinal chemistry.
Both annotators hold graduate-level training in pharmacology and are familiarized with DrugBank's data schema and interaction description conventions during a 30-minute onboarding session.
Annotators label independently. Neither has access to the other's labels or to the automated labels during annotation.
Annotators received no monetary compensation. Participation was voluntary as part of an academic collaboration.

\subsubsection{Annotation Rubric}
For each item, annotators are shown two pieces of information. 
\textbf{(a).}~The DrugBank mechanism description and the corresponding DDInter~\cite{DDInter-2022} mechanism description authored by clinical pharmacists. 
\textbf{(b).}~All relevant biomedical entities from the DrugBank-derived KG used in this paper, together with each drug's role (substrate, inducer, inhibitor, agonist, antagonist) as recorded in DrugBank.

Annotators assign two labels per item:
\begin{itemize}[leftmargin=*,itemsep=1pt,topsep=1pt,parsep=0pt]
    \item \textbf{Task 1 (PK/PD).} The mechanism category (PK, PD, or Mixed) based on whether the described mechanism is ADME-related, effect-related, or both.
    
    \item \textbf{Task 2 (A/B).} Whether the automatically identified shared entity is pharmacologically valid and actually mediates the interaction via the proposed action pair. Select Type~A or Type~B.
\end{itemize}

As reference material, annotators may consult Stockley's Drug Interactions~\cite{Stockley-book-2018}, Goodman \& Gilman's Pharmacological Basis of Therapeutics~\cite{G-and-G-book-2018}, and the FDA Drug Interaction Guidance for Industry (2020). 
Annotators may additionally consult general-purpose large language models for inspiration when interpreting ambiguous mechanism descriptions.
There is no time limit per item, and annotators are instructed to flag any item with ambiguous wording for later discussion.

\subsubsection{Conflict Resolution}
After independent annotation, disagreeing items are first resolved by discussion between the two annotators, who together re-examine DrugBank evidence and the reference material.
Items that remain unresolved after discussion are forwarded to a third senior expert (not a co-author, with faculty-level pharmacology credentials) for adjudication. This third-party label is taken as the consensus.

\subsubsection{Metrics}
We report four metrics.
\begin{enumerate}[leftmargin=*,itemsep=2pt]
    \item \textbf{Inter-annotator agreement (Cohen's $\kappa$)} for PK/PD per-pair ($n{=}500$), PK/PD type-level ($n{=}50$ via majority vote of pair labels), and A/B per-pair ($n{=}500$).
    \item \textbf{Automated-vs-consensus Precision, Recall, F1} for PK/PD classification at both per-pair and type-level granularity, reported with PK as positive and with PD as positive.
    \item \textbf{Automated-vs-consensus Precision, Recall, F1} for A/B classification at the per-pair level, reported with A as positive and with B as positive, together with the overall agreement \% and a per-quadrant agreement breakdown over PK-A / PK-B / PD-A / PD-B.
\end{enumerate}

\subsubsection{Results}
\label{app-A.4.6:results}

We report two-annotator results on the $N{=}500$ pairs. Annotator~1's $10$ Mixed PK/PD labels are collapsed to PK by the \cref{app-A.2:pkpd-keywords} PK-precedence convention. The $39$ PK/PD disagreements and $36$ A/B disagreements between the two annotators ($68$ unique pairs in total, with $7$ overlapping both tasks) are resolved by a third-party judge into a final consensus label per pair, yielding complete consensus coverage on all $500$ pairs.

\paragraph{Inter-Annotator Agreement.}
Cohen's $\kappa$ between the two annotators is $\mathbf{0.842}$ for PK/PD per-pair ($n{=}500$), $\mathbf{0.839}$ for PK/PD type-level ($n{=}50$ types via majority vote of pair labels), and $\mathbf{0.805}$ for A/B per-pair ($n{=}500$). All three values fall in the substantial-to-almost-perfect range. See \cref{tab:taxonomy-pilot-quadrants}.

\paragraph{Auto-vs-Consensus PK/PD Agreement.}
The automated keyword matcher matches the final consensus on $445/500$ pairs (overall agreement $\mathbf{89.0\%}$). PK-as-positive achieves $F1{=}0.884$ with $P{=}0.840$ and $R{=}0.933$. PD-as-positive achieves $F1{=}0.895$ with $P{=}0.940$ and $R{=}0.855$. Aggregating to type level via majority vote within type yields $47/50$ types agreement ($\mathbf{94.0\%}$) with $F1{=}0.936$ for PK and $F1{=}0.943$ for PD. Full numbers in \cref{tab:taxonomy-pilot-primary}.

\paragraph{Auto-vs-Consensus A/B Agreement.}
Overall agreement with the automated taxonomy is $469/500{=}\mathbf{93.8\%}$ (\cref{tab:taxonomy-pilot-primary}). A-as-positive achieves $F1{=}0.893$ with $P{=}0.806$ and $R{=}1.000$. B-as-positive achieves $F1{=}0.956$ with $P{=}1.000$ and $R{=}0.916$. The asymmetry (perfect A recall but lower A precision, and perfect B precision but lower B recall) shows that the automated procedure over-claims A rather than missing it. The per-quadrant breakdown (\cref{tab:taxonomy-pilot-quadrants}) is consistent with this. Type-B groups achieve perfect agreement (PK-B $119/119{=}100\%$, PD-B $221/221{=}100\%$), and PD-A reaches $27/29{=}93.1\%$. PK-A is the weakest quadrant at $102/131{=}77.9\%$. The residual PK-A disagreements concentrate in cases where the automated procedure marks the pair PK-A whenever any compatible CYP shared entity exists, while the consensus demands an unambiguous mechanistic driver.

\begin{table}[h]
\centering
\caption{Auto-vs-consensus agreement on the final judge-resolved consensus ($N{=}500$). F1 columns report PK and PD as positive class for the PK/PD task, and A and B as positive class for the A/B task.}
\label{tab:taxonomy-pilot-primary}
\small
\setlength{\tabcolsep}{6pt}
\renewcommand{\arraystretch}{1.05}
\begin{tabular}{@{}llrrrr@{}}
\toprule
\textbf{Task} & \textbf{Subset} & $n$ & \textbf{Agreement} & \textbf{F1 (pos.\,1)} & \textbf{F1 (pos.\,2)} \\
\midrule
PK/PD per-pair    & all                 & 500 & 89.0\% & 0.884 (PK) & 0.895 (PD) \\
PK/PD type-level  & majority-vote types &  50 & 94.0\% & 0.936 (PK) & 0.943 (PD) \\
A/B per-pair      & overall             & 500 & 93.8\% & 0.893 (A)  & 0.956 (B)  \\
\bottomrule
\end{tabular}
\end{table}

\begin{table}[h]
\centering
\caption{A/B per-quadrant breakdown (computed on the automated assignment) and inter-annotator agreement (Cohen's $\kappa$, computed on the full $N{=}500$ paired labels). Per-quadrant agreement equals the precision of the automated label restricted to that quadrant.}
\label{tab:taxonomy-pilot-quadrants}
\small
\setlength{\tabcolsep}{6pt}
\renewcommand{\arraystretch}{1.05}
\begin{tabular}{@{}llrr@{}}
\toprule
\textbf{Task} & \textbf{Subset} & $n$ & \textbf{Agreement\,/\,$\kappa$} \\
\midrule
\multicolumn{4}{@{}l}{\emph{A/B per-quadrant (auto vs.\ consensus)}} \\
A/B per-pair      & PK-A                & 131 & 77.9\%  \\
A/B per-pair      & PK-B                & 119 & 100.0\% \\
A/B per-pair      & PD-A                &  29 & 93.1\%  \\
A/B per-pair      & PD-B                & 221 & 100.0\% \\
\midrule
\multicolumn{4}{@{}l}{\emph{Inter-annotator (Cohen's $\kappa$)}} \\
PK/PD per-pair    & all                 & 500 & $\kappa{=}0.842$ \\
PK/PD type-level  & majority-vote types &  50 & $\kappa{=}0.839$ \\
A/B per-pair      & all                 & 500 & $\kappa{=}0.805$ \\
\bottomrule
\end{tabular}
\end{table}

\paragraph{Implication for the Taxonomy.}
The PK/PD per-pair $\kappa{=}0.842$ and A/B per-pair $\kappa{=}0.805$ place the automated stratification in the substantial-to-almost-perfect range, with $100\%$ Type B and $93.1\%$ PD-A agreement.
The PK-A $77.9\%$ comes from multi-CYP pairs where the automated procedure accepts any match while annotators require an unambiguous driver, but \cref{sec:stratified-analysis} compares PK-A and PK-B in aggregate rather than per-pair, so this noise does not affect the bucket-level conclusion.

\subsection{Dataset Statistics and Prior Comparison}
\label{app-A.5:dataset-comparison}

\begin{figure}[h]
\centering
\includegraphics[width=0.8\textwidth]{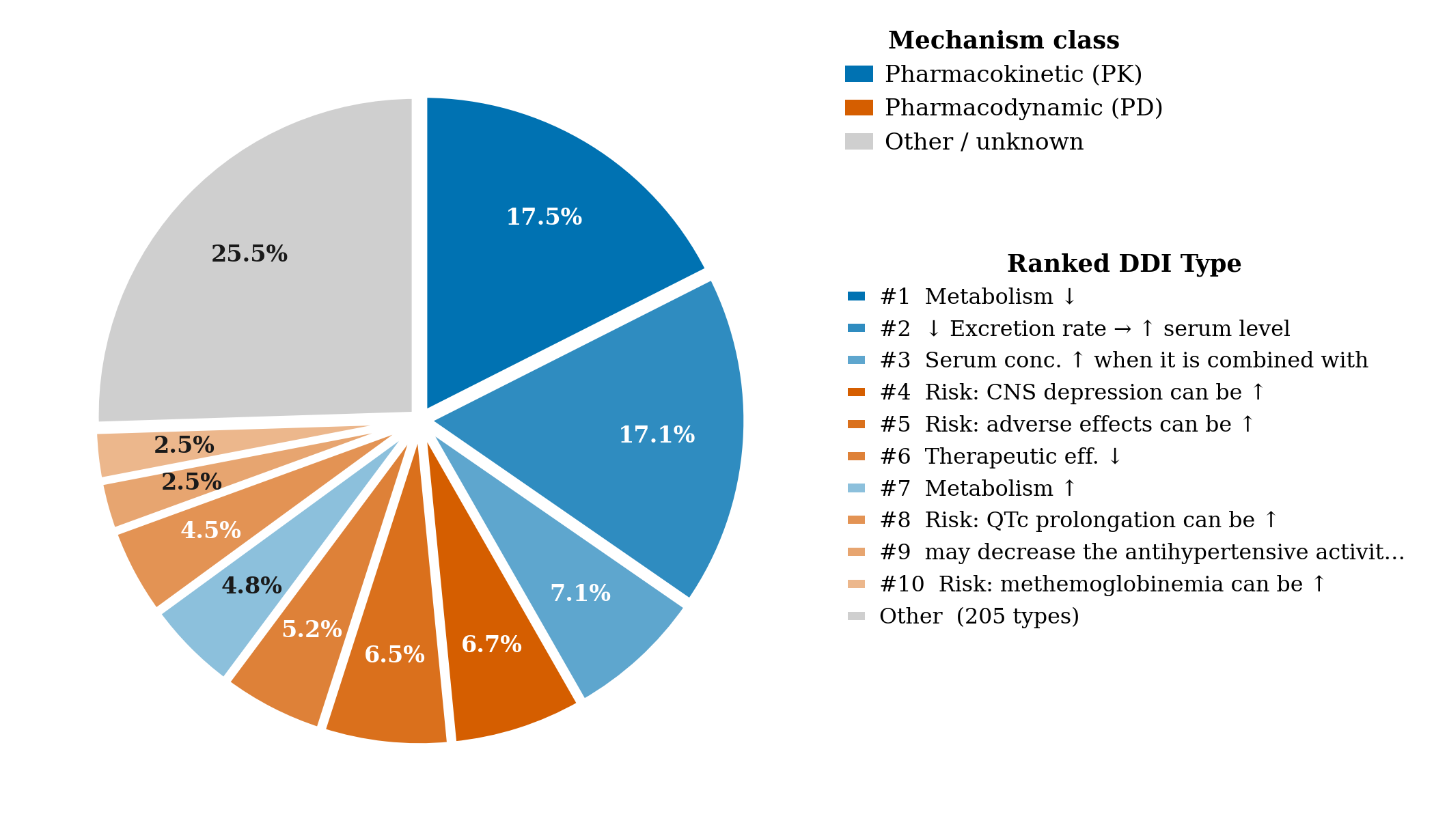}
\caption{Top-10 DDI types by edge count. Mechanism class encoded by hue (blue = PK, orange = PD), with within-class luminance gradient encoding rank. The top-10 types cover 74.5\% of all 565{,}731 edges. The remaining 205 types are pooled into the grey ``Other'' slice.}
\label{fig:ddi-pie}
\end{figure}

\begin{figure}[h]
\centering
\includegraphics[width=0.8\textwidth]{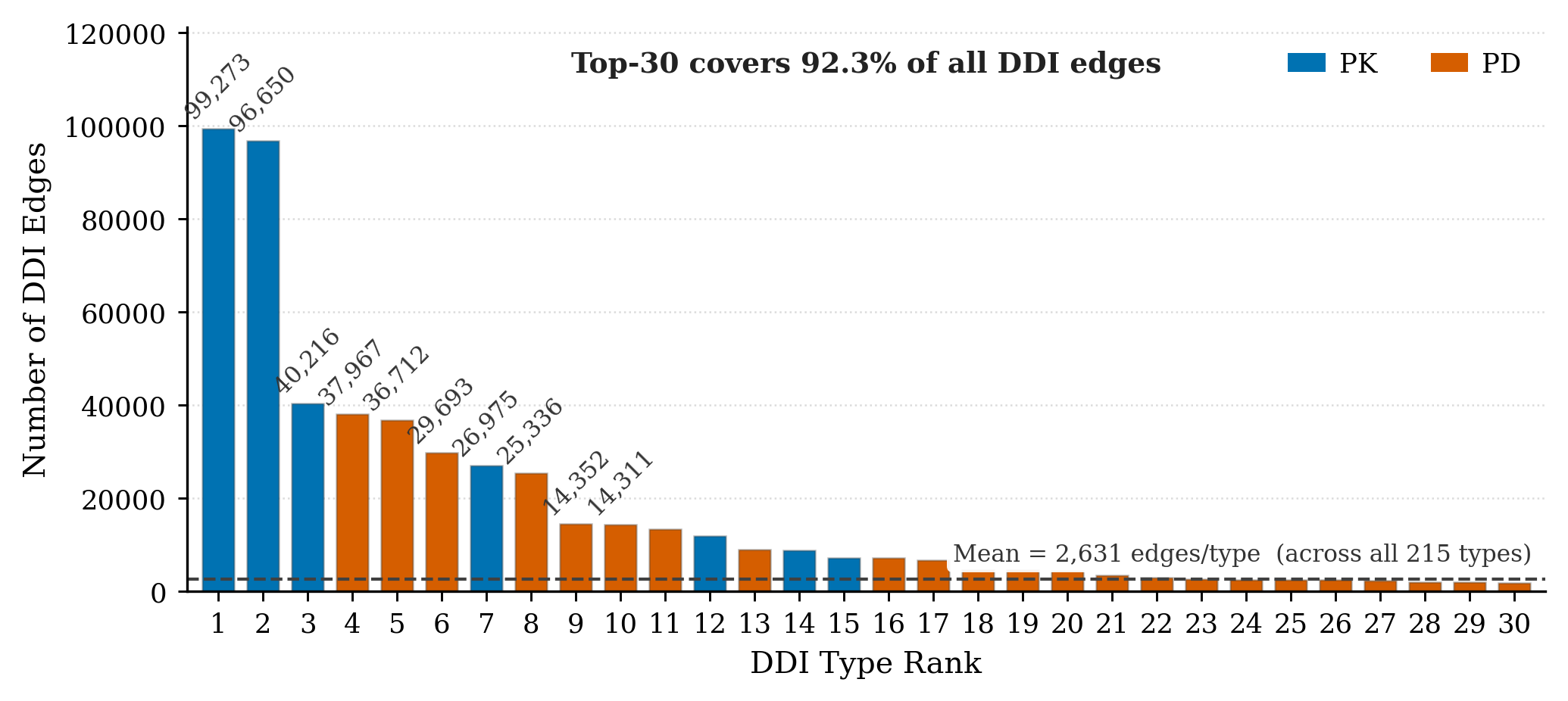}
\caption{Top-30 DDI types on a linear y-axis. Bars are coloured by mechanism class (blue = PK, orange = PD) with raw edge counts annotating the top-10. The dashed horizontal line marks the mean across all 215 types (2{,}631 edges/type). These 30 types cover 92.3\% of the 565{,}731 edges, reflecting the dominance of a small number of ADME mechanisms.}
\label{fig:ddi-top30}
\end{figure}

Even after Step~5 of our pipeline (\cref{app-A.1:filtering-pipeline}) already removes all DDI types with fewer than 10 occurrences, the 215 retained types still follow a pronounced heavy-tailed distribution.
The top-5 types account for 54.9\% of all edges, led by \textit{``The metabolism of can be decreased when combined with''} (99{,}273 edges, 17.5\%) and \textit{``may decrease the excretion rate of''} (96{,}650 edges, 17.1\%).
The top-10 types collectively cover 74.5\% of all edges (\cref{fig:ddi-pie}), and the top-30 cover 92.3\% (\cref{fig:ddi-top30}). 90\% cumulative coverage is reached at rank 24.
In contrast, the bottom 100 types contribute only 0.7\%, with a median of 28 edges per type.

\begin{figure}[h]
\centering
\includegraphics[width=0.8\textwidth]{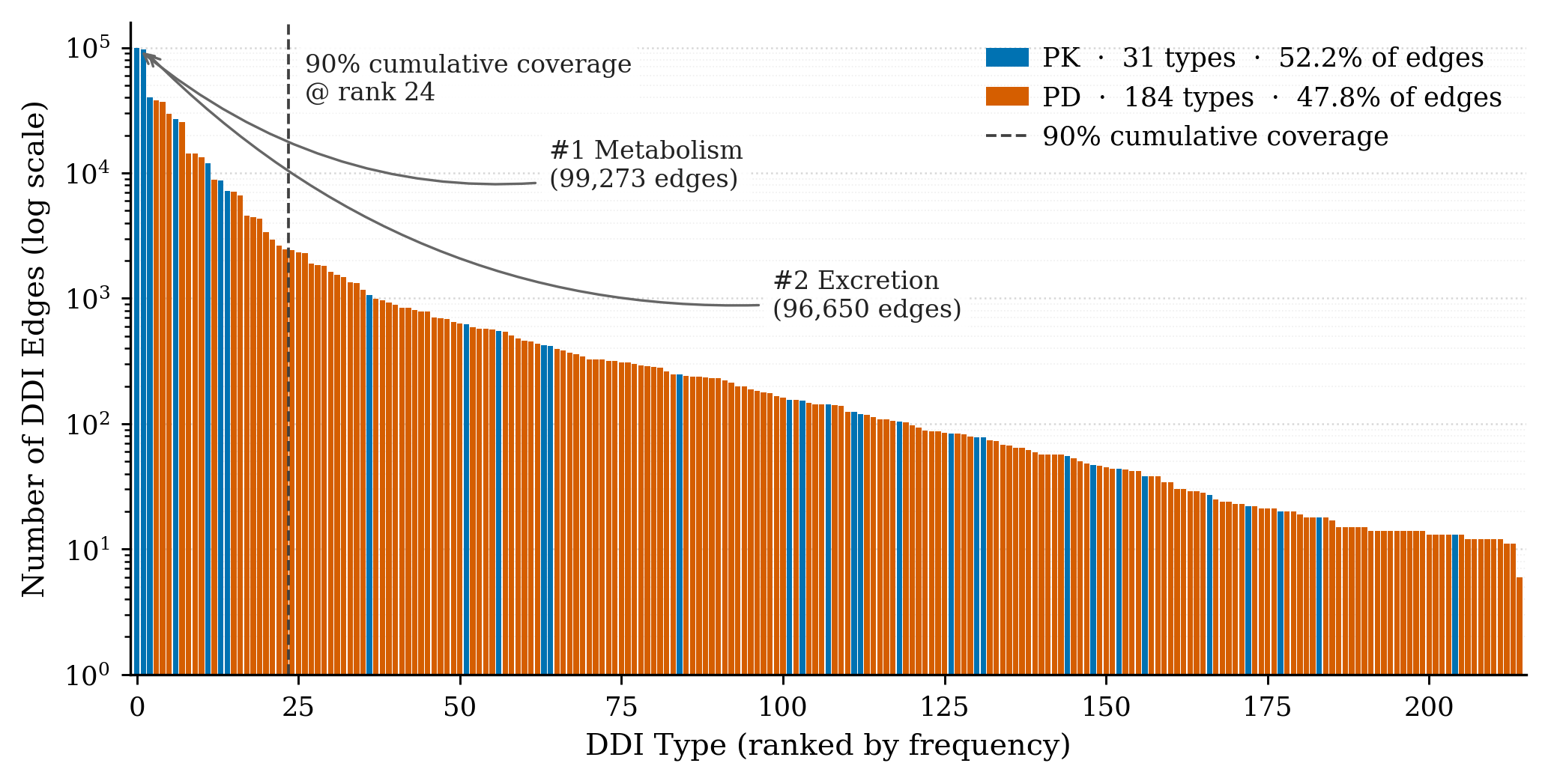}
\caption{All 215 DDI types ranked by frequency on a log-scale y-axis. PK types (blue, 31 incl.\ the 1 Mixed type collapsed to PK per \cref{app-A.2:pkpd-keywords}) concentrate at the high-frequency end while PD types (orange, 184) span a broader, lower-frequency range. The dashed vertical line marks the rank at which cumulative coverage reaches 90\% (rank 24).}
\label{fig:ddi-longtail}
\end{figure}

\cref{fig:ddi-longtail} shows the full rank-frequency curve across all 215 types.
PK types (blue) dominate the high-frequency end while PD types (orange) populate the long tail.
\cref{fig:ddi-top30} zooms into the top-30 types.
The top-3 types (Metabolism, Excretion, Serum concentration) are all PK. PK and PD types alternate from rank 4 onward, with PK dominating the very high-frequency end.

\subsection{Licensing, Artifact, and Reproduction Pipeline}
\label{app-A.6:release}

\subsubsection{Reviewer-accessible artifacts and ED compliance}

\label{app-A.6.0:reviewer-artifacts}

We provide an artifact bundle containing (i) derivative mechanism annotations, split indices, prompt templates, diagnostic scripts, and evaluation code, (ii) a deterministic reconstruction pipeline from a user-provided DrugBank 5.1.13 XML file, (iii) a toy/synthetic XML fixture for testing the pipeline without DrugBank access, (iv) integrity checks for locally reconstructed files, and (v) a Croissant metadata file with core and Responsible AI fields. DrugBank records themselves are not redistributed because of the DrugBank license. To reconstruct the full benchmark, users must obtain DrugBank 5.1.13 in XML format independently. The derivative artifacts and code are hosted in the project's GitHub repository.

\textsc{ColdDDI} is derived from DrugBank~5.1.13 (released January 2025). DrugBank's academic license explicitly prohibits data redistribution, and DrugBank support confirmed this constraint in writing in response to our inquiry.
Consequently, we do not release the underlying DrugBank records (drug descriptions, SMILES strings, DDI edge tables, or entity associations). Users reproduce the full benchmark by applying for DrugBank academic access\footnote{\url{https://go.drugbank.com/releases/5-1-13}} and running our reconstruction pipeline on their own DrugBank XML. Our own derivative artifacts (mechanism annotations, drug-wise split indices, prompt templates, and evaluation code) are not subject to DrugBank's redistribution restriction and are available in the release repository\footnote{\url{https://github.com/0217ljh/ColdDDI-NeurIPS2026}}.

\paragraph{Out-of-the-Box Reproduction.}
The repository provides three top-level entry scripts for reconstruction, validation, and conventional baseline evaluation, together with a separate P4 LLM runner. \texttt{reconstruct.py} rebuilds annotations, splits, and prompt-ready artifacts. \texttt{sanity\_check.py} checks data consistency and, with \texttt{--strict}, verifies the checksums recorded during reconstruction. \texttt{evaluate.py} runs conventional baselines, while \texttt{scripts/run\_benchmark.py} runs P4 LLM training, checkpoint selection, evaluation, and diagnostics. The end-to-end five-stage walkthrough with hardware budgets and expected outputs is in Appendix~\ref{app-A.6.2:walkthrough}, and the full path-by-path repository layout is in Appendix~\ref{app-A.6.1:repo-structure}.

\subsubsection{Repository Structure}
\label{app-A.6.1:repo-structure}

The release repository is organized as an installable Python package (\texttt{coldddi/}) plus three top-level entry scripts and a set of released data artifacts. Every directory is consumed by at least one paper section, and every section that asserts a quantitative result has a corresponding artifact directory. \cref{tab:repo-structure} summarizes the layout.

\begin{table}[h]
\centering
\caption{Release repository layout (\texttt{ColdDDI/}). Top-level entry scripts are immediately runnable. The library code lives under \texttt{coldddi/}. Paths beginning with \texttt{coldddi\_data/} are generated locally by the walkthrough below.}
\label{tab:repo-structure}
\small
\setlength{\tabcolsep}{4pt}
\renewcommand{\arraystretch}{1.05}
\resizebox{\textwidth}{!}{%
\begin{tabular}{@{}lll@{}}
\toprule
\textbf{Path} & \textbf{Contents} & \textbf{Paper anchor} \\
\midrule
\multicolumn{3}{@{}l}{\emph{Entry scripts (top level)}} \\
\texttt{reconstruct.py}              & DrugBank XML $\to$ filtered edges + KG + annotations + splits & Appendices~\ref{app-A.1:filtering-pipeline} and~\ref{app-B.1:split-stats}  \\
\texttt{evaluate.py} & Conventional baseline training and evaluation & \cref{sec:cold-start-results} \\
\texttt{sanity\_check.py} & Checks data consistency and reconstruction checksums & Appendix~\ref{app-A.6.2:walkthrough} \\
\midrule
\multicolumn{3}{@{}l}{\emph{Library package}} \\
\texttt{coldddi/data/filter.py}      & 7-step filtering pipeline                                      & Appendix~\ref{app-A.1:filtering-pipeline}          \\
\texttt{coldddi/data/splits.py}      & $G_1/G_2$ partition + S0/S1/S2 builder                         & \cref{sec:cold-start-settings} and Appendix~\ref{app-B.1:split-stats} \\
\texttt{coldddi/data/negatives.py}   & Default and ratio/strategy negative samplers                   & Appendix~\ref{app-B.2:negative-sampling}           \\
\texttt{coldddi/data/kg.py}          & DrugBank KG (enzymes, transporters, targets, carriers, pathways) loader & Appendix~\ref{app-A.3:ab-subdivision}    \\
\texttt{coldddi/annotations/pkpd\_keywords.py} & PK/PD keyword matcher                            & Appendix~\ref{app-A.2:pkpd-keywords}               \\
\texttt{coldddi/annotations/ab\_subdivision.py} & A/B shared-entity protocol                      & Appendix~\ref{app-A.3:ab-subdivision}              \\
\texttt{coldddi/baselines/base.py}   & \texttt{BaselineModel} abstract class (fit / predict / save / load) & Appendix~\ref{app-C.1:baselines}              \\
\texttt{coldddi/baselines/\{deepddi,ssi\_ddi,dsn\_ddi,hdn\_ddi,emergnn,tiger,mkg\_fenn,textddi\}/} & Eight conventional baselines & Appendix~\ref{app-C.1:baselines}              \\
\texttt{coldddi/llm/prompts/binary\_cls.py} & Renders P1--P5 templates and R0--R7 masking variants & \cref{sec:llm-patterns} and Appendix~\ref{app-D.1:prompts} \\
\texttt{coldddi/llm/\{trainer,select\_best\}.py} & LoRA training and checkpoint selection & Appendix~\ref{app-D.2:lora} \\
\texttt{coldddi/llm/inference.py}    & Greedy scoring with elicited probability                       & Appendix~\ref{app-D.2:lora}                        \\
\texttt{coldddi/diagnostics/\{kps,ksai,masking\}.py} & KPS and KSAI diagnostic interfaces     & \cref{sec:modality-results} and Appendices~\ref{app-E.1:diagnostic-metrics},~\ref{app-E.2:masking-full},~\ref{app-E.3:kps-full} \\
\texttt{coldddi/evaluate.py} & Per-run baseline predictions and evaluation metrics & \cref{sec:cold-start-results} \\
\texttt{coldddi/eval/aggregate.py} & Overall, stratified, and diagnostic summaries across runs & \cref{sec:cold-start-results} \\
\midrule
\multicolumn{3}{@{}l}{\emph{Repository artifacts and local reconstruction outputs}} \\
\texttt{annotations/pkpd.parquet}    & 215 DDI types $\to$ \{PK, PD, Mixed\}                          & Appendix~\ref{app-A.2:pkpd-keywords}               \\
\texttt{coldddi\_data/outputs\_full/annotations/ab.parquet} & Locally reconstructed A/B annotations for the full benchmark & Appendix~\ref{app-A.3:ab-subdivision} \\
\texttt{coldddi\_data/outputs\_full/annotations/mediating\_entities.parquet} & Locally reconstructed mediating-entity annotations & Appendix~\ref{app-A.3:ab-subdivision} \\
\texttt{coldddi\_data/outputs\_full/annotations/action\_pairs.parquet} & Locally reconstructed entity-role annotations & Appendix~\ref{app-A.3:ab-subdivision} \\
\texttt{coldddi\_data/intermediate/splits/seed\{42,43,44\}/} & Full-dataset Parquet splits and JSON manifests containing $G_1/G_2$ & Appendix~\ref{app-B.1:split-stats} \\
\texttt{coldddi\_data/subsets/800/seed\{42,43,44\}/intermediate/} & Per-seed 800-drug datasets and splits, generated with \texttt{--include-subset 800/both} & Appendix~\ref{app-B.3.1:subset-construction} \\
\texttt{coldddi\_data/intermediate/splits/checksums.txt} & MD5 checksums recorded during reconstruction & Appendix~\ref{app-A.6.2:walkthrough} \\
\midrule
\multicolumn{3}{@{}l}{\emph{Reproducibility helpers}} \\
\texttt{exps/sec5\_\{overall,stratified,kps\}.sh} & Conventional baseline sweeps and metric aggregation & \cref{sec:cold-start-results,sec:stratified-analysis,sec:modality-results} \\
\texttt{annotation/}                 & Human-validation package (sampling, rubric, IAA scripts, final consensus) & Appendix~\ref{app-A.4:taxonomy-validation} \\
\texttt{requirements.txt} & Dependencies for reconstruction, conventional baselines, and the LLM pipeline & Appendix~\ref{app-A.6.3:hardware-software} \\
\texttt{LICENSE}                     & MIT for code, CC BY 4.0 for annotations                     & Appendix~\ref{app-A.6:release}                     \\
\texttt{README.md}                   & Quick-start + pointer to this appendix                         & Appendix~\ref{app-A.6.2:walkthrough}         \\
\bottomrule
\end{tabular}%
}
\end{table}

\subsubsection{Reproduction Pipeline and Public APIs}
\label{app-A.6.2:walkthrough}

The library exposes three top-level entry scripts (\texttt{reconstruct.py}, \texttt{sanity\_check.py}, and \texttt{evaluate.py}), the P4 LLM runner \texttt{scripts/run\_benchmark.py}, and Python interfaces for baselines, prompts, and diagnostics. The CLI signatures are listed first, followed by the end-to-end walkthrough and the Python interfaces.

\paragraph{Entry-Script CLI.}
Each entry script uses the data-path and seed arguments listed below.

\begin{tcolorbox}[colback=gray!5, colframe=black!60, title=Entry-script CLI signatures, fonttitle=\bfseries\small, boxsep=2pt, left=4pt, right=4pt, top=2pt, bottom=2pt]
\small
\begin{verbatim}
python reconstruct.py
       --drugbank   PATH                # DrugBank 5.1.13 XML
       --output     DIR                 # output dataset directory
       --seeds      INT [INT ...]       # default: 42 43 44
       --include-subset {800,full,both} # default: full

python sanity_check.py
       --data       DIR                 # output of reconstruct.py
       [--strict]                       # also verify every per-seed split MD5

python evaluate.py
       --method NAME                   # e.g., dsn_ddi or emergnn
       (--data DIR | --subset {800,1900})
       --seed INT                      # one reconstructed seed
       --setting {S0,S1,S2,all}         # default: all
       [--ab-parquet PATH]              # matching A/B annotations
       [--adapter PATH]                 # load a saved baseline
       [--device {cuda,cpu}]
       [--out DIR]                     # predictions and metrics
python scripts/run_benchmark.py
       --data DIR                      # contains filtered/ and splits/
       --ab-parquet PATH               # matching A/B annotations
       --model MODEL                   # Hugging Face ID or local path
       [--prompt P4]
       [--seed INT]                    # default: 42
       [--device {cuda,cpu,auto}]
       [--output DIR]
\end{verbatim}
\end{tcolorbox}

\paragraph{Step-by-Step Walkthrough.}
End-to-end reproduction chains the entry scripts in five stages, after a one-time DrugBank XML download (Stage~0). Reproduction on any 32--80~GB GPU (depending on model size) should fall within $\pm 20\%$ of our internal RTX~5090 / A40 wall-clock measurements. Per-baseline budgets are in \cref{app-C.1:baselines}.

\begin{tcolorbox}[colback=gray!5, colframe=black!60, title=Reproduction Walkthrough, fonttitle=\bfseries\small, boxsep=2pt, left=4pt, right=4pt, top=2pt, bottom=2pt]
\small
\begin{verbatim}
# Stage 0: Obtain DrugBank XML (one-time, ~1-3 days approval lead time).
#   Apply at https://go.drugbank.com/releases/5-1-13
#   Download drugbank_5.1.13.xml (~1.5 GB compressed).

# Stage 1: Clone the code repository and install.
git clone https://github.com/0217ljh/ColdDDI-NeurIPS2026.git
cd ColdDDI-NeurIPS2026
python -m venv .venv && source .venv/bin/activate    # or conda env
python -m pip install -r requirements.txt

# Stage 2: Reconstruct (deterministic, ~15 min on a recent x86 desktop, CPU-bound).
python reconstruct.py \
       --drugbank /path/to/drugbank_5.1.13.xml \
       --output   ./coldddi_data \
       --seeds    42 43 44

# Stage 3: Sanity-check (~30 s, verifies data consistency
#                         and checksums recorded during reconstruction).
python sanity_check.py --data ./coldddi_data --strict

# Stage 4: Run a baseline end-to-end (~30-90 min per baseline, GPU).
python evaluate.py \
       --method dsn_ddi \
       --seed 42 \
       --data ./coldddi_data/intermediate \
       --ab-parquet ./coldddi_data/outputs_full/annotations/ab.parquet
python evaluate.py \
       --method emergnn \
       --seed 42 \
       --data ./coldddi_data/intermediate \
       --ab-parquet ./coldddi_data/outputs_full/annotations/ab.parquet

# Stage 5 (optional): LoRA fine-tune Llama-3.2-1B under P4 (~6 GPU-hours).
python scripts/run_benchmark.py \
       --data ./coldddi_data/intermediate \
       --ab-parquet ./coldddi_data/outputs_full/annotations/ab.parquet \
       --model meta-llama/Llama-3.2-1B \
       --prompt P4 \
       --seed 42 \
       --device cuda
\end{verbatim}
\end{tcolorbox}

\noindent Cross-CUDA-version drift is below $\pm 0.005$ on each metric in our internal tests.

\paragraph{Baseline Plugin Contract.}
Every method in \texttt{coldddi/baselines/} subclasses \texttt{BaselineModel} (\texttt{coldddi/baselines/base.py}), an abstract class with four methods. New baselines implement this class, register their method name with the \texttt{register} decorator, and add their module to \texttt{NAME\_TO\_MODULE} in \texttt{coldddi/baselines/base.py}.

\begin{tcolorbox}[colback=gray!5, colframe=black!60, title=BaselineModel ABC, fonttitle=\bfseries\small, boxsep=2pt, left=4pt, right=4pt, top=2pt, bottom=2pt]
\small
\begin{verbatim}
class BaselineModel(ABC):
    name: str                                # registered method alias
    def fit(self, train: PairDataset,
                  val:   PairDataset) -> None: ...
    def predict_proba(self, pairs: pd.DataFrame) -> np.ndarray: ...
    def save(self, path: str) -> None: ...
    @classmethod
    def load(cls, path: str) -> "BaselineModel": ...
\end{verbatim}
\end{tcolorbox}

\noindent\texttt{PairDataset} wraps the drug-pair indices, the per-epoch negative pool, and a handle to the KG / SMILES tables, so a new baseline only consumes the abstractions and never touches raw DrugBank.

\paragraph{Prompt-Template Plugin.}
LLM prompt patterns and masking variants are implemented in \texttt{coldddi/llm/prompts/binary\_cls.py}. New patterns are added by extending the Python prompt builder and the corresponding runner mapping. The release does not use external Jinja templates or a YAML masking registry.

\paragraph{Diagnostic-Indicator Plugin.}
\texttt{coldddi/diagnostics/} exposes \texttt{compute\_kps\_f}, \texttt{compute\_kps\_channel}, and \texttt{compute\_ksai}. KPS-F takes base predictions and swap candidates; channel KPS additionally takes masked predictions and a channel name; KSAI takes predictions under R0--R3. The functions return per-bucket \texttt{pandas.DataFrame} results. New indicators require an explicit implementation and integration into the evaluation pipeline.

\subsubsection{Hardware and Software Requirements}
\label{app-A.6.3:hardware-software}

\paragraph{Software.}
Python~3.10 with the dependencies specified in \texttt{requirements.txt}. Later Python versions trigger a known \texttt{torch\_scatter} ABI mismatch in EmerGNN and are not supported.

\paragraph{Hardware.}
Reconstruction is CPU-bound and runs in $\sim$15~min on a 16~GB-RAM desktop. Baseline training and LoRA fine-tuning fit on a single 32--80~GB GPU depending on model size. Per-method wall-clock and memory budgets are in \cref{tab:baseline-hparams} and Appendix~\ref{app-D.2:lora}. Disk usage is $\sim$2.0~GB for the reconstructed dataset directory and $\sim$15--40~MB per LLM adapter.

\subsubsection{Licensing, Maintenance, and Availability}
\label{app-A.6.4:license}

\paragraph{Licensing.}
Code is released under the MIT License. Mechanism annotations and split indices are released under CC BY 4.0. The underlying DrugBank records are not redistributed and remain governed by the DrugBank academic license.

\paragraph{Maintenance.}
The benchmark is maintained by the authors. Planned annual updates align with DrugBank releases (next target DrugBank~5.1.14). Issues, bug reports, and baseline contributions are accepted through the GitHub repository. 

\paragraph{Perpetual Availability.}
The GitHub repository provides the derivative artifacts and reconstruction pipeline. Because the derivatives are self-contained and all released artifacts are pinned to DrugBank~5.1.13, the benchmark remains reconstructible for any user who can independently obtain DrugBank academic access.

The GitHub repository provides the derivative artifacts and reconstruction pipeline. Users must obtain DrugBank~5.1.13 independently to reconstruct the full benchmark.

\section{Evaluation Splits and Sampling}
\label{app-B:splits}

\subsection{Split Construction and Statistics}
\label{app-B.1:split-stats}

This subsection details the construction protocol for the cold-start splits introduced in \cref{sec:cold-start-settings} and reports the resulting train / validation / test edge counts.

\paragraph{Drug-Wise Partition.}
The full benchmark contains 1{,}900 drugs after the filtering pipeline of \cref{app-A.1:filtering-pipeline}. We partition the drug set into $G_1$ (seen drugs) and $G_2$ (unseen drugs) by uniform random sampling under a fixed seed. The split ratio is $|G_1| : |G_2| = 80\% : 20\%$, which yields 1{,}520 drugs in $G_1$ and 380 drugs in $G_2$ on the full 1{,}900-drug benchmark, and 640 / 160 on the 800-drug subset (\cref{app-B.3.1:subset-construction}).

\paragraph{Edge Pool Decomposition.}
The positive DDI edges are partitioned into three pools by the partition status of their endpoint drugs, extending the $\mathcal{E}_{\text{all}}^{(G_1)}$ notation of \cref{sec:cold-start-settings}.
\begin{itemize}[leftmargin=*,itemsep=1pt,topsep=1pt,parsep=0pt]
    \item $\mathcal{E}_{\text{all}}^{(G_1)}$, both endpoints in $G_1$, used for transductive evaluation (S0).
    \item $\mathcal{E}_{\text{all}}^{(G_1, G_2)}$, exactly one endpoint in $G_2$, used for semi-inductive evaluation (S1).
    \item $\mathcal{E}_{\text{all}}^{(G_2)}$, both endpoints in $G_2$, used for fully-inductive evaluation (S2).
\end{itemize}
The exact pool sizes are seed-dependent because the $G_1$/$G_2$ partition is sampled per seed. \cref{tab:split-counts} reports per-seed pool sizes and per-split positive edge counts on both scales.

\paragraph{Train / Val / Test Allocation.}
Within each edge pool, edges are allocated to splits as follows. The $\mathcal{E}_{\text{all}}^{(G_1)}$ pool follows a 90/5/5 split (train / val-S0 / test-S0). The two cross-pools $\mathcal{E}_{\text{all}}^{(G_1, G_2)}$ and $\mathcal{E}_{\text{all}}^{(G_2)}$ use a 50/50 split (val / test) and contribute zero edges to the training set, ensuring that test-S1 and test-S2 evaluation is performed exclusively on edges involving at least one unseen drug.

\begin{table}[h]
\centering
\caption{Per-seed positive edge counts on the full 1{,}900-drug benchmark and the 800-drug subset across three random seeds. Drug-set sizes are $|G_1|/|G_2| = 1{,}520/380$ on the full set and $640/160$ on the subset for every seed. Negative edges are sampled at a 1:1 ratio per split (\cref{app-B.2:negative-sampling}) and have identical counts.}
\label{tab:split-counts}
\small
\setlength{\tabcolsep}{4pt}
\renewcommand{\arraystretch}{1.1}
\begin{tabular}{@{}l rrr rrr@{}}
\toprule
                                           & \multicolumn{3}{c}{\textbf{Full 1{,}900-drug}} & \multicolumn{3}{c}{\textbf{800-drug subset}} \\
\cmidrule(lr){2-4}\cmidrule(lr){5-7}
\textbf{Quantity}                          & \textbf{Seed 42} & \textbf{Seed 43} & \textbf{Seed 44} & \textbf{Seed 42} & \textbf{Seed 43} & \textbf{Seed 44} \\
\midrule
\multicolumn{7}{@{}l}{\emph{Edge pool sizes}} \\
\quad $\mathcal{E}_{\text{all}}^{(G_1)}$       & 361{,}485 & 366{,}641 & 360{,}330 & 59{,}715 & 62{,}861 & 61{,}554 \\
\quad $\mathcal{E}_{\text{all}}^{(G_1, G_2)}$  & 181{,}571 & 177{,}769 & 182{,}380 & 30{,}515 & 29{,}867 & 32{,}448 \\
\quad $\mathcal{E}_{\text{all}}^{(G_2)}$       &  22{,}675 &  21{,}321 &  23{,}021 &  3{,}837 &  3{,}450 &  4{,}188 \\
\midrule
\multicolumn{7}{@{}l}{\emph{Per-split positive counts}} \\
\quad Train                                & 325{,}336 & 329{,}976 & 324{,}297 & 53{,}743 & 56{,}574 & 55{,}398 \\
\quad Val-S0                               &  18{,}074 &  18{,}332 &  18{,}016 &  2{,}986 &  3{,}143 &  3{,}078 \\
\quad Test-S0                              &  18{,}075 &  18{,}333 &  18{,}017 &  2{,}986 &  3{,}144 &  3{,}078 \\
\quad Val-S1                               &  90{,}785 &  88{,}884 &  91{,}190 & 15{,}257 & 14{,}933 & 16{,}224 \\
\quad Test-S1                              &  90{,}786 &  88{,}885 &  91{,}190 & 15{,}258 & 14{,}934 & 16{,}224 \\
\quad Val-S2                               &  11{,}337 &  10{,}660 &  11{,}510 &  1{,}918 &  1{,}725 &  2{,}094 \\
\quad Test-S2                              &  11{,}338 &  10{,}661 &  11{,}511 &  1{,}919 &  1{,}725 &  2{,}094 \\
\midrule
\textbf{Total positive edges}              & \textbf{565{,}731} & \textbf{565{,}731} & \textbf{565{,}731} & \textbf{94{,}067} & \textbf{96{,}178} & \textbf{98{,}190} \\
\bottomrule
\end{tabular}
\end{table}

\subsubsection{Structural Similarity Across Drug-Wise Splits}
\label{app-B.1.1:structural-overlap}

We examine structural overlap between $G_1$ and $G_2$ across the three full-set partitions generated with seeds 42, 43, and 44.
All 1{,}900 drugs have distinct canonical SMILES, so no exact SMILES duplicates occur across the drug-wise splits.
For each held-out drug in $G_2$, the maximum Tanimoto similarity to any drug in $G_1$ is computed using Morgan fingerprints with radius 2 and 2{,}048 bits.
The within-train baseline compares each drug in $G_1$ with the remaining drugs in $G_1$, excluding self-comparisons.

\begin{table}[h]
\centering
\caption{Percentage of drugs whose maximum Tanimoto similarity to the reference set exceeds each cutoff (3-seed mean$\pm$std on the full 1{,}900-drug benchmark). Train--Test compares held-out drugs against training drugs; Train--Train uses the remaining training drugs as the reference.}
\label{tab:structural-overlap}
\small
\setlength{\tabcolsep}{6pt}
\renewcommand{\arraystretch}{1.05}
\begin{tabular}{@{}lcc@{}}
\toprule
\textbf{Tanimoto cutoff}
& \textbf{Train--Test}
& \textbf{Train--Train} \\
\midrule
$>0.85$ & $5.18 \pm 0.45$ & $5.07 \pm 0.19$ \\
$>0.90$ & $3.77 \pm 0.12$ & $3.88 \pm 0.05$ \\
\bottomrule
\end{tabular}
\end{table}

At both cutoffs, high-similarity neighbors occur at comparable rates across and within the training partition (\cref{tab:structural-overlap}).
The observed train--test similarity therefore reflects structural similarity already present in the drug pool, with no marked excess introduced by the random split.

\subsection{Negative Sampling Strategy}
\label{app-B.2:negative-sampling}

A DDI benchmark derived from curated databases is unavoidably open-world: the absence of a recorded interaction does not guarantee that no pharmacological interaction exists.
This subsection documents (i)~the default sampling procedure used throughout the main paper, (ii)~the per-epoch re-sampling protocol, two robustness analyses quantifying the effects of (iii)~positive-to-negative ratio and (iv)~negative-sampling strategy, and (v)~temporal backfill across historical datasets.
Both robustness analyses follow a \emph{train-only} design that rewrites only the training-side negatives while leaving validation and test negatives identical to the default fair bundle. Every (method, configuration) cell is therefore evaluated on the same fixed 1:1 test set, so AUC differences attribute cleanly to training-sample quality rather than test-difficulty drift. The ratio and sampling-strategy analyses are conducted on the 800-drug subset under a single random seed (48 train runs total under the configurations enumerated below). TextDDI is omitted from the robustness analyses due to its prohibitive training time.

\subsubsection{Default Sampling Procedure}
\label{app-B.2.1:neg-default}

Let $\mathcal{E}_{\text{all}}$ denote the set of positive DDI pairs and $\overline{\mathcal{E}}^{(G_1)}$, $\overline{\mathcal{E}}^{(G_1, G_2)}$, $\overline{\mathcal{E}}^{(G_2)}$ the setting-specific complement spaces defined in \cref{sec:cold-start-settings} (for S0, S1, S2 respectively). For each setting we sample negative pairs uniformly at random from the corresponding complement space at a $1{:}1$ positive-to-negative ratio. Sampling is performed without replacement within a single draw and restricted to pairs not appearing in $\mathcal{E}_{\text{all}}$ under any setting.

\subsubsection{Per-Epoch Re-Sampling Protocol}
\label{app-B.2.2:neg-resampling}

For \emph{training}, a fresh negative set $\mathcal{N}_{\text{train}}^{(t)} \sim \overline{\mathcal{E}}^{(G_1)}$ is sampled at the start of every epoch $t$ (in practice rotated across four pre-sampled sets to match LLM-FT's finite negative budget, see Appendix~\ref{app-C.2:training-protocol}).
This dynamic re-sampling (i)~reduces overfitting to any particular draw of presumed negatives, and (ii)~mimics the open-world variability that a deployed model encounters.
For \emph{validation and test}, negatives are sampled \emph{once} using a fixed seed and held constant across all methods and across training, ensuring identical evaluation conditions.

\subsubsection{Ratio Robustness Analysis}
\label{app-B.2.3:neg-ratio}

We evaluate seven conventional baselines and the best LLM-FT configuration (Llama-3.2-1B, P4) on S2 under three training-side positive-to-negative ratios $r \in \{1, 3, 5\}$.
For each ratio, the training negatives are sampled uniformly from $\overline{\mathcal{E}}^{(G_1)}$ at the target multiplicity per epoch (per-epoch re-sampling, \cref{app-B.2.2:neg-resampling}).
\cref{tab:neg-ratio-robustness} reports the resulting S2 AUC.

\begin{table}[h]
\centering
\caption{S2 AUC under three training-side positive-to-negative ratios (800-drug subset, single random seed). The test set is held fixed at 1:1 across all cells. The $r{=}1$ column reuses the default fair bundle.}
\label{tab:neg-ratio-robustness}
\centering
\small
\setlength{\tabcolsep}{6pt}
\renewcommand{\arraystretch}{1.05}
\begin{tabular}{lccc}
\toprule
\textbf{Method} & \textbf{1:1} & \textbf{1:3} & \textbf{1:5} \\ 
\midrule
DeepDDI   & 0.678 & 0.680 & 0.674 \\ 
SSI-DDI   & 0.656 & 0.646 & 0.642 \\ 
DSN-DDI   & 0.814 & 0.755 & 0.793 \\ 
HDN-DDI   & 0.675 & 0.668 & 0.667 \\ 
EmerGNN   & 0.724 & 0.714 & 0.706 \\ 
TIGER     & 0.581 & 0.589 & 0.557 \\ 
MKG-FENN  & 0.569 & 0.607 & 0.568 \\ 
LLM-FT    & 0.799 & 0.788 & 0.755 \\ 
\bottomrule
\end{tabular}
\end{table}

Across all cells, S2 AUC stays within $\pm 0.04$ of the $r{=}1$ default for most methods (DeepDDI, SSI-DDI, HDN-DDI, EmerGNN, TIGER, MKG-FENN), indicating that the headline ranking is largely insensitive to the train-side ratio.
DSN-DDI is the most ratio-sensitive method, with AUC oscillating between $0.755$ and $0.814$. LLM-FT shows a small monotone drop from $r{=}1$ ($0.799$) to $r{=}5$ ($0.755$). The two top methods at $r{=}1$ (LLM-FT $0.799$, DSN-DDI $0.814$) remain the top two in every ratio column.

\subsubsection{Sampling Strategy Comparison}
\label{app-B.2.4:neg-strategy}

We compare three training-side negative-sampling strategies on S2 using seven conventional baselines and the best LLM-FT configuration (Llama-3.2-1B, P4). 
Pair-pair similarity uses Morgan fingerprints (radius 2, 2048 bits) and the symmetric pair-similarity formula $\mathrm{sim}((a, b), (c, d)) = \max\bigl(\min(T_{a,c}, T_{b,d}),\ \min(T_{a,d}, T_{b,c})\bigr)$, where $T$ is Tanimoto similarity.

\begin{itemize}[leftmargin=*,itemsep=1pt,topsep=1pt,parsep=0pt]
    \item \textbf{Uniform random (default)}, sampled uniformly from $\overline{\mathcal{E}}^{(G_1)}$. Reuses the default fair training bundle.
    \item \textbf{Hard negative}, rank-based selection from $\overline{\mathcal{E}}^{(G_1)}$. We score each candidate by $s(n) = \max_{p \in \mathcal{E}_{\text{train}}} \mathrm{sim}(n, p)$ against all training-side positive pairs, retain the top 10\% of the candidate pool, and then sample at the target ratio. Rank-based selection is required because, for ColdDDI's drug-pair similarity distribution, even the median positive-pair similarity admits nearly the entire candidate pool.
    \item \textbf{Structure-matched}, stratified sampling over 10 Tanimoto-similarity bins, matching the candidate-score histogram to the positive-pair similarity histogram. Empty bins are redistributed iteratively to non-empty neighbors (weighted by target frequency), removing easy-negative bias without inducing a single-bin collapse.
\end{itemize}

\cref{tab:neg-strategy} reports the resulting S2 AUC.

\begin{table}[h]
\centering
\caption{S2 AUC under three training-side negative-sampling strategies (800-drug subset, single random seed). The Random column reuses the default fair bundle.}
\label{tab:neg-strategy}
\centering
\small
\setlength{\tabcolsep}{6pt}
\renewcommand{\arraystretch}{1.05}
\begin{tabular}{lccc}
\toprule
\textbf{Method} & \textbf{Random} & \textbf{Hard} & \textbf{Structure-matched} \\
\midrule
DeepDDI   & 0.678 & 0.558 & 0.672 \\
SSI-DDI   & 0.656 & 0.581 & 0.645 \\
DSN-DDI   & 0.814 & 0.718 & 0.770 \\
HDN-DDI   & 0.675 & 0.623 & 0.621 \\
EmerGNN   & 0.724 & 0.619 & 0.723 \\
TIGER     & 0.581 & 0.572 & 0.569 \\
MKG-FENN  & 0.569 & 0.473 & 0.618 \\
LLM-FT    & 0.799 & 0.789 & 0.789 \\
\bottomrule
\end{tabular}
\end{table}

Hard-negative training depresses AUC by roughly $0.10$ for four baselines (DeepDDI $0.120$, EmerGNN $0.105$, DSN-DDI $0.096$, MKG-FENN $0.096$) and by less than $0.08$ for the rest (SSI-DDI, HDN-DDI, TIGER, LLM-FT), confirming that the hard-negative pool is genuinely harder than uniform random.
Structure-matched training stays within $\pm 0.06$ of the random baseline for every method, with MKG-FENN ($0.618$ vs.\ $0.569$) the only noticeable improvement.
LLM-FT is the most strategy-robust ($0.799$ random, $0.789$ hard, $0.789$ structure-matched, all within $\pm 0.01$). LLM-FT and DSN-DDI remain top-two across all three sampling strategies.

\subsubsection{Temporal Backfill Analysis}
\label{app-B.2.5:temporal-backfill}

We examine how often unrecorded pairs become documented interactions in later DrugBank-derived datasets.
The released RYU, DENG, and KGNN datasets correspond to DrugBank versions 5.0.3, 5.1.3, and 5.1.4, respectively.
Within each shared-drug pool, the backfill rate is the fraction of pairs unrecorded in the earlier dataset that appear as positive interactions in the later one.
The backfill rates are $0.47\%$ for $5.0.3 \to 5.1.3$ (548 shared drugs), $2.58\%$ for $5.0.3 \to 5.1.4$ (990 shared drugs), and $3.98\%$ for $5.1.3 \to 5.1.4$ (387 shared drugs).
These modest backfill rates indicate that identifiable contamination affects only a small fraction of previously unrecorded pairs in the three version comparisons.

\subsection{Subset Representativeness and Stability}
\label{app-B.3:subset-stability}

The full 1{,}900-drug ColdDDI dataset is the primary analysis target, and all conventional baseline evaluations are run on it directly. 
LLM fine-tuning and the related diagnostic analyses are instead run on a stratified 800-drug subset, because their compute cost scales super-linearly with drug count and full-set evaluation across the LLM matrix is intractable under our budget.
To justify this scope reduction, this subsection shows that the 800-drug subset closely tracks the full benchmark along three complementary angles: (i)~cross-seed preservation of the mechanism-stratified distribution (Appendix~\ref{app-B.3.1:subset-construction}), (ii)~head-to-head comparison of conventional baselines between the full 1{,}900-drug set and the 800-drug subset (Appendix~\ref{app-B.3.2:subset-vs-full}), and (iii)~cross-size stability of model performance (Appendix~\ref{app-B.3.3:subset-size}).

\subsubsection{Subset Construction and Cross-Seed Distributional Stability}
\label{app-B.3.1:subset-construction}

The 800-drug subset is sampled uniformly without replacement from the 1{,}900-drug pool under a random seed, then partitioned and split following the same protocol as the full set (Appendix~\ref{app-B.1:split-stats}).
Main-paper results and the distributional verification below both use three random seeds (\cref{tab:subset-overview}).

\begin{table}[h]
\centering
\caption{Full dataset and 800-drug subsets across the three seeds.}
\label{tab:subset-overview}
\small
\setlength{\tabcolsep}{6pt}
\renewcommand{\arraystretch}{1.05}
\begin{tabular}{lrrrr}
\toprule
\textbf{Statistic} & \textbf{Full (1,900)} & \textbf{Seed 42} & \textbf{Seed 43} & \textbf{Seed 44} \\
\midrule
Total drugs              & 1,900   & 800     & 800     & 800     \\
$G_1$ (seen)             & 1,520   & 640     & 640     & 640     \\
$G_2$ (unseen)           & 380     & 160     & 160     & 160     \\
Total positive DDI pairs & 565,731 & 94,067  & 96,178  & 98,190  \\
DDI types covered        & 215     & 165     & 152     & 165     \\
Train positive           & 325,336 & 53,743  & 56,574  & 55,398  \\
Test S0 positive         & 18,075  & 2,986   & 3,144   & 3,078   \\
Test S1 positive         & 90,786  & 15,258  & 14,934  & 16,224  \\
Test S2 positive         & 11,338  & 1,919   & 1,725   & 2,094   \\
Test S2 DDI types        & 215     & 73      & 59      & 66      \\
\bottomrule
\end{tabular}
\end{table}

Each subset retains 152--165 of 215 DDI types (71--77\%), covering more than 99\% of all DDI edges by type frequency. 
The missing types are all low-frequency PD types (median count 18 edges in the full dataset), whose absence does not affect the PK/PD distributional conclusions.

\begin{figure}[h]
\centering
\includegraphics[width=0.7\textwidth]{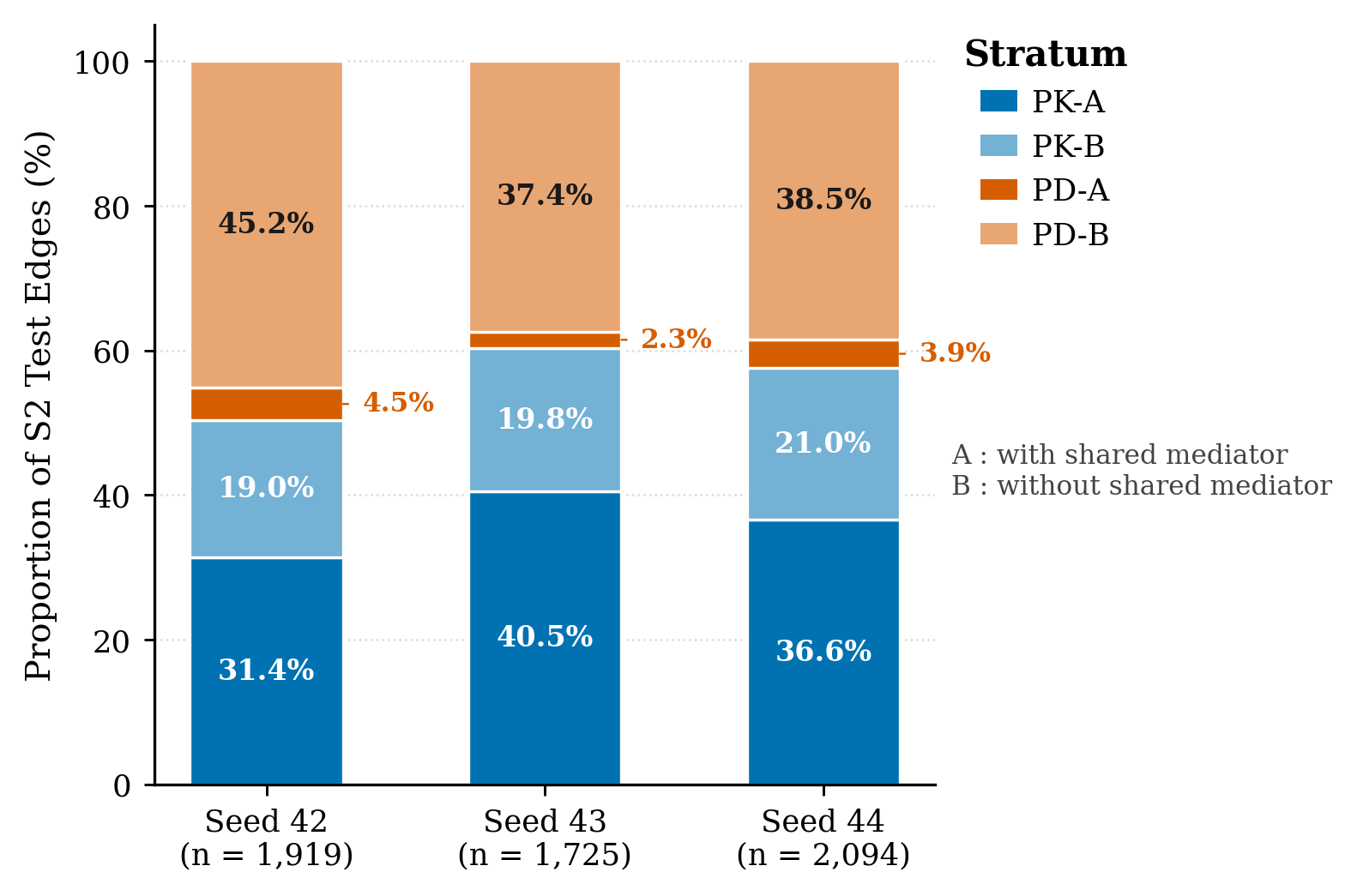}
\caption{PK/PD $\times$ A/B proportion in Test~S2 across the three seeds on the 800-drug subset. Within the PD class, PD-B dominates over PD-A in every seed, confirming that the mechanism-level composition is stable across the three random seeds.}
\label{fig:seed-pkpd}
\end{figure}

\cref{fig:seed-pkpd} shows the PK/PD $\times$ A/B distribution of Test~S2 positive edges across the three seeds.
Despite the random variation in drug composition, the broad composition is consistent across seeds.
Within the PD class, PD-B dominates over PD-A in every seed (PD-B 37.4\%--45.2\% vs.\ PD-A 2.3\%--4.5\%), and PD-A is consistently the smallest stratum overall.
The full 1{,}900-drug dataset shows the same dominance pattern (PD-B 43.9\%, PK-A 29.6\%, PK-B 22.6\%, PD-A 3.8\%, see \cref{sec:taxonomy}), confirming that the subset preserves the mechanism-level composition of the full benchmark.
The main source of cross-seed variation is the PK-A proportion (31.4\%--40.5\%), driven by whether CYP-substrate hub drugs fall into $G_2$. 
This variation does not affect the qualitative conclusions, since the A/B gap and PK/PD asymmetry are reproduced under all three seeds.

\paragraph{Cross-Seed Performance Stability.}
Beyond distributional stability, we report per-method S2 performance over three random seeds on the 800-drug subset (\cref{tab:subset-baseline-stability}).
AUC standard deviation stays under $0.06$ for every method, indicating that subset-induced variation in test composition does not cause large swings in absolute performance.
LLM-FT (Llama-3.2-1B, P4) shows comparable cross-seed stability to the conventional baselines (AUC std $0.022$).

\begin{table}[h]
\centering
\caption{Per-method S2 performance on the 800-drug subset over three random seeds, reported as mean $\pm$ standard deviation.}
\label{tab:subset-baseline-stability}
\small
\setlength{\tabcolsep}{6pt}
\renewcommand{\arraystretch}{1.1}
\begin{tabular}{@{}l ccc@{}}
\toprule
\textbf{Method} & \textbf{AUC} & \textbf{Recall} & \textbf{F1} \\
\midrule
DeepDDI                   & $0.659 \pm 0.018$ & $0.486 \pm 0.030$ & $0.555 \pm 0.013$ \\
SSI-DDI                   & $0.614 \pm 0.037$ & $0.513 \pm 0.116$ & $0.548 \pm 0.079$ \\
DSN-DDI                   & $0.758 \pm 0.052$ & $0.524 \pm 0.006$ & $0.617 \pm 0.019$ \\
HDN-DDI                   & $0.634 \pm 0.035$ & $0.633 \pm 0.079$ & $0.606 \pm 0.035$ \\
EmerGNN                   & $0.716 \pm 0.020$ & $0.545 \pm 0.060$ & $0.614 \pm 0.039$ \\
TIGER                     & $0.566 \pm 0.014$ & $0.473 \pm 0.047$ & $0.511 \pm 0.022$ \\
MKG-FENN                  & $0.566 \pm 0.009$ & $0.989 \pm 0.010$ & $0.667 \pm 0.000$ \\
TextDDI                   & $0.675 \pm 0.024$ & $0.590 \pm 0.059$ & $0.614 \pm 0.034$ \\
\midrule
\rowcolor{gray!20}
LLM-FT (Llama-3.2-1B, P4) & $0.776 \pm 0.020$ & $0.697 \pm 0.021$ & $0.705 \pm 0.018$ \\
\bottomrule
\end{tabular}
\end{table}

\subsubsection{Baseline Comparison: Full 1{,}900 vs 800 Subset}
\label{app-B.3.2:subset-vs-full}


To verify that the 800-drug subset is a faithful proxy for the full 1{,}900-drug benchmark, we compare each baseline's S0 / S1 / S2 AUC under both scales on a single random seed.
\cref{tab:subset-full-delta} reports the per-method absolute AUC difference $|\Delta\mathrm{AUC}|$ across the three settings.

\begin{table}[h]
\centering
\caption{Per-baseline $|\Delta\mathrm{AUC}|$ between the full 1{,}900-drug benchmark and the 800-drug subset on S0 / S1 / S2 (single random seed). The rightmost column is the mean across the three settings.}
\label{tab:subset-full-delta}
\small
\setlength{\tabcolsep}{6pt}
\renewcommand{\arraystretch}{1.1}
\begin{tabular}{@{}l cccc@{}}
\toprule
\textbf{Method} & \textbf{S0} & \textbf{S1} & \textbf{S2} & \textbf{Mean} \\
\midrule
DeepDDI  & 0.011 & 0.005 & 0.005 & 0.007 \\
SSI-DDI  & 0.033 & 0.021 & 0.031 & 0.028 \\
DSN-DDI  & 0.081 & 0.141 & 0.190 & 0.137 \\
HDN-DDI  & 0.041 & 0.007 & 0.032 & 0.026 \\
EmerGNN  & 0.001 & 0.005 & 0.009 & 0.005 \\
TIGER    & 0.007 & 0.047 & 0.053 & 0.036 \\
MKG-FENN & 0.002 & 0.017 & 0.016 & 0.011 \\
TextDDI  & 0.008 & 0.002 & 0.034 & 0.015 \\
\midrule
mean (8 methods) & 0.023 & 0.031 & 0.046 & 0.033 \\
\bottomrule
\end{tabular}
\end{table}

The mean $|\Delta\mathrm{AUC}|$ across all eight baselines is $0.033$ over the three settings and $0.046$ on S2, consistent with the subset-induced spread reported in \cref{sec:cold-start-results}.
Seven of the eight baselines remain within $|\Delta\mathrm{AUC}| \le 0.06$ on every setting. DSN-DDI is the only outlier ($|\Delta\mathrm{AUC}| = 0.190$ on S2), reflecting its known sensitivity to drug-pair composition under fully-inductive evaluation.
%


\subsubsection{Cross-Size Performance Stability}
\label{app-B.3.3:subset-size}

\begin{figure}[h]
\centering
\includegraphics[width=\textwidth]{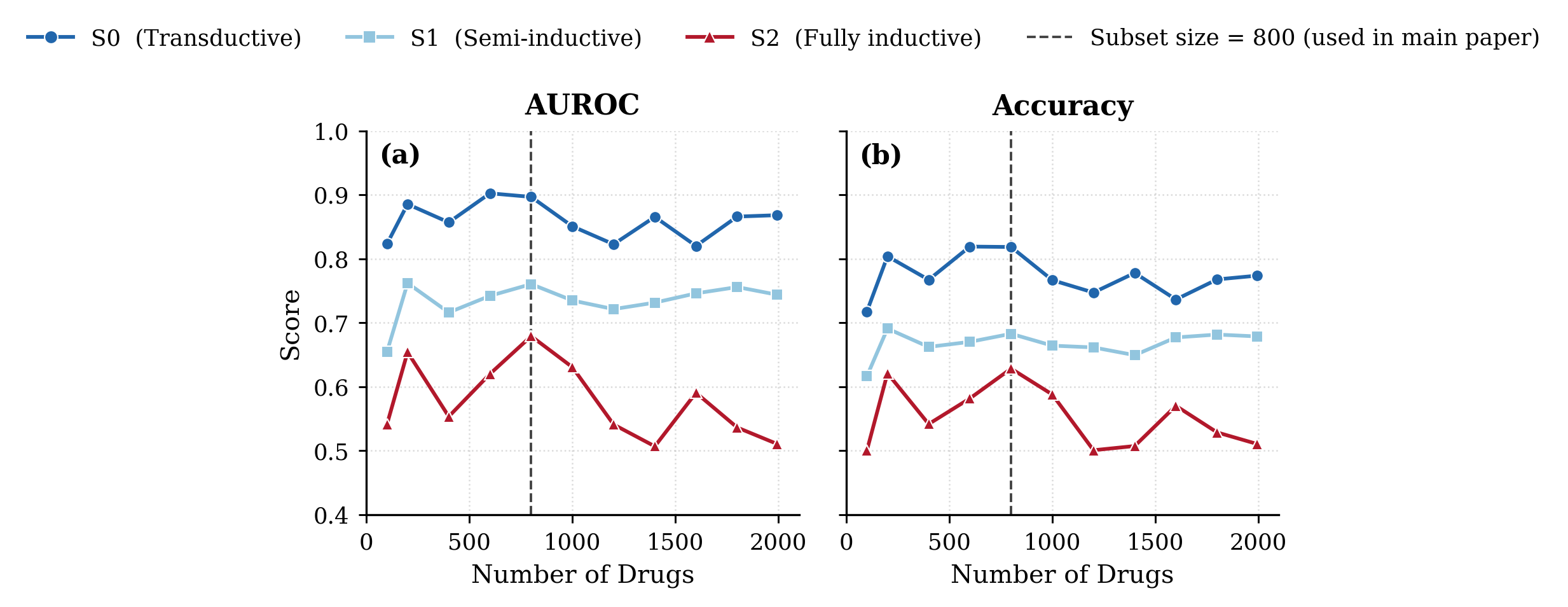}
\caption{AUROC and Accuracy across subset sizes (100--1,900 drugs) under the three cold-start settings (S0 / S1 / S2). All runs use LoRA-fine-tuned Qwen2.5-0.5B with prompt template P1. The dashed vertical line marks the 800-drug subset used in the main paper.}
\label{fig:subset-size}
\end{figure}

We scan 11 subset sizes from 100 to 1,900 drugs (full dataset) on Qwen2.5-0.5B under the P1 (Zero-Shot) prompt to characterize how performance depends on subset size.
\cref{fig:subset-size} plots AUROC and Accuracy across S0/S1/S2 as a function of subset size.
\cref{tab:subset-size-metrics} reports the complete numerical results.

\begin{table}[h]
\centering
\caption{Full metrics across subset sizes and evaluation settings (Qwen2.5-0.5B, LoRA fine-tuned).}
\label{tab:subset-size-metrics}
\resizebox{\textwidth}{!}{%
\begin{tabular}{rlrrrrrrr}
\toprule
\textbf{Drugs} & \textbf{Split} & \textbf{AUROC} & \textbf{AUPRC} & \textbf{Recall} & \textbf{Accuracy} & \textbf{Precision} & \textbf{F1} & \textbf{$n_\text{pos}$} \\
\midrule
100  & S0 & 0.823 & 0.817 & 0.630 & 0.717 & 0.763 & 0.690 & 46 \\
100  & S1 & 0.654 & 0.620 & 0.568 & 0.616 & 0.628 & 0.597 & 241 \\
100  & S2 & 0.541 & 0.542 & 0.138 & 0.500 & 0.500 & 0.216 & 29 \\
\midrule
200  & S0 & 0.885 & 0.868 & 0.801 & 0.804 & 0.805 & 0.803 & 186 \\
200  & S1 & 0.762 & 0.735 & 0.752 & 0.691 & 0.670 & 0.709 & 975 \\
200  & S2 & 0.654 & 0.662 & 0.702 & 0.621 & 0.604 & 0.649 & 124 \\
\midrule
400  & S0 & 0.857 & 0.853 & 0.793 & 0.767 & 0.754 & 0.773 & 792 \\
400  & S1 & 0.716 & 0.678 & 0.733 & 0.662 & 0.642 & 0.684 & 4,075 \\
400  & S2 & 0.553 & 0.549 & 0.619 & 0.541 & 0.536 & 0.575 & 494 \\
\midrule
600  & S0 & 0.902 & 0.906 & 0.817 & 0.819 & 0.820 & 0.819 & 1,845 \\
600  & S1 & 0.742 & 0.718 & 0.669 & 0.670 & 0.671 & 0.670 & 8,781 \\
600  & S2 & 0.620 & 0.628 & 0.780 & 0.582 & 0.558 & 0.651 & 1,042 \\
\midrule
\textbf{800}  & \textbf{S0} & \textbf{0.897} & \textbf{0.903} & \textbf{0.805} & \textbf{0.818} & \textbf{0.827} & \textbf{0.816} & \textbf{2,986} \\
\textbf{800}  & \textbf{S1} & \textbf{0.760} & \textbf{0.745} & \textbf{0.645} & \textbf{0.683} & \textbf{0.698} & \textbf{0.670} & \textbf{15,258} \\
\textbf{800}  & \textbf{S2} & \textbf{0.679} & \textbf{0.662} & \textbf{0.650} & \textbf{0.628} & \textbf{0.623} & \textbf{0.636} & \textbf{1,919} \\
\midrule
1,000  & S0 & 0.851 & 0.726 & 0.822 & 0.766 & 0.740 & 0.779 & 5,012 \\
1,000  & S1 & 0.735 & 0.694 & 0.763 & 0.664 & 0.637 & 0.694 & 23,883 \\
1,000  & S2 & 0.630 & 0.631 & 0.423 & 0.588 & 0.631 & 0.506 & 2,918 \\
\midrule
1,200  & S0 & 0.823 & 0.678 & 0.799 & 0.747 & 0.724 & 0.759 & 7,485 \\
1,200  & S1 & 0.721 & 0.689 & 0.738 & 0.661 & 0.640 & 0.685 & 35,761 \\
1,200  & S2 & 0.541 & 0.665 & 0.006 & 0.501 & 0.558 & 0.011 & 4,242 \\
\midrule
1,400  & S0 & 0.865 & 0.691 & 0.691 & 0.778 & 0.836 & 0.757 & 9,747 \\
1,400  & S1 & 0.731 & 0.692 & 0.484 & 0.649 & 0.723 & 0.580 & 48,191 \\
1,400  & S2 & 0.507 & 0.501 & 0.805 & 0.507 & 0.505 & 0.620 & 5,912 \\
\midrule
1,600  & S0 & 0.820 & 0.701 & 0.817 & 0.736 & 0.703 & 0.756 & 12,541 \\
1,600  & S1 & 0.746 & 0.710 & 0.753 & 0.677 & 0.654 & 0.700 & 65,816 \\
1,600  & S2 & 0.590 & 0.565 & 0.615 & 0.570 & 0.564 & 0.588 & 8,628 \\
\midrule
1,800  & S0 & 0.866 & 0.706 & 0.880 & 0.768 & 0.718 & 0.791 & 16,204 \\
1,800  & S1 & 0.756 & 0.719 & 0.816 & 0.681 & 0.643 & 0.719 & 82,705 \\
1,800  & S2 & 0.536 & 0.527 & 0.520 & 0.528 & 0.529 & 0.525 & 10,506 \\
\midrule
1,900  & S0 & 0.868 & 0.697 & 0.873 & 0.773 & 0.728 & 0.794 & 18,075 \\
1,900  & S1 & 0.744 & 0.715 & 0.778 & 0.678 & 0.649 & 0.708 & 90,786 \\
1,900  & S2 & 0.510 & 0.505 & 0.939 & 0.510 & 0.506 & 0.657 & 11,338 \\
\bottomrule
\end{tabular}%
}
\end{table}

The S0 $>$ S1 $>$ S2 difficulty ordering is preserved at every subset size. S0 and S1 metrics are stable across the full size range (AUROC std $0.029$ and $0.030$ respectively, Accuracy std $0.032$ and $0.021$).
S2 metrics decline as the drug pool grows (AUROC std $0.059$, range $0.51$--$0.68$, Accuracy std $0.048$), which is expected because the Zero-Shot prompt provides no KG mediating context.
Without explicit mediating evidence, scaling the drug pool adds difficulty (more cold-start pairs, broader mechanism diversity) without adding usable signal, so the S2 trend reflects model and prompt limits rather than instability of the benchmark itself. 
The qualitative cold-start difficulty gradient is preserved at every size, including the 800-drug scale used in the main paper.

\subsection{Task Formulation Extension: Multi-Class Event-Type Prediction}
\label{app-B.4:multiclass}

The main paper adopts binary detection. To verify that the same drug-wise cold-start protocol extends to fine-grained event-type prediction, we run a reference 215-class baseline on the 800-drug subset using the same S0 / S1 / S2 splits.

\paragraph{Setup.} The model is Llama-3.2-1B with a LoRA adapter ($r{=}16$, $\alpha{=}16$, dropout~$0.1$, target $q/k/v/o$\_proj) and a randomly initialised softmax head over the 215 DrugBank event types.
Inputs use the P4 prompt as in the binary fine-tuned baseline, with the trailing Yes/No assistant turn dropped. 
Training uses cross-entropy on positives only, since DDI event types are defined only for positive pairs. We train for 4 epochs (batch size 16, lr $5\!\times\!10^{-4}$ cosine) and select the best LoRA checkpoint per split.
We verify that the per-sample composition matches the binary baseline before training. Results are 3-seed mean$\pm$std on the full S0 / S1 / S2 test sets.

\begin{table}[h]
\centering
\caption{Multi-class event-type prediction on the 800-drug subset (3-seed mean$\pm$std on the full S0 / S1 / S2 test sets).}
\label{tab:app-multiclass}
\small
\setlength{\tabcolsep}{4pt}
\begin{tabular}{@{}lrcccccc@{}}
\toprule
\textbf{Split} & $\boldsymbol{n}$ & \textbf{Top-1} & \textbf{Top-3} & \textbf{Top-5} & \textbf{Macro-F1} & \textbf{Macro-AUROC} & \textbf{Macro-AUPRC} \\
\midrule
S0 & 3{,}069  & $0.757_{\pm0.028}$ & $0.943_{\pm0.017}$ & $0.976_{\pm0.008}$ & $0.323_{\pm0.011}$ & $0.995_{\pm0.001}$ & $0.633_{\pm0.023}$ \\
S1 & 15{,}472 & $0.517_{\pm0.030}$ & $0.778_{\pm0.027}$ & $0.869_{\pm0.018}$ & $0.207_{\pm0.016}$ & $0.941_{\pm0.014}$ & $0.348_{\pm0.034}$ \\
S2 & 1{,}913  & $0.336_{\pm0.055}$ & $0.590_{\pm0.070}$ & $0.706_{\pm0.062}$ & $0.036_{\pm0.005}$ & $0.838_{\pm0.020}$ & $0.192_{\pm0.044}$ \\
\bottomrule
\end{tabular}
\end{table}

\paragraph{Findings.} The cold-start difficulty ordering S0 $>$ S1 $>$ S2 is preserved on every metric, with top-1 dropping from $0.757$ on S0 to $0.517$ on S1 and $0.336$ on S2.
Macro-F1 collapses faster than ranking metrics (S2 macro-F1 $0.036$ vs.\ macro-AUROC $0.838$), reflecting the long-tailed nature of the 215-type label space where many rare types attract zero correct predictions on the cold-start fold. 
The same mechanism-stratified evaluation framework therefore transfers cleanly to the multi-class formulation, while exposing a long-tail challenge that binary detection sidesteps.


\section{Baseline Implementations}
\label{app-C:baselines-section}

\subsection{Method Descriptions and Hyperparameters}
\label{app-C.1:baselines}

We implement eight conventional baselines under a unified training protocol so that all methods see identical drug-pair splits, identical positive edges, and identical per-epoch negative sets. 
This subsection describes each baseline in two paragraphs, the original method and our adaptation to \textsc{ColdDDI}'s binary cold-start setting, followed by the consolidated hyperparameter table. Code, configs, and re-run scripts are released alongside the paper.

\paragraph{DeepDDI~\cite{DeepDDI_2018}.}
Representative matrix-based baseline. Each drug is encoded as a Structural Similarity Profile (SSP) computed by Tanimoto similarity over Morgan fingerprints (radius 2, 2048 bits) against a reference set of training drugs, then PCA-projected to 50 dimensions. 
The original release uses an 86-way softmax over interaction types.

We re-key the SSP reference set to the per-seed $G_1$ pool so that the encoder never sees test-side drugs, swap the 86-way classification head for a single sigmoid logit emitting a binary DDI probability, and keep all other layers (9 fully-connected blocks, hidden size 2048, dropout 0.3) unchanged.
Code is provided in \texttt{coldddi/baselines/deepddi/} and selected through \texttt{evaluate.py} with \texttt{--method deepddi}.

\paragraph{SSI-DDI~\cite{SSI-DDI-2021}.}
Substructure-substructure interaction GNN.
 Each drug is encoded as a molecular graph over RDKit atom features and passed through a multi-layer GAT, then cross-drug substructure attention produces the pair representation.

The original release already supports binary DDI classification under their KGE-style head, so we keep the architecture and only re-route the data pipeline through our cold-start dataset. 
Code is provided in \texttt{coldddi/baselines/ssi\_ddi/} and selected through \texttt{evaluate.py} with \texttt{--method ssi\_ddi}.

\paragraph{DSN-DDI~\cite{DSN-DDI-2023}.}
Dual-channel attention GNN over the bipartite atom-level graph of each drug pair. 
Per-pair molecular GAT layers are interleaved with intra-graph and inter-graph attention to mix local atom features with cross-drug substructure interactions, and the resulting drug embeddings are scored through a RESCAL relation head.

The released code targets the 86-class DrugBank type space. We set the relation count to one so that the RESCAL head reduces to a single binary score, and re-route the data pipeline through our cold-start FoldBundle.
Code is provided in \texttt{coldddi/baselines/dsn\_ddi/} and selected through \texttt{evaluate.py} with \texttt{--method dsn\_ddi}.

\paragraph{HDN-DDI~\cite{HDN-DDI-2025}.}
Hierarchical drug-drug network that extends DSN-DDI with substructure-level message passing.

We apply the same single-relation RESCAL reduction as DSN-DDI, keep the hierarchical encoder unchanged, and re-route the data pipeline through our cold-start dataset.
Code is provided in \texttt{coldddi/baselines/hdn\_ddi/} and selected through \texttt{evaluate.py} with \texttt{--method hdn\_ddi}.

\paragraph{EmerGNN~\cite{EmerGNN-2023}.}
Bi-directional flow GNN over a DrugBank-derived knowledge graph for emerging drug pairs.
Each entity is initialized with Morgan fingerprints, and per-layer attention-weighted relation aggregation propagates information from both endpoint drugs through the KG.

The original release relies on \texttt{torchdrug}'s CUDA kernels for relation message passing, which our environment does not support.
We reimplement EmerGNN in pure PyTorch, replacing \texttt{generalized\_rspmm} with dense \texttt{einsum} and \texttt{torch\_scatter.scatter\_add} with \texttt{tensor.index\_add\_}. 
The architecture and update equations are otherwise identical. 
The KG is built from our 5 entity-association tables (transporter, enzyme, carrier, pathway, target). 
The original relation-prediction head is replaced with a single binary head scoring the (head, tail) pair. 
The pure-PyTorch implementation is provided in \texttt{coldddi/baselines/emergnn/} and selected through \texttt{evaluate.py} with \texttt{--method emergnn}.

\paragraph{TIGER~\cite{TIGER-2024}.}
KG-integrated graph attention model.
Each drug's molecular graph is fused with a KG random-walk subgraph rooted at the drug, and a transformer-style attention encoder produces drug embeddings.

The original head supports multi-class prediction. 
We use a 2-class head with binary cross-entropy, restrict the random-walk subgraph to relations whose endpoints lie in $G_1$ at training time, and route the data pipeline through our cold-start dataset. 
Code is provided in \texttt{coldddi/baselines/tiger/} and selected through \texttt{evaluate.py} with \texttt{--method tiger}.

\paragraph{MKG-FENN~\cite{MKG-FENN-2024}.}
Multi-modal KG fusion network with neighbor-attention pooling over drug-entity neighborhoods.

The original release classifies into 215 types. 
We replace the final softmax layer with a binary sigmoid head, keep the multi-channel KG fusion unchanged, and route the data pipeline through our cold-start dataset so that the neighbor-attention pool excludes $G_2$ drugs at training time. 
Code is provided in \texttt{coldddi/baselines/mkg\_fenn/} and selected through \texttt{evaluate.py} with \texttt{--method mkg\_fenn}.

\paragraph{TextDDI~\cite{TextDDI-2023}.}
RoBERTa-based text encoder that fine-tunes a binary classifier on each drug pair's concatenated natural-language descriptions.
Per-drug descriptions are drawn from the original TextDDI dictionary (drug profiles in medical text), with a knowledge-graph fallback over targets, enzymes, transporters, and carriers if the dictionary entry is missing.

The released model already targets binary DDI detection and we reuse it as-is, only re-routing the data pipeline through our cold-start dataset so that test-side drugs never appear in the textual context retrieved at training time.
Code is provided in \texttt{coldddi/baselines/textddi/} and selected through \texttt{evaluate.py} with \texttt{--method textddi}.

\paragraph{Hyperparameter Quick-Reference.}
Hyperparameters largely follow the settings reported in each baseline's original publication, with light task-specific tuning on the cold-start splits.
We list only the final chosen values, which are already wired in as the script defaults in the released repository.
All eight conventional baselines were trained on every one of the six datasets covered by this paper (1{,}900-drug full benchmark and 800-drug subset, each over three random seeds. See Appendix~\ref{app-B.1:split-stats} and Appendix~\ref{app-B.3.1:subset-construction}).
The reported hyperparameters are shared across all six runs per method. Wall-clock figures are approximate end-to-end training on a single RTX~5090 GPU for one run on the 1{,}900-drug full benchmark and the 800-drug subset (Appendix~\ref{app-A.6.3:hardware-software}).

\begin{table}[h]
\centering
\caption{Hyperparameters used for the eight conventional baselines, matching the paper presets defined in \texttt{coldddi/baselines/<method>/baseline.py}. Wall-clock is approximate end-to-end training on a single RTX~5090 for one run, reported separately for the 800-drug subset and the 1{,}900-drug full benchmark.}
\label{tab:baseline-hparams}
\small
\setlength{\tabcolsep}{4pt}
\renewcommand{\arraystretch}{1.05}
\resizebox{\textwidth}{!}{%
\begin{tabular}{@{}l ccccccc l r rr@{}}
\toprule
\textbf{Method} & \textbf{Optim} & \textbf{LR} & \textbf{Weight decay} & \textbf{Batch} & \textbf{Epochs} & \textbf{Patience} & \textbf{Key arch dims} & \textbf{Params} & \multicolumn{2}{c}{\textbf{Wall-clock (5090)}} \\
\cmidrule(lr){10-11}
& & & & & & & & & \textbf{800} & \textbf{1{,}900} \\
\midrule
DeepDDI    & Adam  & $10^{-3}$ & $0$               & 256  & 100 & 15 & SSP=50, hidden=2048, layers=9, dropout=0.3      & 29.6M  & $\sim$10 min & $\sim$1 h \\
SSI-DDI    & Adam  & $10^{-2}$ & $5\times10^{-4}$  & 1024 & 150 & 50 & atom feats=55, KGE dim=64                       & 0.7M   & $\sim$50 min & $\sim$10 h \\
DSN-DDI    & Adam  & $10^{-3}$ & $5\times10^{-4}$  & 512  &  50 & 10 & hidden=128, KGE dim=128, dropout=0.2            & 1.4M   & $\sim$1 h    & $\sim$14 h \\
HDN-DDI    & Adam  & $10^{-3}$ & $5\times10^{-4}$  & 512  &  50 & 10 & hidden=128, KGE dim=128                         & 1.6M   & $\sim$1 h    & $\sim$14 h \\
EmerGNN    & Adam  & $10^{-3}$ & $10^{-8}$         & 32   &  40 & 10 & $n_\mathrm{dim}$=64, length=3, feat=Morgan FP   & 0.5M   & $\sim$40 min & $\sim$10 h \\
TIGER      & Adam  & $10^{-3}$ & $10^{-4}$         & 128  &  50 & 0  & layer=2, $d_\mathrm{dim}$=64, walk=randomWalk, fixed-num=32, dropout=0.2 & 8.2M & $\sim$3 h & $\sim$19 h \\
MKG-FENN   & Adam  & $10^{-2}$ & $10^{-8}$         & 256  &  50 & 15 & embedding=128, neighbor sample=6, dropout=0.3   & 106.5M & $\sim$2 h    & $\sim$12 h \\
TextDDI    & AdamW & $10^{-5}$ & $10^{-6}$         & 32   &  30 & 10 & RoBERTa-base, max length=256                    & 124.6M & $\sim$8 h    & $\sim$40 h \\
\bottomrule
\end{tabular}%
}
\end{table}

\subsection{Training Protocol and Hardware}
\label{app-C.2:training-protocol}

\paragraph{Unified Training Protocol.}
Three protocol details are shared across all baselines and the LLM-FT pipeline, removing common confounds in cross-method comparison.

\begin{itemize}[leftmargin=*,itemsep=1pt,topsep=1pt,parsep=0pt]
    \item \textbf{Identical Splits.} All methods consume the same FoldBundle pkl produced by our split-construction pipeline (Appendix~\ref{app-B.1:split-stats}), so train / val-S0 / val-S1 / val-S2 / test-S0 / test-S1 / test-S2 partitions are byte-identical across methods.
    
    \item \textbf{Capped Negative Budget for LLM-Baseline Parity.} To keep the binary-classification protocol fair to LLM-FT (which sees one positive and one negative per training pair and cannot draw fresh negatives each epoch), every conventional baseline is held to the same finite negative budget. The bundle pre-samples four independent training-negative sets $\mathcal{N}^{(0)}, \mathcal{N}^{(1)}, \mathcal{N}^{(2)}, \mathcal{N}^{(3)}$ from $\overline{\mathcal{E}}^{(G_1)}$, each at the configured ratio. At training epoch $t$, the active negative set is $\mathcal{N}^{(t \bmod 4)}$. Methods with longer schedules rotate through the same four sets multiple times rather than seeing fresh draws each epoch, so a high-throughput optimizer cannot quietly absorb more negative supervision than LLM-FT does and the resulting gap reflects representational capacity rather than negative-sample volume. This eliminates extra-data variance from the comparison.
    
    \item \textbf{Fixed Validation and Test Negatives.} Validation and test negatives are sampled once from the appropriate setting-specific complement space and held constant across all methods and epochs (Appendix~\ref{app-B.2.1:neg-default}), so model selection and final scoring use identical evaluation sets.
\end{itemize}

\paragraph{Hardware.}
All baseline runs reported in this paper were measured on a mix of NVIDIA RTX~5090 (32~GB, local workstation) and 2$\times$A40 (48~GB, cluster) GPUs. 
Per-method wall-clock and parameter counts are listed in \cref{tab:baseline-hparams}. 
The full 1{,}900-drug benchmark requires roughly $5\times$--$15\times$ longer per baseline depending on architecture. The edge count alone scales $\sim 6\times$ (Appendix~\ref{app-B.1:split-stats}), but methods with pair-level attention or random-walk overhead scale faster.
LLM-FT wall-clock is reported separately in Appendix~\ref{app-D.2:lora}.

\subsection{Full Test Metrics}
\label{app-C.3:baseline-full-metrics}

For each scale we split the per-split metric breakdown into a ranking-metric table (AUC-ROC, AUC-PRC) and a threshold-metric table (F1, accuracy) so that each cell stays legible. \cref{tab:baseline-full-metrics-800-auc} and \cref{tab:baseline-full-metrics-800-f1acc} cover the 800-drug subset, averaged over three random seeds. \cref{tab:baseline-full-metrics-1900-auc} and \cref{tab:baseline-full-metrics-1900-f1acc} cover the 1{,}900-drug full benchmark. All four tables are produced by \texttt{scripts/render\_baseline\_full\_metrics\_tex.py} from the per-fold \texttt{metric.json} files.

\begin{table}[h]
\centering
\caption{Ranking metrics (AUC-ROC, AUC-PRC) for the eight conventional baselines on the 800-drug subset (3-seed mean$\pm$std).}
\label{tab:baseline-full-metrics-800-auc}
\small
\setlength{\tabcolsep}{4pt}
\renewcommand{\arraystretch}{1.15}
\resizebox{\textwidth}{!}{%
\begin{tabular}{@{}l cc cc cc@{}}
\toprule
\textbf{Method} & \multicolumn{2}{c}{\textbf{S0}} & \multicolumn{2}{c}{\textbf{S1}} & \multicolumn{2}{c}{\textbf{S2}} \\
\cmidrule(lr){2-3} \cmidrule(lr){4-5} \cmidrule(lr){6-7}
 & AUC-ROC & AUC-PRC & AUC-ROC & AUC-PRC & AUC-ROC & AUC-PRC \\
\midrule
DeepDDI & 0.988\,\scriptsize$\pm$0.003 & 0.989\,\scriptsize$\pm$0.002 & 0.814\,\scriptsize$\pm$0.008 & 0.797\,\scriptsize$\pm$0.014 & 0.659\,\scriptsize$\pm$0.018 & 0.646\,\scriptsize$\pm$0.011 \\
SSI-DDI & 0.876\,\scriptsize$\pm$0.026 & 0.871\,\scriptsize$\pm$0.027 & 0.716\,\scriptsize$\pm$0.015 & 0.699\,\scriptsize$\pm$0.011 & 0.614\,\scriptsize$\pm$0.037 & 0.610\,\scriptsize$\pm$0.025 \\
DSN-DDI & 0.981\,\scriptsize$\pm$0.004 & 0.980\,\scriptsize$\pm$0.005 & 0.876\,\scriptsize$\pm$0.013 & 0.876\,\scriptsize$\pm$0.008 & 0.758\,\scriptsize$\pm$0.052 & 0.747\,\scriptsize$\pm$0.061 \\
HDN-DDI & 0.909\,\scriptsize$\pm$0.011 & 0.908\,\scriptsize$\pm$0.015 & 0.741\,\scriptsize$\pm$0.005 & 0.721\,\scriptsize$\pm$0.008 & 0.634\,\scriptsize$\pm$0.035 & 0.628\,\scriptsize$\pm$0.041 \\
EmerGNN & 0.982\,\scriptsize$\pm$0.001 & 0.984\,\scriptsize$\pm$0.002 & 0.805\,\scriptsize$\pm$0.012 & 0.793\,\scriptsize$\pm$0.024 & 0.716\,\scriptsize$\pm$0.020 & 0.701\,\scriptsize$\pm$0.032 \\
TIGER & 0.971\,\scriptsize$\pm$0.004 & 0.973\,\scriptsize$\pm$0.004 & 0.729\,\scriptsize$\pm$0.008 & 0.694\,\scriptsize$\pm$0.013 & 0.566\,\scriptsize$\pm$0.014 & 0.556\,\scriptsize$\pm$0.016 \\
MKG-FENN & 0.997\,\scriptsize$\pm$0.001 & 0.998\,\scriptsize$\pm$0.000 & 0.725\,\scriptsize$\pm$0.002 & 0.685\,\scriptsize$\pm$0.002 & 0.566\,\scriptsize$\pm$0.009 & 0.544\,\scriptsize$\pm$0.011 \\
TextDDI & 0.992\,\scriptsize$\pm$0.002 & 0.993\,\scriptsize$\pm$0.002 & 0.785\,\scriptsize$\pm$0.002 & 0.762\,\scriptsize$\pm$0.012 & 0.675\,\scriptsize$\pm$0.024 & 0.674\,\scriptsize$\pm$0.021 \\
\bottomrule
\end{tabular}%
}
\end{table}

\begin{table}[h]
\centering
\caption{Threshold metrics (F1, accuracy) on the 800-drug subset, same runs as \cref{tab:baseline-full-metrics-800-auc}.}
\label{tab:baseline-full-metrics-800-f1acc}
\small
\setlength{\tabcolsep}{4pt}
\renewcommand{\arraystretch}{1.15}
\resizebox{\textwidth}{!}{%
\begin{tabular}{@{}l cc cc cc@{}}
\toprule
\textbf{Method} & \multicolumn{2}{c}{\textbf{S0}} & \multicolumn{2}{c}{\textbf{S1}} & \multicolumn{2}{c}{\textbf{S2}} \\
\cmidrule(lr){2-3} \cmidrule(lr){4-5} \cmidrule(lr){6-7}
 & F1 & Acc & F1 & Acc & F1 & Acc \\
\midrule
DeepDDI & 0.941\,\scriptsize$\pm$0.007 & 0.943\,\scriptsize$\pm$0.006 & 0.721\,\scriptsize$\pm$0.005 & 0.735\,\scriptsize$\pm$0.004 & 0.555\,\scriptsize$\pm$0.013 & 0.611\,\scriptsize$\pm$0.005 \\
SSI-DDI & 0.801\,\scriptsize$\pm$0.025 & 0.798\,\scriptsize$\pm$0.026 & 0.656\,\scriptsize$\pm$0.004 & 0.659\,\scriptsize$\pm$0.012 & 0.548\,\scriptsize$\pm$0.079 & 0.585\,\scriptsize$\pm$0.037 \\
DSN-DDI & 0.929\,\scriptsize$\pm$0.004 & 0.928\,\scriptsize$\pm$0.005 & 0.764\,\scriptsize$\pm$0.034 & 0.784\,\scriptsize$\pm$0.018 & 0.617\,\scriptsize$\pm$0.019 & 0.674\,\scriptsize$\pm$0.029 \\
HDN-DDI & 0.828\,\scriptsize$\pm$0.011 & 0.826\,\scriptsize$\pm$0.008 & 0.670\,\scriptsize$\pm$0.016 & 0.680\,\scriptsize$\pm$0.005 & 0.606\,\scriptsize$\pm$0.035 & 0.590\,\scriptsize$\pm$0.037 \\
EmerGNN & 0.932\,\scriptsize$\pm$0.002 & 0.933\,\scriptsize$\pm$0.002 & 0.714\,\scriptsize$\pm$0.022 & 0.735\,\scriptsize$\pm$0.015 & 0.614\,\scriptsize$\pm$0.039 & 0.659\,\scriptsize$\pm$0.022 \\
TIGER & 0.907\,\scriptsize$\pm$0.007 & 0.905\,\scriptsize$\pm$0.008 & 0.654\,\scriptsize$\pm$0.014 & 0.667\,\scriptsize$\pm$0.007 & 0.511\,\scriptsize$\pm$0.022 & 0.549\,\scriptsize$\pm$0.010 \\
MKG-FENN & 0.976\,\scriptsize$\pm$0.002 & 0.976\,\scriptsize$\pm$0.002 & 0.663\,\scriptsize$\pm$0.029 & 0.661\,\scriptsize$\pm$0.003 & 0.667\,\scriptsize$\pm$0.000 & 0.506\,\scriptsize$\pm$0.005 \\
TextDDI & 0.959\,\scriptsize$\pm$0.007 & 0.960\,\scriptsize$\pm$0.007 & 0.699\,\scriptsize$\pm$0.021 & 0.717\,\scriptsize$\pm$0.005 & 0.614\,\scriptsize$\pm$0.034 & 0.630\,\scriptsize$\pm$0.016 \\
\bottomrule
\end{tabular}%
}
\end{table}

\begin{table}[h]
\centering
\caption{Ranking metrics (AUC-ROC, AUC-PRC) on the 1{,}900-drug full benchmark (3-seed mean$\pm$std).}
\label{tab:baseline-full-metrics-1900-auc}
\small
\setlength{\tabcolsep}{4pt}
\renewcommand{\arraystretch}{1.15}
\resizebox{\textwidth}{!}{%
\begin{tabular}{@{}l cc cc cc@{}}
\toprule
\textbf{Method} & \multicolumn{2}{c}{\textbf{S0}} & \multicolumn{2}{c}{\textbf{S1}} & \multicolumn{2}{c}{\textbf{S2}} \\
\cmidrule(lr){2-3} \cmidrule(lr){4-5} \cmidrule(lr){6-7}
 & AUC-ROC & AUC-PRC & AUC-ROC & AUC-PRC & AUC-ROC & AUC-PRC \\
\midrule
DeepDDI & 0.997\,\scriptsize$\pm$0.000 & 0.997\,\scriptsize$\pm$0.000 & 0.819\,\scriptsize$\pm$0.003 & 0.804\,\scriptsize$\pm$0.008 & 0.672\,\scriptsize$\pm$0.009 & 0.663\,\scriptsize$\pm$0.016 \\
SSI-DDI & 0.829\,\scriptsize$\pm$0.010 & 0.819\,\scriptsize$\pm$0.013 & 0.706\,\scriptsize$\pm$0.006 & 0.688\,\scriptsize$\pm$0.006 & 0.627\,\scriptsize$\pm$0.003 & 0.624\,\scriptsize$\pm$0.014 \\
DSN-DDI & 0.950\,\scriptsize$\pm$0.038 & 0.947\,\scriptsize$\pm$0.050 & 0.827\,\scriptsize$\pm$0.056 & 0.818\,\scriptsize$\pm$0.065 & 0.711\,\scriptsize$\pm$0.063 & 0.696\,\scriptsize$\pm$0.062 \\
HDN-DDI & 0.853\,\scriptsize$\pm$0.006 & 0.846\,\scriptsize$\pm$0.010 & 0.736\,\scriptsize$\pm$0.003 & 0.718\,\scriptsize$\pm$0.003 & 0.645\,\scriptsize$\pm$0.002 & 0.638\,\scriptsize$\pm$0.009 \\
EmerGNN & 0.980\,\scriptsize$\pm$0.001 & 0.982\,\scriptsize$\pm$0.001 & 0.799\,\scriptsize$\pm$0.000 & 0.776\,\scriptsize$\pm$0.007 & 0.707\,\scriptsize$\pm$0.006 & 0.680\,\scriptsize$\pm$0.012 \\
TIGER & 0.976\,\scriptsize$\pm$0.002 & 0.977\,\scriptsize$\pm$0.002 & 0.771\,\scriptsize$\pm$0.027 & 0.746\,\scriptsize$\pm$0.032 & 0.630\,\scriptsize$\pm$0.015 & 0.616\,\scriptsize$\pm$0.022 \\
MKG-FENN & 0.999\,\scriptsize$\pm$0.000 & 0.999\,\scriptsize$\pm$0.000 & 0.703\,\scriptsize$\pm$0.005 & 0.664\,\scriptsize$\pm$0.014 & 0.563\,\scriptsize$\pm$0.007 & 0.545\,\scriptsize$\pm$0.004 \\
TextDDI & 0.996\,\scriptsize$\pm$0.002 & 0.996\,\scriptsize$\pm$0.003 & 0.786\,\scriptsize$\pm$0.005 & 0.759\,\scriptsize$\pm$0.005 & 0.658\,\scriptsize$\pm$0.004 & 0.649\,\scriptsize$\pm$0.011 \\
\bottomrule
\end{tabular}%
}
\end{table}

\begin{table}[h]
\centering
\caption{Threshold metrics (F1, accuracy) on the 1{,}900-drug full benchmark, same runs as \cref{tab:baseline-full-metrics-1900-auc}.}
\label{tab:baseline-full-metrics-1900-f1acc}
\small
\setlength{\tabcolsep}{4pt}
\renewcommand{\arraystretch}{1.15}
\resizebox{\textwidth}{!}{%
\begin{tabular}{@{}l cc cc cc@{}}
\toprule
\textbf{Method} & \multicolumn{2}{c}{\textbf{S0}} & \multicolumn{2}{c}{\textbf{S1}} & \multicolumn{2}{c}{\textbf{S2}} \\
\cmidrule(lr){2-3} \cmidrule(lr){4-5} \cmidrule(lr){6-7}
 & F1 & Acc & F1 & Acc & F1 & Acc \\
\midrule
DeepDDI & 0.973\,\scriptsize$\pm$0.000 & 0.973\,\scriptsize$\pm$0.000 & 0.726\,\scriptsize$\pm$0.007 & 0.743\,\scriptsize$\pm$0.004 & 0.575\,\scriptsize$\pm$0.019 & 0.624\,\scriptsize$\pm$0.006 \\
SSI-DDI & 0.756\,\scriptsize$\pm$0.012 & 0.751\,\scriptsize$\pm$0.013 & 0.643\,\scriptsize$\pm$0.011 & 0.652\,\scriptsize$\pm$0.007 & 0.587\,\scriptsize$\pm$0.021 & 0.593\,\scriptsize$\pm$0.002 \\
DSN-DDI & 0.887\,\scriptsize$\pm$0.055 & 0.879\,\scriptsize$\pm$0.064 & 0.742\,\scriptsize$\pm$0.041 & 0.744\,\scriptsize$\pm$0.059 & 0.667\,\scriptsize$\pm$0.076 & 0.656\,\scriptsize$\pm$0.061 \\
HDN-DDI & 0.775\,\scriptsize$\pm$0.011 & 0.772\,\scriptsize$\pm$0.007 & 0.669\,\scriptsize$\pm$0.001 & 0.675\,\scriptsize$\pm$0.002 & 0.597\,\scriptsize$\pm$0.018 & 0.603\,\scriptsize$\pm$0.002 \\
EmerGNN & 0.932\,\scriptsize$\pm$0.004 & 0.932\,\scriptsize$\pm$0.004 & 0.728\,\scriptsize$\pm$0.002 & 0.734\,\scriptsize$\pm$0.001 & 0.644\,\scriptsize$\pm$0.011 & 0.657\,\scriptsize$\pm$0.007 \\
TIGER & 0.917\,\scriptsize$\pm$0.006 & 0.917\,\scriptsize$\pm$0.006 & 0.659\,\scriptsize$\pm$0.043 & 0.694\,\scriptsize$\pm$0.026 & 0.500\,\scriptsize$\pm$0.074 & 0.581\,\scriptsize$\pm$0.020 \\
MKG-FENN & 0.989\,\scriptsize$\pm$0.001 & 0.989\,\scriptsize$\pm$0.001 & 0.612\,\scriptsize$\pm$0.033 & 0.642\,\scriptsize$\pm$0.013 & 0.667\,\scriptsize$\pm$0.000 & 0.500\,\scriptsize$\pm$0.000 \\
TextDDI & 0.973\,\scriptsize$\pm$0.011 & 0.973\,\scriptsize$\pm$0.011 & 0.711\,\scriptsize$\pm$0.016 & 0.724\,\scriptsize$\pm$0.011 & 0.564\,\scriptsize$\pm$0.014 & 0.617\,\scriptsize$\pm$0.001 \\
\bottomrule
\end{tabular}%
}
\end{table}

\subsection{Stratified Test Metrics by Mechanism $\times$ Similarity}

\label{app-C.4:mech-sim-baselines}

This subsection complements \cref{sec:stratified-analysis} (D2) by cross-tabulating S2 Recall along two axes simultaneously, namely the four mechanism subtypes (PK-A, PK-B, PD-A, PD-B) and three structural-similarity tiers based on Tanimoto similarity over Morgan fingerprints (radius 2, 2048 bits) of the two drug SMILES.
Tier cutoffs follow our companion analysis. Low corresponds to $t \le 0.08$, Mid to $0.08 < t \le 0.115$, and High to $t > 0.115$. All numbers are 3-seed mean on the 800-drug subset under pattern P4 (One-Hop KG sequence).
MKG-FENN values carry $\dagger$ as in \cref{tab:stratified-recall} and are excluded from per-row bolding.

\cref{tab:app-mech-sim-baselines} reports per-cell Recall for the eight conventional baselines. 
Two patterns hold across most baselines.
\textbf{(a)}~Within each mechanism subtype, Recall shows a roughly downward trend from High to Low similarity tier, indicating that structural overlap with training drugs remains a primary signal. 
\textbf{(b)}~PK-B (no annotated mediating entity) is consistently among the hardest columns, with even High-similarity cells barely above 0.5 for most baselines.

\begin{table}[t]
\caption{S2 Recall on the 800-drug subset, broken down by mechanism subtype $\times$ Tanimoto similarity tier (3-seed mean$\pm$std). Conventional baselines. $n$ is the per-cell positive count averaged across seeds. Best per row in \textbf{bold} (excluding MKG-FENN).}
\label{tab:app-mech-sim-baselines}
\centering
\footnotesize
\setlength{\tabcolsep}{3pt}
\renewcommand{\arraystretch}{1}
\resizebox{\textwidth}{!}{%
\begin{tabular}{@{}ll r cccccccc@{}}
\toprule
\textbf{Type} & \textbf{Tier} & $n$ & \textbf{DeepDDI} & \textbf{SSI-DDI} & \textbf{DSN-DDI} & \textbf{HDN-DDI} & \textbf{EmerGNN} & \textbf{TIGER} & \textbf{MKG-FENN} & \textbf{TextDDI} \\
\midrule
PK-A & Low & 183 & $0.411_{\pm0.026}$ & $0.454_{\pm0.134}$ & $0.464_{\pm0.053}$ & $0.602_{\pm0.088}$ & $\boldsymbol{0.734_{\pm0.004}}$ & $0.373_{\pm0.097}$ & $0.974^{\dagger}_{\pm0.029}$ & $0.565_{\pm0.051}$ \\
PK-A & Mid & 250 & $0.497_{\pm0.074}$ & $0.564_{\pm0.210}$ & $0.532_{\pm0.075}$ & $0.720_{\pm0.055}$ & $\boldsymbol{0.772_{\pm0.058}}$ & $0.457_{\pm0.081}$ & $0.994^{\dagger}_{\pm0.005}$ & $0.575_{\pm0.129}$ \\
PK-A & High & 255 & $0.605_{\pm0.042}$ & $0.624_{\pm0.236}$ & $0.584_{\pm0.008}$ & $\boldsymbol{0.841_{\pm0.047}}$ & $0.815_{\pm0.051}$ & $0.518_{\pm0.034}$ & $0.994^{\dagger}_{\pm0.007}$ & $0.583_{\pm0.168}$ \\
\hdashline
PK-B & Low & 157 & $0.279_{\pm0.048}$ & $0.288_{\pm0.103}$ & $\boldsymbol{0.450_{\pm0.064}}$ & $0.305_{\pm0.142}$ & $0.216_{\pm0.022}$ & $0.357_{\pm0.045}$ & $0.986^{\dagger}_{\pm0.012}$ & $0.392_{\pm0.058}$ \\
PK-B & Mid & 129 & $0.374_{\pm0.025}$ & $0.360_{\pm0.136}$ & $0.453_{\pm0.068}$ & $0.460_{\pm0.180}$ & $0.356_{\pm0.057}$ & $0.511_{\pm0.079}$ & $0.991^{\dagger}_{\pm0.009}$ & $\boldsymbol{0.560_{\pm0.070}}$ \\
PK-B & High & 95 & $0.511_{\pm0.050}$ & $0.444_{\pm0.209}$ & $0.511_{\pm0.089}$ & $\boldsymbol{0.629_{\pm0.153}}$ & $0.404_{\pm0.075}$ & $0.490_{\pm0.004}$ & $0.990^{\dagger}_{\pm0.017}$ & $0.498_{\pm0.117}$ \\
\hdashline
PD-A & Low & 15 & $0.519_{\pm0.274}$ & $0.489_{\pm0.041}$ & $0.481_{\pm0.170}$ & $0.595_{\pm0.086}$ & $\boldsymbol{0.799_{\pm0.191}}$ & $0.516_{\pm0.144}$ & $1.000^{\dagger}_{\pm0.000}$ & $0.754_{\pm0.084}$ \\
PD-A & Mid & 17 & $0.543_{\pm0.051}$ & $0.655_{\pm0.095}$ & $0.544_{\pm0.230}$ & $0.781_{\pm0.054}$ & $0.781_{\pm0.093}$ & $0.589_{\pm0.106}$ & $1.000^{\dagger}_{\pm0.000}$ & $\boldsymbol{0.848_{\pm0.073}}$ \\
PD-A & High & 45 & $0.691_{\pm0.087}$ & $0.776_{\pm0.108}$ & $0.678_{\pm0.045}$ & $\boldsymbol{0.820_{\pm0.106}}$ & $0.733_{\pm0.137}$ & $0.667_{\pm0.068}$ & $1.000^{\dagger}_{\pm0.000}$ & $0.811_{\pm0.061}$ \\
\hdashline
PD-B & Low & 242 & $0.401_{\pm0.089}$ & $0.408_{\pm0.058}$ & $0.454_{\pm0.030}$ & $0.415_{\pm0.078}$ & $0.457_{\pm0.129}$ & $0.407_{\pm0.082}$ & $0.988^{\dagger}_{\pm0.011}$ & $\boldsymbol{0.610_{\pm0.011}}$ \\
PD-B & Mid & 251 & $0.479_{\pm0.074}$ & $0.542_{\pm0.061}$ & $0.539_{\pm0.015}$ & $\boldsymbol{0.672_{\pm0.082}}$ & $0.379_{\pm0.091}$ & $0.498_{\pm0.084}$ & $0.994^{\dagger}_{\pm0.007}$ & $0.638_{\pm0.024}$ \\
PD-B & High & 271 & $0.628_{\pm0.108}$ & $0.653_{\pm0.069}$ & $0.622_{\pm0.023}$ & $\boldsymbol{0.769_{\pm0.093}}$ & $0.471_{\pm0.136}$ & $0.551_{\pm0.092}$ & $0.984^{\dagger}_{\pm0.021}$ & $0.701_{\pm0.022}$ \\
\bottomrule
\end{tabular}%
}
\par\vspace{2pt}
{\footnotesize $^{\dagger}$~MKG-FENN saturates at near-unit Recall on every cell (overall AUC 0.566 on S2 indicates trivial output bias rather than genuine prediction). Excluded from bolding.\par}
\end{table}

\subsection{Negative-Pool Sensitivity of Stratified Metrics}
\label{app-C.5:negative-pool-sensitivity}

We examine whether the Type~A--Type~B performance gap depends
on the sampled-unrecorded comparison pool.
We use the saved S2 predictions of LLM-FT (P4) and EmerGNN
on the 800-drug subset for seeds 42, 43, and 44, without retraining.
For this supplementary analysis, positive pairs are evaluated
when both drugs have DrugBank mediator annotations.
Type~A requires a shared enzyme or transporter for PK pairs,
or a shared target for PD pairs; the remaining evaluable pairs
are Type~B.
This shared-mediator rule omits the action-role compatibility
filter used in \cref{tab:stratified-recall}.

\paragraph{Comparison pools.}
We hold the positive subsets, prediction scores, and binary
decisions fixed across three constructions:
\begin{itemize}[leftmargin=*,noitemsep,topsep=2pt]
    \item \textbf{C1: shared pool.}
    Each positive subtype is compared against the complete
    S2 sampled-unrecorded pool.
    \item \textbf{C2: size-matched pool.}
    For each positive subtype, we sample an equal number of
    unrecorded pairs without replacement from the S2 pool.
    Sampling is performed separately for each method, seed,
    and subtype, using a random generator initialized with
    seed 20260725.
    \item \textbf{C3: mediator-kind-matched pool.}
    We retain unrecorded pairs only when each endpoint has
    at least two annotations of the relevant mediator kinds:
    the enzyme and transporter counts are summed for PK,
    whereas the target count is used for PD.
\end{itemize}

For each metric $m$, we compute the A--B gap within each seed as
\[
\Delta_m =
\frac{1}{2}
\left[
m_{\mathrm{PK\text{-}A}} - m_{\mathrm{PK\text{-}B}}
+
m_{\mathrm{PD\text{-}A}} - m_{\mathrm{PD\text{-}B}}
\right],
\]
and report its mean across the three seeds.
AUPRC is computed as average precision; F1 and MCC use the
saved binary predictions.

\begin{table}[t]
\centering
\caption{Type~A--Type~B metric gaps under three
sampled-unrecorded comparison pools on the 800-drug subset
(3-seed mean). Positive values indicate higher scores for
Type~A under the supplementary shared-mediator labeling rule.}
\label{tab:negative-pool-sensitivity}
\small
\setlength{\tabcolsep}{4pt}
\renewcommand{\arraystretch}{1.15}
\begin{tabular}{@{}l ccc ccc@{}}
\toprule
\textbf{Metric}
& \multicolumn{3}{c}{\textbf{LLM-FT (P4)}}
& \multicolumn{3}{c}{\textbf{EmerGNN}} \\
\cmidrule(lr){2-4}
\cmidrule(lr){5-7}
& C1 & C2 & C3 & C1 & C2 & C3 \\
\midrule
Recall
& $+0.433$ & $+0.433$ & $+0.433$
& $+0.436$ & $+0.436$ & $+0.436$ \\
AUC-ROC
& $+0.250$ & $+0.233$ & $+0.296$
& $+0.255$ & $+0.259$ & $+0.266$ \\
AUPRC
& $+0.230$ & $+0.212$ & $+0.195$
& $+0.177$ & $+0.237$ & $+0.154$ \\
MCC
& $+0.238$ & $+0.405$ & $+0.335$
& $+0.277$ & $+0.437$ & $+0.343$ \\
F1
& $+0.108$ & $+0.300$ & $+0.142$
& $+0.159$ & $+0.357$ & $+0.208$ \\
\bottomrule
\end{tabular}
\end{table}

\cref{tab:negative-pool-sensitivity} shows positive A--B gaps for both methods under every metric and pool construction.
Thus, the Type~A gain for these two methods is not specific to Recall.

\subsection{Confidence Intervals for the A--B Recall Gap}
\label{app-C.6:recall-gap-ci}

We compute confidence intervals for the Recall gaps in \cref{tab:stratified-recall}, using the same subtype assignments and saved predictions.
For each seed $s \in \{42,43,44\}$, the gap is
\[
g_s = \frac{1}{2}\left[
R_{\mathrm{PK\text{-}A},s} - R_{\mathrm{PK\text{-}B},s}
+
R_{\mathrm{PD\text{-}A},s} - R_{\mathrm{PD\text{-}B},s}
\right].
\]
We report pointwise two-sided $95\%$ Student's $t$ confidence intervals, $\bar{g} \pm t_{0.975,2}s_g/\sqrt{3}$, where $\bar{g}$ and $s_g$ are the mean and sample standard deviation of the three paired gaps.

\begin{table}[t]
\centering
\caption{S2 A--B Recall gaps (3-seed mean$\pm$std) and pointwise $95\%$ confidence intervals on the 800-drug subset, using the labeling protocol of \cref{tab:stratified-recall}.}
\label{tab:recall-gap-ci}
\small
\setlength{\tabcolsep}{8pt}
\renewcommand{\arraystretch}{1.15}
\begin{tabular}{@{}lcc@{}}
\toprule
\textbf{Method}
& \textbf{Recall gap}
& \textbf{95\% CI} \\
\midrule
DeepDDI
& $+0.125_{\pm0.053}$
& $[-0.007,\,+0.258]$ \\
SSI-DDI
& $+0.182_{\pm0.031}$
& $[+0.104,\,+0.260]$ \\
DSN-DDI
& $+0.067_{\pm0.063}$
& $[-0.089,\,+0.224]$ \\
HDN-DDI
& $+0.215_{\pm0.077}$
& $[+0.025,\,+0.405]$ \\
EmerGNN
& $+0.388_{\pm0.078}$
& $[+0.194,\,+0.582]$ \\
TIGER
& $+0.071_{\pm0.052}$
& $[-0.057,\,+0.199]$ \\
MKG-FENN
& $+0.006_{\pm0.006}$
& $[-0.009,\,+0.021]$ \\
TextDDI
& $+0.129_{\pm0.048}$
& $[+0.009,\,+0.249]$ \\
\midrule
\rowcolor{gray!20}
LLM-FT
& $+0.398_{\pm0.105}$
& $[+0.137,\,+0.659]$ \\
\bottomrule
\end{tabular}
\end{table}

The intervals in \cref{tab:recall-gap-ci} lie entirely above zero for five of the nine methods: LLM-FT, EmerGNN, HDN-DDI, SSI-DDI, and TextDDI, supporting the Type~A advantage for these methods.

\subsection{Sensitivity to External Mediator Annotations}
\label{app-C.7:external-kg}

We examine whether the Type~A--Type~B performance gap depends on DrugBank's mediator coverage by adding annotations from PharmGKB, ChEMBL, and KEGG.
We hold the S2 predictions of LLM-FT (P4) and EmerGNN fixed on the 800-drug subset for seeds 42, 43, and 44, and recompute the A/B assignments under each enriched graph.
This analysis evaluates the sensitivity of the stratification and the associated performance gap, without retraining either model on the enriched graphs.

\paragraph{Sources and alignment.}
We use the PharmGKB drug, gene, and relationship exports dated July 5, 2026, ChEMBL 37, and KEGG DRUG annotations retrieved via the REST API on July 25, 2026.
DrugBank IDs anchor the drug namespace, PharmGKB drugs are matched through their DrugBank cross-references, ChEMBL compounds through full InChIKey matches derived from DrugBank SMILES, and KEGG drug IDs through PharmGKB cross-references.
For PharmGKB, we retain drug--gene associations with nonempty evidence fields and exclude records marked ``not associated.''
For ChEMBL, we use curated \texttt{drug\_mechanism} entries restricted to single-protein targets.
KEGG annotations are extracted from the \texttt{TARGET} and \texttt{METABOLISM} fields.
Mediator entities are aligned to the DrugBank catalog by gene symbol or normalized protein name.
Unmatched candidates receive new identifiers, with their mediator kinds assigned by name-based rules for PharmGKB, the curated target table for ChEMBL, and the source field for KEGG.
We combine each source with DrugBank and deduplicate drug--entity--kind triples, considering each source separately and all three together.
The DrugBank baseline contains 4,724 enzyme edges, 7,292 target edges, and 2,758 transporter edges.

\paragraph{Relabeling protocol.}
We use the supplementary shared-mediator rule described in Appendix~\ref{app-C.5:negative-pool-sensitivity}.
A PK pair is Type~A if its drugs share an enzyme or transporter, and a PD pair is Type~A if they share a target. 
The remaining evaluable pairs are Type~B.
Unlike the original protocol in \cref{tab:stratified-recall}, this analysis does not require action-role compatibility.
A positive pair is evaluable when both drugs have mediator annotations in the graph being evaluated.
For each seed, we compute the macro-averaged A--B Recall gap across PK and PD, then average the gaps over the three seeds.

\begin{table}[t]
\centering
\caption{Mediator-graph enrichment and A--B Recall gaps. Every row uses DrugBank as the base graph; ``All three'' adds PharmGKB, ChEMBL, and KEGG. Edge increases and drug coverage refer to the 1,900-drug pool. Recall gaps are 3-seed means on the 800-drug S2 subset. Enz., Tgt., and Trn.\ denote enzymes, targets, and transporters.}
\label{tab:external-kg}
\small
\setlength{\tabcolsep}{4pt}
\renewcommand{\arraystretch}{1.15}
\begin{tabular}{@{}l ccc c cc@{}}
\toprule
\multirow{2}{*}{\textbf{Added source}}
& \multicolumn{3}{c}{\textbf{Edge increase (\%)}}
& \multirow{2}{*}{\textbf{Drugs}}
& \multicolumn{2}{c}{\textbf{Recall gap}} \\
\cmidrule(lr){2-4}
\cmidrule(lr){6-7}
& Enz. & Tgt. & Trn.
& & LLM-FT & EmerGNN \\
\midrule
None
& $0.0$ & $0.0$ & $0.0$
& $1{,}793$ & $+0.433$ & $+0.436$ \\
PharmGKB
& $+36.3$ & $+12.9$ & $+19.8$
& $1{,}793$ & $+0.432$ & $+0.436$ \\
ChEMBL
& $+2.1$ & $+1.7$ & $+1.1$
& $1{,}803$ & $+0.435$ & $+0.436$ \\
KEGG
& $+3.2$ & $+43.5$ & $+1.5$
& $1{,}794$ & $+0.390$ & $+0.392$ \\
All three
& $+40.8$ & $+57.3$ & $+21.7$
& $1{,}804$ & $+0.395$ & $+0.392$ \\
\bottomrule
\end{tabular}
\end{table}

\paragraph{Reassignments and results.}
Among positive pairs evaluable under the DrugBank baseline in the 1{,}900-drug pool, enrichment with PharmGKB, ChEMBL, KEGG, and their union reassigns 4{,}997, 356, 33{,}942, and 38{,}590 pairs from Type~B to Type~A, respectively.
The reported results from \cref{tab:external-kg} retain a positive A--B Recall gap for both LLM-FT and EmerGNN across PharmGKB, ChEMBL, and KEGG enrichment.

\subsection{Performance Across Training-Frequency Bins}
\label{app-C.8:frequency-analysis}

We examine whether frequent training patterns explain cold-start performance on the 800-drug S2 subset.
The analysis distinguishes two levels: shared mediators and DDI types.
At the mediator level, each Type~A positive pair is indexed by its annotated primary shared mediator $M$, with $n_{\mathrm{train}}(M)$ counting training positive pairs whose two drugs share $M$.
For DDI types, all S2 positive pairs are grouped by $n_{\mathrm{train}}(t)$, the number of training positive pairs carrying type $t$.
Recall is computed within each bin for Llama-3.2-1B (P4), Gemma-3-12B (P4), and EmerGNN, pooling test examples across seeds 42, 43, and 44.

\begin{table}[t]
\centering
\caption{Frequency-stratified Recall on the 800-drug S2 subset, pooling test examples across three seeds. The mediator analysis includes Type~A positives, and the DDI-type analysis includes all positives. $n_{\mathrm{test}}$ denotes the number of test examples
in each bin.}
\label{tab:frequency-stratified-recall}
\small
\setlength{\tabcolsep}{6pt}
\renewcommand{\arraystretch}{1.10}
\begin{tabular}{@{}l r ccc@{}}
\toprule
\textbf{Training-frequency bin}
& $\boldsymbol{n_{\mathrm{test}}}$
& \textbf{Llama-3.2-1B}
& \textbf{Gemma-3-12B}
& \textbf{EmerGNN} \\
\midrule
\multicolumn{5}{l}{
\textbf{Shared mediator: $n_{\mathrm{train}}(M)$}} \\
\midrule
$1$--$10$
& $4$ & $0.750$ & $1.000$ & $1.000$ \\
$11$--$50$
& $10$ & $0.700$ & $0.800$ & $0.700$ \\
$51$--$200$
& $14$ & $0.786$ & $0.643$ & $0.571$ \\
$201$--$1{,}000$
& $108$ & $0.991$ & $0.954$ & $0.778$ \\
$>1{,}000$
& $2{,}137$ & $0.887$ & $0.737$ & $0.779$ \\
\midrule
\multicolumn{5}{l}{
\textbf{DDI type: $n_{\mathrm{train}}(t)$}} \\
\midrule
$1$--$100$
& $61$ & $0.770$ & $0.770$ & $0.639$ \\
$101$--$1{,}000$
& $340$ & $0.674$ & $0.665$ & $0.476$ \\
$1{,}001$--$5{,}000$
& $581$ & $0.657$ & $0.614$ & $0.489$ \\
$5{,}001$--$20{,}000$
& $1{,}829$ & $0.699$ & $0.660$ & $0.527$ \\
$>20{,}000$
& $2{,}927$ & $0.684$ & $0.585$ & $0.579$ \\
\bottomrule
\end{tabular}
\end{table}

Across both frequency axes, Recall is non-monotonic for all three models in
\cref{tab:frequency-stratified-recall}.
The highest-frequency bins do not consistently yield the best performance.
These patterns suggest that training-set frequency alone is insufficient to explain the observed cold-start performance or the Type~A--Type~B gap.

\section{LLM Configuration and Experiments}
\label{app-D:llm}

\definecolor{promptShared}{HTML}{555555}
\definecolor{promptP1}{HTML}{1F77B4}
\definecolor{promptP2}{HTML}{2CA02C}
\definecolor{promptP3}{HTML}{D62728}
\definecolor{promptP4}{HTML}{9467BD}
\definecolor{promptP5}{HTML}{E07B00}

\newtcolorbox{promptbox}[2]{%
  enhanced,
  colback=#1!4!white,
  colframe=#1!75!black,
  coltitle=white,
  fonttitle=\bfseries\small,
  title={#2},
  boxrule=0.5pt, arc=2pt,
  left=6pt, right=6pt, top=3pt, bottom=3pt,
  before skip=4pt, after skip=4pt,
}

\subsection{Prompt Templates (P1--P5)}
\label{app-D.1:prompts}

Each prompt is composed of six ordered content blocks and formatted into a three-role chat message (system~/~user~/~assistant). $B_1$ is the task name, a one-line identifier delivered in the system role. $B_2$ is the task instruction, which frames the model as an expert pharmacologist and states the binary DDI prediction objective. $B_3$ is the method instruction, which selects the inference strategy (zero-shot, similarity-based few-shot, KG-augmented, or mechanism-aware few-shot) and tells the model how to use whatever auxiliary content is provided. $B_4$ is the auxiliary block, which carries any retrieved reference pairs or per-drug KG facts that the chosen strategy depends on. $B_5$ is the output constraint, which pins the response to a single token (Yes or No). $B_6$ is the query block, which lists the two drugs to be predicted and any per-pattern context such as SMILES or 1-hop facts.

\begin{center}
\small
\begin{tabular}{@{}l p{0.78\linewidth}@{}}
\toprule
\textbf{Role} & \textbf{Content blocks} \\
\midrule
System    & Task name ($B_1$) + Task instruction ($B_2$, shared) \\
User      & Method instruction ($B_3$, pattern-specific) + Auxiliary block ($B_4$, pattern-specific) + Output constraint ($B_5$, shared) + Query pair ($B_6$, pattern-specific) \\
Assistant & ``Yes'' or ``No'' (single token) \\
\bottomrule
\end{tabular}
\end{center}

All five patterns share $B_1$, $B_2$, and $B_5$, and differ in $B_3$, $B_4$, and $B_6$. We first list the two shared blocks (grey), which are byte-identical across P1--P5, then show one colour-coded box per pattern bundling that pattern's $B_3$ method instruction with its $B_6$ query template (and $B_4$ auxiliary block for the few-shot patterns P2 and P5).

\begin{figure}[!htbp]
\noindent
\begin{promptbox}{promptShared}{Task instruction ($B_2$, shared)}
\small\itshape
You are an expert pharmacologist and clinical research assistant. Your objective is to determine whether a clinically significant drug-drug interaction (DDI) exists between the two provided drugs identified by the `[Prediction]' tag. Consider potential pharmacokinetic interactions and pharmacodynamic interactions in your assessment.
\end{promptbox}
\end{figure}

\begin{figure}[!htbp]
\noindent
\begin{promptbox}{promptShared}{Output constraint ($B_5$, shared)}
\small\itshape
Output Constraint: Respond strictly with exactly one token: `Yes' or `No'. Do not provide any explanations, confidence scores, punctuation, or additional text.
\end{promptbox}
\end{figure}

The five pattern-specific boxes follow. Within each box, \texttt{\{a\_name\}}, \texttt{\{b\_name\}}, \texttt{\{a\_smiles\}}, \texttt{\{b\_smiles\}} stand for the canonical DrugBank names and SMILES of the query pair, and \texttt{\{\ldots\}} under ``Facts'' stands for the one-hop biomedical entities (targets, enzymes, transporters, carriers, pathways) of each drug. Reference examples in $B_4$ are sampled from the training pool only, with self-exclusion enforced to prevent trivial identity matches.

\begin{figure}[!htbp]
\noindent
\begin{promptbox}{promptP1}{P1 Zero-Shot}
\small
\textit{\%~Method instruction.} 

This is a Zero-Shot inference mode. No external context, drug descriptions, or labeled examples are provided. You must rely exclusively on your pre-trained internal knowledge regarding drug mechanisms and chemical knowledge to make the prediction.

\smallskip
\textit{\%~Query template.}
\begin{lstlisting}[basicstyle=\small\ttfamily,frame=none,xleftmargin=4pt]
### [Prediction]:
Drug A: {a_name}
Drug B: {b_name}
### Answer:
\end{lstlisting}
\end{promptbox}
\end{figure}

\begin{figure}[!htbp]
\noindent
\begin{promptbox}{promptP2}{P2 Few-Shot SMILES}
\small
\textit{\%~Method instruction.}

This is a Few-Shot inference task enhanced by molecular structural similarity. You are provided with reference examples selected based on high SMILES similarity (Morgan fingerprints) to the target drug pair. Then you can infer the interaction status of the target pair by analyzing the shared molecular substructures and interaction patterns observed in these structural analogs.

\smallskip
\textit{\%~Auxiliary block + query template.}
\begin{lstlisting}[basicstyle=\small\ttfamily,frame=none,xleftmargin=4pt]
### Reference Examples (Structural Analogs):
### [Example k]
Drug A: {ref_a_name}, SMILES: {ref_a_smiles}
Drug B: {ref_b_name}, SMILES: {ref_b_smiles}
### Answer: {Yes|No}
... (k-shot)

### [Prediction]:
Drug A: {a_name}, SMILES: {a_smiles}
Drug B: {b_name}, SMILES: {b_smiles}
### Answer:
\end{lstlisting}
\end{promptbox}
\end{figure}

\begin{figure}[!htbp]
\noindent
\begin{promptbox}{promptP3}{P3 One-Hop KG triplets}
\small
\textit{\%~Method instruction.} 

This inference is augmented by external biomedical knowledge. For each drug, you are provided with a set of explicit 1-hop Knowledge Graph Triples (Subject, Relation, Object) representing its immediate biological associations (e.g., (Drug A, has targets, xxx target)). Integrate these structured facts with your internal knowledge to assess the likelihood of an interaction.

\smallskip
\textit{\%~Query template.}
\begin{lstlisting}[basicstyle=\small\ttfamily,frame=none,xleftmargin=4pt]
### [Prediction]:
Drug A: {a_name}
Drug A's Facts: ({a_name}, has targets, ...),
                ({a_name}, has enzymes, ...), ...
Drug B: {b_name}
Drug B's Facts: ({b_name}, has targets, ...),
                ({b_name}, has enzymes, ...), ...
### Answer:
\end{lstlisting}
\end{promptbox}
\end{figure}

\begin{figure}[!htbp]
\noindent
\begin{promptbox}{promptP4}{P4 One-Hop KG sequence}
\small
\textit{\%~Method instruction.} 

This inference is augmented by external biomedical knowledge transformed into natural language descriptions. Structured Knowledge Graph facts regarding each drug have been linearized into coherent sentences grouping biological associations by category (e.g., whose targets are xxx, xxx, and xxx). Utilize these semantic descriptions to identify potential interaction mechanisms.

\smallskip
\textit{\%~Query template.}
\begin{lstlisting}[basicstyle=\small\ttfamily,frame=none,xleftmargin=4pt]
### [Prediction]:
Drug A: {a_name}
whose targets are ...
whose enzymes are ...
whose transporters are ...
Drug B: {b_name}
whose targets are ...
whose enzymes are ...
whose transporters are ...
### Answer:
\end{lstlisting}
\end{promptbox}
\end{figure}

\begin{figure}[!htbp]
\noindent
\begin{promptbox}{promptP5}{P5 Few-Shot 2-hop}
\small
\textit{\%~Method instruction.}

This task utilizes a Mechanism-Aware Few-Shot strategy. The provided reference examples are not selected randomly, they are retrieved based on shared biological entities (targets, transporters, enzymes) from a biomedical Knowledge Graph. Specifically, the reference drugs share critical protein associations with the query drugs (with tag `[Prediction]'), implying a potential 2nd-order mechanistic similarity (Query Drug $\to$ Shared Protein $\leftarrow$ Reference Drug). Use these mechanistically related examples to infer the interaction status of the target pair.

\smallskip
\textit{\%~Auxiliary block + query template.}
\begin{lstlisting}[basicstyle=\small\ttfamily,frame=none,xleftmargin=4pt]
### Reference Examples (Mechanism-Aware):
### [Example k]
Drug A: {ref_a_name}
Drug B: {ref_b_name}
Matching Evidence:
(Query Drug A -> {Target: shared} <- {ref_a_name});
(Query Drug B -> {Target: shared} <- {ref_b_name})
### Answer: {Yes|No}
... (k-shot)

### [Prediction]:
Query Drug A: {a_name}
Query Drug B: {b_name}
### Answer:
\end{lstlisting}
\end{promptbox}
\end{figure}

When no shared biomedical entity exists for P5 retrieval, the block falls back to the SMILES-similarity retrieval used by P2.

\paragraph{Description-Leakage Verification.}
We explicitly confirm that no prompt template contains DrugBank interaction descriptions or any textual hint about whether the query pair interacts:
\begin{enumerate}[leftmargin=*,itemsep=1pt,topsep=1pt,parsep=0pt]
    \item The task instruction ($B_2$) and all method instructions ($B_3$) describe the task and the method. They never reference a specific interaction.
    \item The few-shot blocks in P2 and P5 contain reference drug pairs with Yes/No labels, but these pairs are drawn from the training pool and do not reveal the label of the query pair.
    \item The one-hop KG blocks in P3 and P4 describe each drug's individual biomedical associations (targets, enzymes, transporters, carriers, pathways), not drug-drug relations.
    \item The query format ($B_6$) lists drug names (or \texttt{[DRUG\_A]}/\texttt{[DRUG\_B]} placeholders under R1/R3/R5/R7), optionally SMILES or KG context, but never the interaction description.
\end{enumerate}
Consequently, the model must infer the interaction from drug-level features rather than from leaked annotations.

\subsection{LoRA Fine-Tuning Configuration}
\label{app-D.2:lora}

All fine-tuned LLM results in \cref{sec:cold-start-results,sec:modality-results,sec:model-diversity} are produced by Low-Rank Adaptation (LoRA)~\cite{lora-2022} of the base model, with all base-model parameters frozen.
The configuration is shared across the eleven open-weight LLMs of \cref{tab:llm-family} and across the five prompt patterns P1--P5. Only the prompt template varies.
The released training preset is defined in \texttt{coldddi/llm/trainer.py}. The P4 runner records its effective configuration in \texttt{config.json} under the run directory.

\paragraph{Hyperparameter Search.}
The LoRA settings reported below are not hand-picked. They are the best-trial output of an Optuna study run on Qwen2.5-0.5B (chosen as the cheapest base for a fast search) over 50 trials maximising val-S2 AUC-ROC on a 1{,}000-pair training subset and a 100-pair validation subset on a single random seed. 
The full search space and the best assignment are listed in \cref{tab:lora-hpo-space}. The best trial reached val AUC $0.747$. The median across the 50 trials was $0.506$, indicating that LoRA performance on cold-start S2 is highly sensitive to this configuration.
Marginal-mean comparisons over the 50 trials confirm the qualitative ranking.
Adding $o\_\text{proj}$, reducing the batch from $64$ to $16$, and lifting $(r, \alpha)$ from $(8,8)$ to $(16,16)$ each raise mean val AUC monotonically.
We carry this assignment to every other backbone in \cref{tab:llm-family} unchanged, since per-backbone re-tuning would inflate compute by $11\times$ for variance smaller than the cross-backbone S2 spread.

\begin{table}[h]
\centering
\caption{LoRA hyperparameter search space and the best trial. The Optuna study ran for 50 trials on Qwen2.5-0.5B (1{,}000 train / 100 val pairs, single random seed), maximising val-S2 AUC-ROC. Best trial value $0.747$, median across all 50 trials $0.506$.}
\label{tab:lora-hpo-space}
\small
\setlength{\tabcolsep}{6pt}
\renewcommand{\arraystretch}{1.1}
\begin{tabular}{@{}lll@{}}
\toprule
\textbf{Parameter} & \textbf{Search space} & \textbf{Best} \\
\midrule
learning rate              & $[1\!\times\!10^{-5},\;5\!\times\!10^{-4}]$ (step $5\!\times\!10^{-6}$) & $5\!\times\!10^{-4}$ \\
$(r, \alpha)$              & $\{(8,8),\,(8,16),\,(16,16),\,(16,32)\}$    & $(16, 16)$ \\
dropout                    & $[0, 0.1]$ (step $0.02$)                    & $0.1$ \\
target modules             & $\{q\!+\!v,\; q\!+\!k\!+\!v,\; q\!+\!k\!+\!v\!+\!o\}$ & $q\!+\!k\!+\!v\!+\!o$ \\
batch size                 & $\{16,\,32,\,64\}$                          & $16$ \\
\bottomrule
\end{tabular}
\end{table}

\paragraph{Adapter Hyperparameters.}
\cref{tab:lora-hparams} lists the LoRA-side configuration. Rank~$r{=}16$ and $\alpha{=}16$ keep the effective scaling at $1$, and dropout~$0.1$ matches the base-model attention dropout used during pretraining. The adapter targets the four self-attention projections ($q, k, v, o$). We deliberately do not adapt the FFN to keep the per-LLM adapter size below $40$~MB and to avoid over-writing the pretrained pharmacological-text prior.

\begin{table}[h]
\centering
\caption{LoRA adapter and training hyperparameters used for every fine-tuned LLM in the paper. The micro-batch column lists the values used on 2 A40~(96~GB).}
\label{tab:lora-hparams}
\small
\setlength{\tabcolsep}{6pt}
\renewcommand{\arraystretch}{1.1}
\begin{tabular}{@{}ll@{}}
\toprule
\textbf{Parameter} & \textbf{Value} \\
\midrule
\multicolumn{2}{@{}l}{\emph{Adapter (PEFT/LoRA)}} \\
\quad rank $r$                  & 16 \\
\quad $\alpha$                  & 16 \\
\quad dropout                   & 0.10 \\
\quad target modules            & \texttt{q\_proj, k\_proj, v\_proj, o\_proj} \\
\quad layers transformed        & 11 (last 11 transformer blocks) \\
\quad bias                      & none (frozen) \\
\quad task type                 & \texttt{CAUSAL\_LM} \\
\midrule
\multicolumn{2}{@{}l}{\emph{Training}} \\
\quad optimizer                 & \texttt{adamw\_torch} \\
\quad learning rate             & $5 \times 10^{-4}$ \\
\quad LR schedule               & cosine, warmup~ratio $0.05$ \\
\quad weight decay              & $0.0$ \\
\quad gradient clipping         & $1.0$ \\
\quad effective batch size      & 16 \\
\quad micro-batch (0.5B / 1B)   & 8 (\texttt{grad\_accum}~$=2$) \\
\quad micro-batch (3B / 4B)     & 4 (\texttt{grad\_accum}~$=4$) \\
\quad micro-batch (7B)          & 2 (\texttt{grad\_accum}~$=8$) \\
\quad micro-batch (12B--14B)    & 1 (\texttt{grad\_accum}~$=16$) \\
\quad epochs                    & 4 \\
\quad max sequence length       & 1{,}250 (Top-3 KG) / 4{,}096 (Full KG, R4--R7) \\
\quad negatives per batch       & 16 (matches default 1:1 ratio of \cref{app-B.2.1:neg-default}) \\
\quad save\_total\_limit        & 20 (per-epoch ckpts, best-ckpt selection) \\
\midrule
\multicolumn{2}{@{}l}{\emph{Tokenization \& heads}} \\
\quad answer tokens             & ``\,Yes'' / ``\,No'' (single tokens with leading space) \\
\quad loss                      & next-token cross-entropy on the answer position only \\
\quad inference temperature     & 0 (greedy) \\
\quad probability               & $\mathrm{softmax}(\mathit{logits})_{\text{Yes}}$ \\
\bottomrule
\end{tabular}
\end{table}


\paragraph{Hardware and Wall-Clock.}
The full LLM matrix (11 LLMs $\times$ 5 patterns $\times$ 3 seeds = 165 cells) was trained on 2 NVIDIA A40 (96~GB) GPUs across approximately 100 days of cluster time, equivalent to $\sim$1{,}400 single-GPU hours.
Per-cell wall-clock breaks down as $\sim$5--7~h for 1B--3B LLMs, $\sim$10--12~h for 7B, $\sim$16--20~h for 13B/14B. 
%
%
Inference (3-seed, 5-pattern, S0/S1/S2 scoring) is collected on the same GPU after training and contributes an additional $\sim 30\%$ wall-clock overhead.

\paragraph{Mask-Condition Sweep.}
The eight mask conditions R0--R7 (\cref{tab:masking-conditions}) inherit the same LoRA configuration. Only the prompt template and \texttt{max\_token\_length} ($1{,}250$ for R0--R3, $4{,}096$ for R4--R7) change.
We do not re-train per mask condition. Masking is applied at inference time on the R0-trained adapter so that all mask deltas isolate the input perturbation from supervised adaptation noise.

\paragraph{Released Artifacts.}
The release includes \texttt{scripts/run\_benchmark.py} for the P4 end-to-end pipeline and \texttt{scripts/run\_llm.py} for P1--P5 runs. Training configuration and inference are implemented in \texttt{coldddi/llm/trainer.py} and \texttt{coldddi/llm/inference.py}, respectively. The selected adapter checkpoints will be released. Base-model weights are not redistributed; users supply them through the standard Hugging Face cache.

\subsection{Direct Inference Results}
\label{app-D.3:direct-inference}

This subsection supplements \cref{sec:model-diversity} (D4) by reporting per-model direct-inference (no fine-tuning) results on S2. 
We report 11 open-weight LLMs across the 5 prompt patterns, then 2 closed-source LLMs under P4 with a probabilistic-elicitation protocol.

\paragraph{Open-Weight LLMs (P1--P5).}
\cref{tab:app-e1-auc} reports S2 AUC-ROC and \cref{tab:app-e1-recall} reports S2 Recall on the 800-drug subset. 
All numbers come from a single random seed since direct inference is deterministic.

\begin{table}[t]
\caption{S2 AUC-ROC under direct inference (no fine-tuning) on the 800-drug subset (single random seed). Best per column in \textbf{bold}.}
\label{tab:app-e1-auc}
\centering
\small
\setlength{\tabcolsep}{4pt}
\renewcommand{\arraystretch}{1.05}
\begin{tabular}{@{}l ccccc@{}}
\toprule
\textbf{Model} & \textbf{P1} & \textbf{P2} & \textbf{P3} & \textbf{P4} & \textbf{P5} \\
\midrule
Llama-3.2-1B    & 0.488 & 0.523 & 0.465 & 0.525 & 0.485 \\
Llama-3.2-3B    & 0.504 & 0.488 & 0.457 & 0.459 & 0.468 \\
Llama-2-7B      & 0.535 & 0.489 & 0.549 & 0.474 & 0.496 \\
Llama-2-13B     & 0.545 & \textbf{0.533} & 0.450 & 0.513 & 0.479 \\
Qwen2.5-0.5B    & 0.525 & 0.473 & 0.511 & 0.489 & 0.513 \\
Qwen2.5-3B      & 0.502 & 0.511 & 0.555 & 0.517 & \textbf{0.579} \\
Qwen2.5-7B      & \textbf{0.563} & 0.532 & 0.534 & 0.486 & 0.560 \\
Qwen2.5-14B     & 0.541 & 0.516 & \textbf{0.584} & 0.510 & 0.555 \\
Gemma-3-1B      & 0.526 & 0.477 & 0.483 & 0.520 & 0.510 \\
Gemma-3-4B      & 0.495 & 0.457 & 0.517    & 0.496    & 0.512 \\
Gemma-3-12B     & 0.498 & 0.520 & 0.502 & \textbf{0.548} & 0.465 \\
\bottomrule
\end{tabular}
\end{table}

\begin{table}[t]
\caption{S2 Recall under direct inference (no fine-tuning) on the 800-drug subset (single random seed). Best per column in \textbf{bold}.}
\label{tab:app-e1-recall}
\centering
\small
\setlength{\tabcolsep}{4pt}
\renewcommand{\arraystretch}{1.05}
\begin{tabular}{@{}l ccccc@{}}
\toprule
\textbf{Model} & \textbf{P1} & \textbf{P2} & \textbf{P3} & \textbf{P4} & \textbf{P5} \\
\midrule
Llama-3.2-1B    & 0.000 & 0.078 & 0.850 & 0.857 & 0.102 \\
Llama-3.2-3B    & 0.000 & 0.003 & 0.000 & 0.000 & 0.000 \\
Llama-2-7B      & 0.000 & 0.008 & 0.000 & 0.000 & 0.000 \\
Llama-2-13B     & 1.000 & 0.411 & 0.964 & 0.995 & 1.000 \\
Qwen2.5-0.5B    & 1.000 & 0.980 & \textbf{1.000} & \textbf{1.000} & 1.000 \\
Qwen2.5-3B      & 0.999 & 0.631 & 1.000 & 1.000 & 0.000 \\
Qwen2.5-7B      & 1.000 & \textbf{0.981} & 0.992 & 1.000 & 1.000 \\
Qwen2.5-14B     & 1.000 & 0.980 & 1.000 & 1.000 & 1.000 \\
Gemma-3-1B      & 0.000 & 0.001 & 0.000 & 0.000 & 0.000 \\
Gemma-3-4B      & 0.000 & 0.001 & 0.000 & 0.000 & 0.000 \\
Gemma-3-12B     & 0.000 & 0.001 & 0.000 & 0.000 & 0.000 \\
\bottomrule
\end{tabular}
\end{table}

\paragraph{Open-Weight Observations.}
Across all 55 reported (LLM, pattern) cells for open-weight models, AUC ranges from $0.450$ to $0.584$ with mean $0.509$, confirming Finding 6 of \cref{sec:model-diversity} that direct inference stays near random and collapses to a single answer regardless of family or scale.
The Recall column reveals a complementary pattern. 
Most LLMs collapse to a single answer. Recall $\approx 0.000$ for the Gemma family and the Llama-3.2-3B / Llama-2-7B variants, indicating they default to ``No''. Recall $\approx 1.000$ for most of the Qwen family (Qwen2.5-3B P2 / P5 are partial at $0.631$ / $0.000$) and Llama-2-13B, indicating they default to ``Yes''.
Only Llama-3.2-1B in P3 / P4 ($0.85$, $0.86$), Llama-2-13B in P2 ($0.411$), and Qwen2.5-3B in P2 ($0.631$) emit non-trivial Recall, but their AUCs remain below $0.55$, so the agreement with ground truth is incidental rather than mechanistic.
This confirms that pretrained knowledge alone is insufficient for cold-start DDI prediction.
%

\paragraph{Closed-Source LLMs (P4).}
To verify that Finding 6 of \cref{sec:model-diversity} generalizes beyond open-weight models, we additionally evaluate two commercial LLMs, \texttt{gpt-4o-2024-08-06} (referred to as ``GPT-4o'') and \texttt{claude-sonnet-4-6} (``Claude Sonnet 4.6''), under prompt pattern P4 on the 800-drug subset S2 test set.
We use the S2 test set from a single random seed, which contains 3{,}838 unique pairs (1{,}919 positive and 1{,}919 negative).
Closed-source APIs do not expose token-level logits, so we cannot directly read $P(\text{Yes})$ from the model.
We instead modify the original P4 binary instruction to ask the model to emit a single floating-point number in $[0, 1]$ representing the probability that a clinically significant DDI exists. 
AUC-ROC is then computed on this elicited scalar and a threshold of $0.5$ is used for thresholded metrics (accuracy, precision, recall, F1). Decoding uses temperature $=0$ (greedy), a single random seed, and a single inference per pair.
Per-model results are in \cref{tab:app-e1-closed}.

\begin{table}[t]
\caption{Closed-source LLM direct inference on the 800-drug subset, S2 test set under P4 (3{,}838 unique pairs, single random seed). AUC is computed on the elicited $P(\text{Yes})$. Thresholded metrics use $0.5$ as the cutoff.}
\label{tab:app-e1-closed}
\centering
\small
\setlength{\tabcolsep}{5pt}
\renewcommand{\arraystretch}{1.05}
\begin{tabular}{@{}l ccccc@{}}
\toprule
\textbf{Model} & \textbf{AUC} & \textbf{Acc.} & \textbf{Prec.} & \textbf{Rec.} & \textbf{F1}  \\
\midrule
GPT-4o            & 0.710 & 0.666 & 0.664 & 0.672 & 0.668 \\
Claude Sonnet 4.6 & 0.751 & 0.617 & 0.886 & 0.267 & 0.411 \\
\bottomrule
\end{tabular}
\end{table}

\paragraph{Closed-Source Observations.}
Three points stand out. \textbf{(a)}~Both closed-source models lie below the open-weight LLM-FT (Llama-3.2-1B, P4) S2 AUC of $0.776$ on the same subset, confirming that Finding 6 of \cref{sec:model-diversity} extends to closed-source.
\textbf{(b)}~Both also clearly outperform every open-weight direct-inference cell in \cref{tab:app-e1-auc} (max $0.584$). Closed-source pretraining therefore raises the no-tuning ceiling, but not enough to overtake supervised LoRA fine-tuning of a 1B open model.
\textbf{(c)}~The two models exhibit notably different decision styles. GPT-4o produces balanced predictions (precision $0.66$, recall $0.67$), whereas Claude Sonnet 4.6 is highly conservative (precision $0.89$, recall $0.27$), strongly preferring the negative class when uncertain. 
AUC is threshold-invariant so this asymmetry does not affect the headline ranking, but it indicates that even at comparable AUC, closed-source LLMs differ considerably in operating selection without supervised tuning.

\subsection{Fine-Tuned Per-Model Per-Pattern Results}
\label{app-D.4:full-results}

This subsection reports the full LLM fine-tuning matrix referenced from \cref{sec:cold-start-results} and \cref{sec:model-diversity}, covering eleven open-weight LLMs across three families (Llama, Qwen, Gemma) and parameter counts spanning 0.5B to 14B, evaluated on five inference patterns (P1--P5, defined in \cref{sec:llm-patterns}) and three cold-start settings (S0 / S1 / S2) on the 800-drug subset. All numbers are 3-seed mean$\pm$std. We report AUC-ROC and Recall here.

\begin{table}[t]
\caption{S0 fine-tuned LLM AUC-ROC (top) and Recall (bottom) on the 800-drug subset, by inference pattern (3-seed mean$\pm$std). Best per column in \textbf{bold}.}
\label{tab:app-e2-s0}
\centering
\small
\setlength{\tabcolsep}{4pt}
\renewcommand{\arraystretch}{1.05}
\begin{tabular}{@{}l ccccc@{}}
\toprule
\textbf{Model} & \textbf{P1} & \textbf{P2} & \textbf{P3} & \textbf{P4} & \textbf{P5} \\
\midrule
\multicolumn{6}{@{}l}{\emph{AUC-ROC}} \\
\midrule
Llama-3.2-1B & 0.982\,\scriptsize$\pm$0.003 & \textbf{0.982}\,\scriptsize$\pm$0.005 & 0.987\,\scriptsize$\pm$0.002 & 0.987\,\scriptsize$\pm$0.001 & 0.980\,\scriptsize$\pm$0.003 \\
Llama-3.2-3B & \textbf{0.987}\,\scriptsize$\pm$0.002 & 0.965\,\scriptsize$\pm$0.036 & 0.946\,\scriptsize$\pm$0.071 & \textbf{0.991}\,\scriptsize$\pm$0.001 & \textbf{0.984}\,\scriptsize$\pm$0.003 \\
Llama-2-7B & 0.883\,\scriptsize$\pm$0.055 & 0.975\,\scriptsize$\pm$0.014 & 0.962\,\scriptsize$\pm$0.023 & 0.893\,\scriptsize$\pm$0.081 & 0.967\,\scriptsize$\pm$0.031 \\
Llama-2-13B & 0.891\,\scriptsize$\pm$0.059 & 0.930\,\scriptsize$\pm$0.007 & 0.915\,\scriptsize$\pm$0.071 & 0.903\,\scriptsize$\pm$0.044 & 0.907\,\scriptsize$\pm$0.027 \\
Qwen2.5-0.5B & 0.972\,\scriptsize$\pm$0.003 & 0.974\,\scriptsize$\pm$0.003 & 0.983\,\scriptsize$\pm$0.004 & 0.979\,\scriptsize$\pm$0.007 & 0.972\,\scriptsize$\pm$0.004 \\
Qwen2.5-3B & 0.870\,\scriptsize$\pm$0.047 & 0.945\,\scriptsize$\pm$0.066 & 0.915\,\scriptsize$\pm$0.071 & 0.953\,\scriptsize$\pm$0.061 & 0.946\,\scriptsize$\pm$0.063 \\
Qwen2.5-7B & 0.866\,\scriptsize$\pm$0.116 & 0.599\,\scriptsize$\pm$0.073 & 0.836\,\scriptsize$\pm$0.077 & 0.822\,\scriptsize$\pm$0.089 & 0.907\,\scriptsize$\pm$0.053 \\
Qwen2.5-14B & 0.889\,\scriptsize$\pm$0.079 & 0.847\,\scriptsize$\pm$0.034 & 0.864\,\scriptsize$\pm$0.062 & 0.883\,\scriptsize$\pm$0.097 & 0.906\,\scriptsize$\pm$0.070 \\
Gemma-3-1B & 0.975\,\scriptsize$\pm$0.006 & 0.975\,\scriptsize$\pm$0.004 & 0.984\,\scriptsize$\pm$0.004 & 0.983\,\scriptsize$\pm$0.002 & 0.974\,\scriptsize$\pm$0.003 \\
Gemma-3-4B & 0.984\,\scriptsize$\pm$0.002 & 0.981\,\scriptsize$\pm$0.000 & \textbf{0.990}\,\scriptsize$\pm$0.000 & 0.990\,\scriptsize$\pm$0.001 & 0.980\,\scriptsize$\pm$0.003 \\
Gemma-3-12B & 0.984\,\scriptsize$\pm$0.002 & 0.954\,\scriptsize$\pm$0.042 & \textbf{0.990}\,\scriptsize$\pm$0.002 & 0.990\,\scriptsize$\pm$0.001 & 0.983\,\scriptsize$\pm$0.003 \\
\midrule
\multicolumn{6}{@{}l}{\emph{Recall}} \\
\midrule
Llama-3.2-1B & 0.923\,\scriptsize$\pm$0.003 & \textbf{0.922}\,\scriptsize$\pm$0.013 & 0.941\,\scriptsize$\pm$0.003 & 0.938\,\scriptsize$\pm$0.007 & 0.922\,\scriptsize$\pm$0.009 \\
Llama-3.2-3B & \textbf{0.938}\,\scriptsize$\pm$0.003 & 0.876\,\scriptsize$\pm$0.092 & 0.881\,\scriptsize$\pm$0.096 & \textbf{0.949}\,\scriptsize$\pm$0.002 & \textbf{0.930}\,\scriptsize$\pm$0.010 \\
Llama-2-7B & 0.752\,\scriptsize$\pm$0.061 & 0.905\,\scriptsize$\pm$0.019 & 0.885\,\scriptsize$\pm$0.060 & 0.917\,\scriptsize$\pm$0.040 & 0.914\,\scriptsize$\pm$0.035 \\
Llama-2-13B & 0.819\,\scriptsize$\pm$0.053 & 0.799\,\scriptsize$\pm$0.020 & 0.795\,\scriptsize$\pm$0.147 & 0.783\,\scriptsize$\pm$0.088 & 0.776\,\scriptsize$\pm$0.064 \\
Qwen2.5-0.5B & 0.906\,\scriptsize$\pm$0.005 & 0.905\,\scriptsize$\pm$0.010 & 0.933\,\scriptsize$\pm$0.008 & 0.925\,\scriptsize$\pm$0.016 & 0.909\,\scriptsize$\pm$0.009 \\
Qwen2.5-3B & 0.795\,\scriptsize$\pm$0.038 & 0.898\,\scriptsize$\pm$0.044 & 0.867\,\scriptsize$\pm$0.079 & 0.896\,\scriptsize$\pm$0.083 & 0.883\,\scriptsize$\pm$0.083 \\
Qwen2.5-7B & 0.813\,\scriptsize$\pm$0.117 & 0.572\,\scriptsize$\pm$0.082 & 0.717\,\scriptsize$\pm$0.058 & 0.710\,\scriptsize$\pm$0.099 & 0.833\,\scriptsize$\pm$0.040 \\
Qwen2.5-14B & 0.822\,\scriptsize$\pm$0.085 & 0.738\,\scriptsize$\pm$0.057 & 0.776\,\scriptsize$\pm$0.055 & 0.789\,\scriptsize$\pm$0.097 & 0.823\,\scriptsize$\pm$0.080 \\
Gemma-3-1B & 0.911\,\scriptsize$\pm$0.013 & 0.904\,\scriptsize$\pm$0.011 & 0.928\,\scriptsize$\pm$0.010 & 0.927\,\scriptsize$\pm$0.002 & 0.905\,\scriptsize$\pm$0.008 \\
Gemma-3-4B & 0.928\,\scriptsize$\pm$0.007 & 0.918\,\scriptsize$\pm$0.005 & \textbf{0.949}\,\scriptsize$\pm$0.003 & 0.946\,\scriptsize$\pm$0.002 & 0.916\,\scriptsize$\pm$0.014 \\
Gemma-3-12B & 0.928\,\scriptsize$\pm$0.003 & 0.876\,\scriptsize$\pm$0.075 & 0.947\,\scriptsize$\pm$0.006 & 0.946\,\scriptsize$\pm$0.004 & 0.927\,\scriptsize$\pm$0.008 \\
\bottomrule
\end{tabular}%
\end{table}

\begin{table}[t]
\caption{S1 fine-tuned LLM AUC-ROC (top) and Recall (bottom) on the 800-drug subset (3-seed mean$\pm$std). Best per column in \textbf{bold}.}
\label{tab:app-e2-s1}
\centering
\small
\setlength{\tabcolsep}{4pt}
\renewcommand{\arraystretch}{1.05}
\begin{tabular}{@{}l ccccc@{}}
\toprule
\textbf{Model} & \textbf{P1} & \textbf{P2} & \textbf{P3} & \textbf{P4} & \textbf{P5} \\
\midrule
\multicolumn{6}{@{}l}{\emph{AUC-ROC}} \\
\midrule
Llama-3.2-1B & 0.806\,\scriptsize$\pm$0.014 & 0.804\,\scriptsize$\pm$0.024 & 0.836\,\scriptsize$\pm$0.001 & 0.843\,\scriptsize$\pm$0.009 & 0.798\,\scriptsize$\pm$0.004 \\
Llama-3.2-3B & 0.819\,\scriptsize$\pm$0.021 & 0.807\,\scriptsize$\pm$0.012 & 0.827\,\scriptsize$\pm$0.010 & 0.838\,\scriptsize$\pm$0.014 & 0.828\,\scriptsize$\pm$0.005 \\
Llama-2-7B & 0.789\,\scriptsize$\pm$0.025 & 0.797\,\scriptsize$\pm$0.006 & 0.827\,\scriptsize$\pm$0.008 & 0.771\,\scriptsize$\pm$0.067 & 0.812\,\scriptsize$\pm$0.011 \\
Llama-2-13B & 0.804\,\scriptsize$\pm$0.017 & \textbf{0.827}\,\scriptsize$\pm$0.010 & 0.839\,\scriptsize$\pm$0.024 & 0.836\,\scriptsize$\pm$0.019 & 0.811\,\scriptsize$\pm$0.007 \\
Qwen2.5-0.5B & 0.779\,\scriptsize$\pm$0.012 & 0.788\,\scriptsize$\pm$0.019 & 0.841\,\scriptsize$\pm$0.007 & 0.839\,\scriptsize$\pm$0.016 & 0.788\,\scriptsize$\pm$0.016 \\
Qwen2.5-3B & 0.718\,\scriptsize$\pm$0.039 & 0.808\,\scriptsize$\pm$0.020 & 0.828\,\scriptsize$\pm$0.023 & 0.841\,\scriptsize$\pm$0.045 & 0.801\,\scriptsize$\pm$0.052 \\
Qwen2.5-7B & 0.742\,\scriptsize$\pm$0.092 & 0.580\,\scriptsize$\pm$0.031 & 0.798\,\scriptsize$\pm$0.068 & 0.778\,\scriptsize$\pm$0.062 & 0.822\,\scriptsize$\pm$0.012 \\
Qwen2.5-14B & 0.749\,\scriptsize$\pm$0.069 & 0.759\,\scriptsize$\pm$0.048 & 0.777\,\scriptsize$\pm$0.049 & 0.812\,\scriptsize$\pm$0.062 & 0.792\,\scriptsize$\pm$0.058 \\
Gemma-3-1B & 0.785\,\scriptsize$\pm$0.013 & 0.794\,\scriptsize$\pm$0.008 & 0.830\,\scriptsize$\pm$0.012 & 0.828\,\scriptsize$\pm$0.005 & 0.798\,\scriptsize$\pm$0.013 \\
Gemma-3-4B & 0.817\,\scriptsize$\pm$0.015 & 0.816\,\scriptsize$\pm$0.009 & 0.845\,\scriptsize$\pm$0.010 & 0.849\,\scriptsize$\pm$0.004 & 0.815\,\scriptsize$\pm$0.013 \\
Gemma-3-12B & \textbf{0.837}\,\scriptsize$\pm$0.010 & 0.806\,\scriptsize$\pm$0.059 & \textbf{0.858}\,\scriptsize$\pm$0.008 & \textbf{0.861}\,\scriptsize$\pm$0.008 & \textbf{0.834}\,\scriptsize$\pm$0.010 \\
\midrule
\multicolumn{6}{@{}l}{\emph{Recall}} \\
\midrule
Llama-3.2-1B & 0.674\,\scriptsize$\pm$0.071 & 0.689\,\scriptsize$\pm$0.071 & 0.762\,\scriptsize$\pm$0.007 & \textbf{0.764}\,\scriptsize$\pm$0.020 & 0.684\,\scriptsize$\pm$0.091 \\
Llama-3.2-3B & \textbf{0.741}\,\scriptsize$\pm$0.051 & 0.683\,\scriptsize$\pm$0.030 & \textbf{0.766}\,\scriptsize$\pm$0.060 & 0.684\,\scriptsize$\pm$0.039 & 0.677\,\scriptsize$\pm$0.024 \\
Llama-2-7B & 0.640\,\scriptsize$\pm$0.020 & 0.633\,\scriptsize$\pm$0.056 & 0.735\,\scriptsize$\pm$0.009 & 0.711\,\scriptsize$\pm$0.138 & 0.687\,\scriptsize$\pm$0.071 \\
Llama-2-13B & 0.702\,\scriptsize$\pm$0.036 & 0.639\,\scriptsize$\pm$0.059 & 0.654\,\scriptsize$\pm$0.063 & 0.675\,\scriptsize$\pm$0.104 & 0.661\,\scriptsize$\pm$0.042 \\
Qwen2.5-0.5B & 0.681\,\scriptsize$\pm$0.020 & 0.697\,\scriptsize$\pm$0.050 & 0.731\,\scriptsize$\pm$0.046 & 0.712\,\scriptsize$\pm$0.065 & 0.628\,\scriptsize$\pm$0.094 \\
Qwen2.5-3B & 0.689\,\scriptsize$\pm$0.022 & \textbf{0.739}\,\scriptsize$\pm$0.014 & 0.763\,\scriptsize$\pm$0.009 & 0.706\,\scriptsize$\pm$0.055 & 0.681\,\scriptsize$\pm$0.029 \\
Qwen2.5-7B & 0.650\,\scriptsize$\pm$0.047 & 0.618\,\scriptsize$\pm$0.147 & 0.663\,\scriptsize$\pm$0.060 & 0.643\,\scriptsize$\pm$0.023 & \textbf{0.753}\,\scriptsize$\pm$0.092 \\
Qwen2.5-14B & 0.640\,\scriptsize$\pm$0.056 & 0.674\,\scriptsize$\pm$0.039 & 0.691\,\scriptsize$\pm$0.020 & 0.713\,\scriptsize$\pm$0.073 & 0.682\,\scriptsize$\pm$0.065 \\
Gemma-3-1B & 0.682\,\scriptsize$\pm$0.015 & 0.677\,\scriptsize$\pm$0.040 & 0.698\,\scriptsize$\pm$0.018 & 0.683\,\scriptsize$\pm$0.016 & 0.677\,\scriptsize$\pm$0.025 \\
Gemma-3-4B & 0.689\,\scriptsize$\pm$0.064 & 0.702\,\scriptsize$\pm$0.057 & 0.760\,\scriptsize$\pm$0.049 & 0.729\,\scriptsize$\pm$0.031 & 0.694\,\scriptsize$\pm$0.061 \\
Gemma-3-12B & 0.694\,\scriptsize$\pm$0.042 & 0.712\,\scriptsize$\pm$0.017 & 0.734\,\scriptsize$\pm$0.012 & 0.746\,\scriptsize$\pm$0.016 & 0.749\,\scriptsize$\pm$0.050 \\
\bottomrule
\end{tabular}%
\end{table}

\begin{table}[t]
\caption{S2 fine-tuned LLM AUC-ROC (top) and Recall (bottom) on the 800-drug subset (3-seed mean$\pm$std). Best per column in \textbf{bold}.}
\label{tab:app-e2-s2}
\centering
\small
\setlength{\tabcolsep}{4pt}
\renewcommand{\arraystretch}{1.05}
\begin{tabular}{@{}l ccccc@{}}
\toprule
\textbf{Model} & \textbf{P1} & \textbf{P2} & \textbf{P3} & \textbf{P4} & \textbf{P5} \\
\midrule
\multicolumn{6}{@{}l}{\emph{AUC-ROC}} \\
\midrule
Llama-3.2-1B & 0.685\,\scriptsize$\pm$0.027 & 0.696\,\scriptsize$\pm$0.008 & 0.772\,\scriptsize$\pm$0.013 & 0.776\,\scriptsize$\pm$0.020 & 0.709\,\scriptsize$\pm$0.025 \\
Llama-3.2-3B & 0.725\,\scriptsize$\pm$0.030 & 0.729\,\scriptsize$\pm$0.025 & 0.773\,\scriptsize$\pm$0.004 & 0.785\,\scriptsize$\pm$0.018 & 0.737\,\scriptsize$\pm$0.023 \\
Llama-2-7B & 0.700\,\scriptsize$\pm$0.015 & 0.704\,\scriptsize$\pm$0.009 & 0.762\,\scriptsize$\pm$0.010 & 0.731\,\scriptsize$\pm$0.039 & 0.716\,\scriptsize$\pm$0.023 \\
Llama-2-13B & 0.721\,\scriptsize$\pm$0.022 & 0.731\,\scriptsize$\pm$0.005 & 0.781\,\scriptsize$\pm$0.017 & 0.781\,\scriptsize$\pm$0.015 & 0.726\,\scriptsize$\pm$0.017 \\
Qwen2.5-0.5B & 0.673\,\scriptsize$\pm$0.012 & 0.675\,\scriptsize$\pm$0.022 & 0.759\,\scriptsize$\pm$0.018 & 0.766\,\scriptsize$\pm$0.013 & 0.693\,\scriptsize$\pm$0.021 \\
Qwen2.5-3B & 0.549\,\scriptsize$\pm$0.083 & \textbf{0.737}\,\scriptsize$\pm$0.007 & 0.756\,\scriptsize$\pm$0.003 & 0.769\,\scriptsize$\pm$0.059 & 0.713\,\scriptsize$\pm$0.085 \\
Qwen2.5-7B & 0.598\,\scriptsize$\pm$0.110 & 0.563\,\scriptsize$\pm$0.014 & 0.762\,\scriptsize$\pm$0.058 & 0.740\,\scriptsize$\pm$0.041 & 0.706\,\scriptsize$\pm$0.079 \\
Qwen2.5-14B & 0.614\,\scriptsize$\pm$0.117 & 0.671\,\scriptsize$\pm$0.105 & 0.704\,\scriptsize$\pm$0.071 & 0.750\,\scriptsize$\pm$0.053 & 0.705\,\scriptsize$\pm$0.082 \\
Gemma-3-1B & 0.680\,\scriptsize$\pm$0.021 & 0.698\,\scriptsize$\pm$0.014 & 0.757\,\scriptsize$\pm$0.036 & 0.756\,\scriptsize$\pm$0.012 & 0.707\,\scriptsize$\pm$0.028 \\
Gemma-3-4B & 0.731\,\scriptsize$\pm$0.017 & 0.726\,\scriptsize$\pm$0.014 & 0.784\,\scriptsize$\pm$0.016 & 0.781\,\scriptsize$\pm$0.015 & 0.735\,\scriptsize$\pm$0.020 \\
Gemma-3-12B & \textbf{0.756}\,\scriptsize$\pm$0.019 & 0.735\,\scriptsize$\pm$0.027 & \textbf{0.794}\,\scriptsize$\pm$0.004 & \textbf{0.798}\,\scriptsize$\pm$0.008 & \textbf{0.767}\,\scriptsize$\pm$0.015 \\
\midrule
\multicolumn{6}{@{}l}{\emph{Recall}} \\
\midrule
Llama-3.2-1B & 0.509\,\scriptsize$\pm$0.141 & 0.609\,\scriptsize$\pm$0.083 & 0.668\,\scriptsize$\pm$0.058 & \textbf{0.697}\,\scriptsize$\pm$0.021 & 0.650\,\scriptsize$\pm$0.113 \\
Llama-3.2-3B & 0.652\,\scriptsize$\pm$0.084 & 0.565\,\scriptsize$\pm$0.016 & \textbf{0.698}\,\scriptsize$\pm$0.063 & 0.584\,\scriptsize$\pm$0.048 & 0.555\,\scriptsize$\pm$0.044 \\
Llama-2-7B & 0.517\,\scriptsize$\pm$0.025 & 0.516\,\scriptsize$\pm$0.032 & 0.622\,\scriptsize$\pm$0.104 & 0.655\,\scriptsize$\pm$0.114 & 0.532\,\scriptsize$\pm$0.056 \\
Llama-2-13B & 0.581\,\scriptsize$\pm$0.109 & 0.527\,\scriptsize$\pm$0.117 & 0.549\,\scriptsize$\pm$0.069 & 0.600\,\scriptsize$\pm$0.119 & 0.556\,\scriptsize$\pm$0.062 \\
Qwen2.5-0.5B & 0.726\,\scriptsize$\pm$0.149 & 0.599\,\scriptsize$\pm$0.044 & 0.661\,\scriptsize$\pm$0.026 & 0.646\,\scriptsize$\pm$0.058 & \textbf{0.676}\,\scriptsize$\pm$0.107 \\
Qwen2.5-3B & \textbf{0.814}\,\scriptsize$\pm$0.263 & 0.612\,\scriptsize$\pm$0.104 & 0.661\,\scriptsize$\pm$0.103 & 0.618\,\scriptsize$\pm$0.050 & 0.590\,\scriptsize$\pm$0.066 \\
Qwen2.5-7B & 0.644\,\scriptsize$\pm$0.309 & 0.555\,\scriptsize$\pm$0.072 & 0.605\,\scriptsize$\pm$0.075 & 0.612\,\scriptsize$\pm$0.046 & 0.646\,\scriptsize$\pm$0.168 \\
Qwen2.5-14B & 0.490\,\scriptsize$\pm$0.104 & 0.506\,\scriptsize$\pm$0.150 & 0.629\,\scriptsize$\pm$0.018 & 0.660\,\scriptsize$\pm$0.053 & 0.508\,\scriptsize$\pm$0.023 \\
Gemma-3-1B & 0.560\,\scriptsize$\pm$0.052 & 0.551\,\scriptsize$\pm$0.038 & 0.658\,\scriptsize$\pm$0.080 & 0.569\,\scriptsize$\pm$0.078 & 0.600\,\scriptsize$\pm$0.018 \\
Gemma-3-4B & 0.594\,\scriptsize$\pm$0.089 & 0.530\,\scriptsize$\pm$0.026 & 0.658\,\scriptsize$\pm$0.111 & 0.613\,\scriptsize$\pm$0.087 & 0.602\,\scriptsize$\pm$0.045 \\
Gemma-3-12B & 0.517\,\scriptsize$\pm$0.016 & \textbf{0.650}\,\scriptsize$\pm$0.024 & 0.578\,\scriptsize$\pm$0.046 & 0.616\,\scriptsize$\pm$0.043 & 0.646\,\scriptsize$\pm$0.112 \\
\bottomrule
\end{tabular}%
\end{table}

\paragraph{Observations Across the Matrix.}
Under LoRA fine-tuning, three patterns appear consistently across the three settings.
First, one-hop KG prompting (P3, P4) is the most reliable prompt family, delivering the highest AUC and Recall for nearly every model in S1 and S2 and confirming that explicit KG context drives the gain over zero-shot or SMILES-only prompts.
Second, by combined AUC-ROC and Recall on S2, Llama-3.2-1B under P4 is the strongest configuration (AUC $0.776$, Recall $0.697$). Llama-3.2-3B under P4 (AUC $0.785$, Recall $0.584$) and Gemma-3-12B under P4 (AUC $0.798$, Recall $0.616$) reach higher AUC but markedly lower Recall.
Third, within this fine-tuned setup larger model scale does not yield a monotonic advantage on S2.
Llama-2-7B, Llama-2-13B, Qwen2.5-7B, and Qwen2.5-14B do not exceed the leading 1B--4B configuration by more than $0.04$ AUC despite a $2$--$30\times$ increase in parameter count, and the wall-clock and memory budgets reported in \cref{tab:lora-hparams} grow more steeply than this gap.
Overall, the resulting compute-to-accuracy trade-off favours the smaller backbones in practical deployment.

\paragraph{Llama-3.2-1B on the 1{,}900-Drug Full Benchmark.}
We additionally fine-tune Llama-3.2-1B under prompt pattern P4 on the 1{,}900-drug full benchmark across three random seeds, to verify that the conclusions drawn from the 800-drug subset (\cref{sec:cold-start-results,sec:model-diversity}) still hold at the full benchmark scale.
We restrict the 1{,}900-drug evaluation to P4 because (i) P4 is the strongest cell on the 800-drug subset (\cref{tab:app-e2-s2}) and is the configuration carried into the diagnostic experiments of \cref{sec:modality-results} and Appendix~\ref{app-E.2:masking-full}, and (ii) covering all five patterns at the 1{,}900-drug scale is infeasible within the submission compute budget. 
\cref{tab:app-llama1b-1900} reports 3-seed mean$\pm$std AUC-ROC and Recall on S0 / S1 / S2. 

\begin{table}[t]
\caption{Llama-3.2-1B fine-tuned under P4 on the 1{,}900-drug full benchmark (3-seed mean$\pm$std).}
\label{tab:app-llama1b-1900}
\centering
\small
\setlength{\tabcolsep}{8pt}
\renewcommand{\arraystretch}{1.15}
\begin{tabular}{@{}l ccccc@{}}
\toprule
\textbf{Setting} & \textbf{AUC-ROC} & \textbf{Accuracy} & \textbf{F1} & \textbf{Precision} & \textbf{Recall} \\
\midrule
S0 & $0.916_{\pm0.026}$ & $0.842_{\pm0.021}$ & $0.841_{\pm0.021}$ & $0.843_{\pm0.022}$ & $0.839_{\pm0.021}$ \\
S1 & $0.806_{\pm0.038}$ & $0.733_{\pm0.037}$ & $0.727_{\pm0.037}$ & $0.745_{\pm0.041}$ & $0.710_{\pm0.034}$ \\
S2 & $0.764_{\pm0.006}$ & $0.701_{\pm0.007}$ & $0.686_{\pm0.006}$ & $0.723_{\pm0.010}$ & $0.652_{\pm0.004}$ \\
\bottomrule
\end{tabular}
\end{table}

\subsubsection{Text Descriptions and Structured KG Context}
\label{app-D.4.1:description-ablation}

We examine textual drug descriptions as an alternative and a supplement to structured KG context on the 800-drug subset.
GPT-4o generates a clinical description for each drug, covering pharmacological class, indications, general mechanism, dosing, and adverse effects.
The generation prompt excludes specific enzyme, transporter, target, carrier, and interacting-drug names to reduce direct duplication of the P4 context.
A subsequent leakage audit masks residual entity mentions in descriptions.

Two additional prompt patterns use these descriptions: P7 provides drug names and clinical descriptions without KG entities, whereas P6 augments P4 with the same descriptions.
Both variants use Llama-3.2-1B with LoRA and seeds 42, 43, and 44.
These supplementary runs use one training epoch and a learning rate of $2\times10^{-4}$, with maximum sequence lengths of 1{,}280 tokens for P6 and 1{,}024 for P7.
Checkpoint selection uses validation S2 AUC-ROC.
For reference, \cref{tab:description-ablation} includes TextDDI and the P1/P4 results from \cref{tab:app-e2-s2}.

\begin{table}[t]
\caption{Text-description and KG-context comparison on the 800-drug S2 subset (3-seed mean$\pm$std). Best mean per column in \textbf{bold}.}
\label{tab:description-ablation}
\centering
\small
\setlength{\tabcolsep}{6pt}
\renewcommand{\arraystretch}{1.10}
\begin{tabular}{@{}lcc@{}}
\toprule
\textbf{Method / Prompt} & \textbf{AUC-ROC} & \textbf{Recall} \\
\midrule
TextDDI (RoBERTa)
& $0.675_{\pm0.024}$ & $0.590_{\pm0.059}$ \\
\midrule
Llama-3.2-1B, P1 (names)
& $0.685_{\pm0.027}$ & $0.509_{\pm0.141}$ \\
Llama-3.2-1B, P7 (names + descriptions)
& $0.734_{\pm0.025}$ & $0.681_{\pm0.062}$ \\
Llama-3.2-1B, P4 (KG context)
& $0.776_{\pm0.020}$ & $\mathbf{0.697}_{\pm0.021}$ \\
Llama-3.2-1B, P6 (KG context + descriptions)
& $\mathbf{0.790}_{\pm0.011}$ & $0.691_{\pm0.042}$ \\
\bottomrule
\end{tabular}
\end{table}

P7 obtains higher mean AUC-ROC than the name-only P1 reference, while P4 retains an advantage over P7.
The combined P6 setting achieves the highest mean AUC-ROC, although its Recall does not exceed P4.

\subsection{Mechanism $\times$ Similarity Analysis (Fine-Tuned LLMs)}
\label{app-D.5:mech-sim-llms}

This subsection complements \cref{sec:stratified-analysis} (D2) by cross-tabulating S2 Recall along two axes simultaneously, namely the four mechanism subtypes (PK-A, PK-B, PD-A, PD-B) and three structural-similarity tiers based on Tanimoto similarity over Morgan fingerprints (radius 2, 2048 bits) of the two drug SMILES.
Tier cutoffs follow our companion analysis. Low corresponds to $t \le 0.08$, Mid to $0.08 < t \le 0.115$, and High to $t > 0.115$. All numbers are 3-seed mean$\pm$std on the 800-drug subset under pattern P4 (One-Hop KG sequence). 
The matching cross-tab for the eight conventional baselines is in Appendix~\ref{app-C.4:mech-sim-baselines}.

To keep each table readable we split the eleven fine-tuned LLMs by family across three sub-tables.
\cref{tab:app-mech-sim-llms-llama} (Llama), \cref{tab:app-mech-sim-llms-qwen} (Qwen), and \cref{tab:app-mech-sim-llms-gemma} (Gemma) together cover models spanning four orders of magnitude in parameter count. The PK-A advantage observed for LLM-FT in \cref{tab:stratified-recall} is broadly shared.
Recall on PK-A reaches $0.84$--$0.91$ for the strongest LLM (Llama-3.2-1B), with Qwen2.5-14B as a close second at $0.79$--$0.88$, well above the strongest baseline (HDN-DDI at $0.84$ in High-tier only).
PK-B remains the universal weak spot for LLMs as well, mirroring the conventional baselines and reinforcing the \cref{sec:stratified-analysis} finding that the Type~A advantage under the supplied KG context recurs across model families.

\begin{table}[h]
\centering
\caption{S2 Recall on the 800-drug subset, broken down by mechanism subtype $\times$ Tanimoto similarity tier. Llama family fine-tuned LLMs under P4 (3-seed mean$\pm$std). $n$ is the per-cell positive count averaged across seeds. Best per row in \textbf{bold}.}
\label{tab:app-mech-sim-llms-llama}
\small
\setlength{\tabcolsep}{3pt}
\renewcommand{\arraystretch}{1.1}
\begin{tabular}{@{}ll r c c c c@{}}
\toprule
\textbf{Type} & \textbf{Tier} & $n$ & \textbf{Llama-3.2-1B} & \textbf{Llama-3.2-3B} & \textbf{Llama-2-7B} & \textbf{Llama-2-13B} \\
\midrule
PK-A & Low & 183 & \textbf{0.838}\,\scriptsize$\pm$0.113 & 0.701\,\scriptsize$\pm$0.191 & 0.776\,\scriptsize$\pm$0.157 & 0.731\,\scriptsize$\pm$0.119 \\
PK-A & Mid & 250 & \textbf{0.908}\,\scriptsize$\pm$0.082 & 0.716\,\scriptsize$\pm$0.173 & 0.818\,\scriptsize$\pm$0.125 & 0.731\,\scriptsize$\pm$0.148 \\
PK-A & High & 255 & \textbf{0.911}\,\scriptsize$\pm$0.076 & 0.782\,\scriptsize$\pm$0.141 & 0.829\,\scriptsize$\pm$0.134 & 0.740\,\scriptsize$\pm$0.128 \\
\hdashline
PK-B & Low & 157 & 0.305\,\scriptsize$\pm$0.108 & 0.240\,\scriptsize$\pm$0.049 & \textbf{0.335}\,\scriptsize$\pm$0.079 & 0.265\,\scriptsize$\pm$0.055 \\
PK-B & Mid & 129 & 0.433\,\scriptsize$\pm$0.168 & 0.352\,\scriptsize$\pm$0.132 & \textbf{0.436}\,\scriptsize$\pm$0.088 & 0.351\,\scriptsize$\pm$0.097 \\
PK-B & High & 95 & \textbf{0.511}\,\scriptsize$\pm$0.151 & 0.379\,\scriptsize$\pm$0.121 & 0.506\,\scriptsize$\pm$0.134 & 0.429\,\scriptsize$\pm$0.149 \\
\hdashline
PD-A & Low & 15 & 0.966\,\scriptsize$\pm$0.030 & 0.952\,\scriptsize$\pm$0.082 & 0.886\,\scriptsize$\pm$0.151 & \textbf{0.984}\,\scriptsize$\pm$0.027 \\
PD-A & Mid & 17 & 0.848\,\scriptsize$\pm$0.064 & 0.895\,\scriptsize$\pm$0.120 & 0.762\,\scriptsize$\pm$0.240 & \textbf{0.934}\,\scriptsize$\pm$0.060 \\
PD-A & High & 45 & \textbf{0.946}\,\scriptsize$\pm$0.059 & 0.900\,\scriptsize$\pm$0.040 & 0.874\,\scriptsize$\pm$0.186 & 0.921\,\scriptsize$\pm$0.033 \\
\hdashline
PD-B & Low & 242 & \textbf{0.569}\,\scriptsize$\pm$0.140 & 0.516\,\scriptsize$\pm$0.076 & 0.567\,\scriptsize$\pm$0.138 & 0.545\,\scriptsize$\pm$0.185 \\
PD-B & Mid & 251 & \textbf{0.602}\,\scriptsize$\pm$0.117 & 0.531\,\scriptsize$\pm$0.052 & 0.595\,\scriptsize$\pm$0.108 & 0.541\,\scriptsize$\pm$0.117 \\
PD-B & High & 271 & \textbf{0.690}\,\scriptsize$\pm$0.133 & 0.632\,\scriptsize$\pm$0.098 & 0.670\,\scriptsize$\pm$0.134 & 0.652\,\scriptsize$\pm$0.134 \\
\bottomrule
\end{tabular}%
\end{table}

\begin{table}[h]
\centering
\caption{S2 Recall on the 800-drug subset, broken down by mechanism subtype $\times$ Tanimoto similarity tier. Qwen family fine-tuned LLMs under P4 (3-seed mean$\pm$std). $n$ is the per-cell positive count averaged across seeds. Best per row in \textbf{bold}.}
\label{tab:app-mech-sim-llms-qwen}
\small
\setlength{\tabcolsep}{3pt}
\renewcommand{\arraystretch}{1.1}
\begin{tabular}{@{}ll r c c c c@{}}
\toprule
\textbf{Type} & \textbf{Tier} & $n$ & \textbf{Qwen2.5-0.5B} & \textbf{Qwen2.5-3B} & \textbf{Qwen2.5-7B} & \textbf{Qwen2.5-14B} \\
\midrule
PK-A & Low & 183 & \textbf{0.820}\,\scriptsize$\pm$0.070 & 0.708\,\scriptsize$\pm$0.076 & 0.726\,\scriptsize$\pm$0.186 & 0.794\,\scriptsize$\pm$0.111 \\
PK-A & Mid & 250 & 0.861\,\scriptsize$\pm$0.072 & 0.766\,\scriptsize$\pm$0.104 & 0.821\,\scriptsize$\pm$0.121 & \textbf{0.863}\,\scriptsize$\pm$0.055 \\
PK-A & High & 255 & 0.863\,\scriptsize$\pm$0.054 & 0.780\,\scriptsize$\pm$0.073 & 0.838\,\scriptsize$\pm$0.099 & \textbf{0.879}\,\scriptsize$\pm$0.054 \\
\hdashline
PK-B & Low & 157 & 0.310\,\scriptsize$\pm$0.075 & 0.296\,\scriptsize$\pm$0.072 & 0.281\,\scriptsize$\pm$0.143 & \textbf{0.318}\,\scriptsize$\pm$0.126 \\
PK-B & Mid & 129 & \textbf{0.488}\,\scriptsize$\pm$0.054 & 0.396\,\scriptsize$\pm$0.069 & 0.418\,\scriptsize$\pm$0.150 & 0.438\,\scriptsize$\pm$0.138 \\
PK-B & High & 95 & 0.438\,\scriptsize$\pm$0.016 & 0.420\,\scriptsize$\pm$0.100 & 0.491\,\scriptsize$\pm$0.189 & \textbf{0.511}\,\scriptsize$\pm$0.138 \\
\hdashline
PD-A & Low & 15 & 0.926\,\scriptsize$\pm$0.128 & 0.913\,\scriptsize$\pm$0.084 & 0.704\,\scriptsize$\pm$0.280 & \textbf{0.934}\,\scriptsize$\pm$0.072 \\
PD-A & Mid & 17 & 0.803\,\scriptsize$\pm$0.245 & 0.842\,\scriptsize$\pm$0.118 & 0.743\,\scriptsize$\pm$0.192 & \textbf{0.861}\,\scriptsize$\pm$0.124 \\
PD-A & High & 45 & \textbf{0.924}\,\scriptsize$\pm$0.088 & 0.816\,\scriptsize$\pm$0.191 & 0.719\,\scriptsize$\pm$0.121 & 0.910\,\scriptsize$\pm$0.066 \\
\hdashline
PD-B & Low & 242 & 0.535\,\scriptsize$\pm$0.164 & 0.543\,\scriptsize$\pm$0.130 & 0.492\,\scriptsize$\pm$0.028 & \textbf{0.558}\,\scriptsize$\pm$0.009 \\
PD-B & Mid & 251 & 0.569\,\scriptsize$\pm$0.106 & \textbf{0.605}\,\scriptsize$\pm$0.098 & 0.505\,\scriptsize$\pm$0.054 & 0.581\,\scriptsize$\pm$0.113 \\
PD-B & High & 271 & 0.611\,\scriptsize$\pm$0.146 & \textbf{0.663}\,\scriptsize$\pm$0.078 & 0.592\,\scriptsize$\pm$0.030 & 0.631\,\scriptsize$\pm$0.061 \\
\bottomrule
\end{tabular}%
\end{table}

\begin{table}[h]
\centering
\caption{S2 Recall on the 800-drug subset, broken down by mechanism subtype $\times$ Tanimoto similarity tier. Gemma family fine-tuned LLMs under P4 (3-seed mean$\pm$std). $n$ is the per-cell positive count averaged across seeds. Best per row in \textbf{bold}.}
\label{tab:app-mech-sim-llms-gemma}
\small
\setlength{\tabcolsep}{3pt}
\renewcommand{\arraystretch}{1.1}
\begin{tabular}{@{}ll r c c c@{}}
\toprule
\textbf{Type} & \textbf{Tier} & $n$ & \textbf{Gemma-3-1B} & \textbf{Gemma-3-4B} & \textbf{Gemma-3-12B} \\
\midrule
PK-A & Low & 183 & 0.643\,\scriptsize$\pm$0.255 & \textbf{0.753}\,\scriptsize$\pm$0.133 & 0.669\,\scriptsize$\pm$0.108 \\
PK-A & Mid & 250 & 0.689\,\scriptsize$\pm$0.235 & \textbf{0.760}\,\scriptsize$\pm$0.121 & 0.733\,\scriptsize$\pm$0.089 \\
PK-A & High & 255 & 0.709\,\scriptsize$\pm$0.212 & \textbf{0.799}\,\scriptsize$\pm$0.050 & 0.776\,\scriptsize$\pm$0.072 \\
\hdashline
PK-B & Low & 157 & 0.262\,\scriptsize$\pm$0.126 & \textbf{0.275}\,\scriptsize$\pm$0.095 & 0.236\,\scriptsize$\pm$0.060 \\
PK-B & Mid & 129 & \textbf{0.380}\,\scriptsize$\pm$0.119 & 0.358\,\scriptsize$\pm$0.110 & 0.367\,\scriptsize$\pm$0.067 \\
PK-B & High & 95 & \textbf{0.444}\,\scriptsize$\pm$0.142 & 0.378\,\scriptsize$\pm$0.128 & 0.407\,\scriptsize$\pm$0.070 \\
\hdashline
PD-A & Low & 15 & 0.876\,\scriptsize$\pm$0.141 & \textbf{0.984}\,\scriptsize$\pm$0.027 & 0.966\,\scriptsize$\pm$0.025 \\
PD-A & Mid & 17 & 0.848\,\scriptsize$\pm$0.073 & 0.795\,\scriptsize$\pm$0.113 & \textbf{0.874}\,\scriptsize$\pm$0.058 \\
PD-A & High & 45 & 0.907\,\scriptsize$\pm$0.040 & 0.903\,\scriptsize$\pm$0.048 & \textbf{0.966}\,\scriptsize$\pm$0.025 \\
\hdashline
PD-B & Low & 242 & 0.490\,\scriptsize$\pm$0.073 & 0.552\,\scriptsize$\pm$0.091 & \textbf{0.575}\,\scriptsize$\pm$0.083 \\
PD-B & Mid & 251 & 0.514\,\scriptsize$\pm$0.024 & 0.569\,\scriptsize$\pm$0.121 & \textbf{0.628}\,\scriptsize$\pm$0.077 \\
PD-B & High & 271 & 0.622\,\scriptsize$\pm$0.059 & 0.645\,\scriptsize$\pm$0.168 & \textbf{0.696}\,\scriptsize$\pm$0.079 \\
\bottomrule
\end{tabular}%
\end{table}

\paragraph{Cross-Axis Observations.}
Three observations emerge across the three tables. 
First, the absolute level of Recall differs sharply across mechanism subtypes regardless of similarity tier. PK-A reaches near-saturation for the leading LLM (Llama-3.2-1B at $0.84$--$0.91$ from Low to High) while PK-B caps below $0.52$ even at High similarity. The PK-A vs.\ PK-B gap therefore reflects the availability of mediating-entity evidence rather than structural overlap, consistent with the \cref{sec:stratified-analysis} view that A-B selectivity reflects access-versus-availability of supporting evidence.
Second, by combined AUC and Recall (\cref{sec:model-diversity} and Appendix~\ref{app-D.4:full-results}), Llama-3.2-1B is the strongest LLM and Qwen2.5-14B is the second-strongest, and this ranking transfers to the mech-similarity grid. Llama-3.2-1B leads PK-A across all three tiers and ties Qwen2.5-14B for PK-B High at $0.511$.
This rules out raw scale as the primary driver and points to fine-tuning data exposure to mediating-entity prompts.
Third, PD-B High receives the strongest baseline performance from HDN-DDI and TextDDI (\cref{tab:app-mech-sim-baselines}), both of which exceed every LLM in that cell.
PD-B (no shared target) is precisely the regime where pretrained pharmacological text and dense KG message-passing carry the most weight, and where the mediating-entity advantage of LLM-FT is least informative.

\subsection{Memorization Control: Synonym Robustness}
\label{app-D.6:synonym}

We test whether the fine-tuned LLM's prediction on a test pair is driven by the chemical identity of the two drugs or by the specific surface string of their canonical DrugBank names.
If the model has internalised drug identity, replacing each canonical name with one of its DrugBank synonyms (a different surface form of the same drug) should leave the prediction nearly unchanged. 
A large drop, by contrast, would indicate that the canonical string is acting as a memorization key beyond what the rest of the prompt encodes.

\paragraph{Protocol.}
For every test pair on the 800-subset S2 split, we render up to $K_{\text{eff}}=3$ synonym variants by replacing both \texttt{\{a\_name\}} and \texttt{\{b\_name\}} with the $i$-th DrugBank synonym of each drug (deterministic, alphabetical order).
All other prompt content is held tied to the original drug IDs. 
In particular, the P4 KG facts (targets, enzymes, transporters, carriers, pathways) are not perturbed. Only the cosmetic name strings change.
Eligible synonyms are extracted from the top-level \texttt{<drug><synonyms><synonym>} list of the DrugBank XML and filtered by seven rules:
\begin{enumerate}[leftmargin=*,itemsep=1pt,topsep=1pt,parsep=0pt,label={(\arabic*)}]
    \item English-readable
    \item length $\in [4, 60]$
    \item ASCII letters / digits / spaces / hyphen only
    \item lower-cased synonym distinct from canonical
    \item character-bigram Jaccard $<0.6$ vs.\ canonical (excludes hyphenation and casing variants)
    \item no leading digit or single letter (excludes CAS numbers and abbreviations)
    \item no $\text{\texttrademark}$ / $\text{\textregistered}$ / $\text{\textcopyright}$ characters.
\end{enumerate}
A pair is included iff both drugs have at least one eligible synonym; this yields $n=556$ pairs out of the original S2 test set. We evaluate two parallel setups, each at three random seeds:

\begin{itemize}[leftmargin=*,itemsep=1pt,topsep=1pt,parsep=0pt]
    \item \textbf{FT-on-P1.} Llama-3.2-1B with the per-seed best-S2 LoRA trained on P1 (Zero-Shot), scored under P1 (zero-shot, name-only). The synonym variant differs from the canonical only in two name tokens.
    
    \item \textbf{FT-on-P4.} Llama-3.2-1B with the per-seed best-S2 LoRA trained on P4 (One-Hop KG sequence), scored under P4 (one-hop KG, sequence). The KG context anchors the prompt to drug identity through entity strings tied to the original DB IDs.
\end{itemize}

All five reported metrics in \cref{tab:synonym-robustness} are computed at the pair level and averaged across the three seeds.
We report the agreement rate $\mathbb{P}\{\hat y(\text{canonical}) = \hat y(\text{synonym})\}$, the mean $|\Delta p_{\text{Yes}}|$ between canonical and synonym, the Pearson correlation between the two $p_{\text{Yes}}$ scalars, and the discriminative AUC under each naming. 

\begin{table}[t]
\centering
\small
\setlength{\tabcolsep}{4pt}
\caption{Synonym robustness on 800-subset S2 ($n=556$ pairs). Mean$\pm$std over 3 seeds. Higher agreement / Pearson and lower $|\Delta p|$ indicate name-robustness.}
\label{tab:synonym-robustness}
\resizebox{\textwidth}{!}{%
\begin{tabular}{l c c c c c c}
\toprule
\textbf{Setup} & $N$ & \textbf{Agreement} & $|\Delta p_{\text{Yes}}|$ & \textbf{Pearson $r$} & \textbf{AUC$_\text{canon}$} & \textbf{AUC$_\text{syn}$} \\
\midrule
P1 (zero-shot)     & 556 & $0.730\pm0.037$ & $0.229\pm0.011$ & $0.399\pm0.039$ & $0.692\pm0.008$ & $0.638\pm0.033$ \\
P4 (one-hop KG) & 556 & $0.846\pm0.003$ & $0.125\pm0.015$ & $0.822\pm0.016$ & $0.771\pm0.038$ & $0.753\pm0.036$ \\
\bottomrule
\end{tabular}%
}
\end{table}

\paragraph{Findings.}
The FT-on-P4 model is markedly more name-robust than the FT-on-P1 model. 
Agreement rises from $0.730$ to $0.846$, the mean canonical/synonym gap $|\Delta p_{\text{Yes}}|$ falls from $0.229$ to $0.125$, and the Pearson correlation between canonical and synonym $p_{\text{Yes}}$ more than doubles, from $0.399$ to $0.822$.
The discriminative AUC under synonym names also tracks the canonical AUC much more tightly under P4 than under P1 (drop of $0.018$ versus $0.054$). 
The pattern is consistent with the KG context anchoring the prediction to the underlying drug identity, so that swapping the surface name string has little effect on the decision. 
Under P1, where the prompt carries no information beyond the two names, the model has no other anchor and the synonym substitution materially shifts predictions on roughly $27\%$ of pairs. 
Together, the two rows indicate that the cold-start performance of the fine-tuned P4 model is not driven by memorization of the drug name strings themselves.

\subsection{Memorization Control: Temporal Analysis}
\label{app-D.7:temporal-control}


We start from the \emph{pretraining-memorization hypothesis}. If pretraining-side memorization drives cold-start performance, test pairs containing older, more thoroughly documented drugs should be easier than those containing newer drugs, because older drugs have a larger footprint in the LLM's pretraining corpus.
We design two complementary analyses to test it, a post-hoc stratification of existing S2 results by FDA approval year (\cref{app-D.7.1:temporal-posthoc}) and a temporal cold-start split where the train/test partition is governed by drug approval date (\cref{app-D.7.2:temporal-split}).
Both use Llama-3.2-1B under P4 to match the main-paper configuration.

\subsubsection{Post-Hoc Stratification by FDA Approval Year}
\label{app-D.7.1:temporal-posthoc}

We stratify the existing S2 test-set predictions of Llama-3.2-1B (P4) on the 800-drug subset by each test pair's earlier-approved drug FDA year (the more recent drug determines the cold-start identity, but the earlier-approved drug carries the longer pretraining footprint, so we anchor each pair on the older drug). 
Earliest approval years are extracted from the DrugBank \texttt{products.start\_dates} field, which yields a year for $680$ of the $800$ drugs ($85.0\%$).
Concretely, each test pair $(a, b)$ is assigned the bucket of $\min(\text{year}_a, \text{year}_b)$. 
A pair where the two drugs sit in different brackets is therefore placed in the bucket of the older one (e.g.\ a 1985--2015 pair goes to Pre-2000), and a pair in which either drug has no parseable approval year is bucketed as \emph{Unknown}. 
Of the 800 drugs, 120 lack any parseable date, which is why the Unknown row of \cref{tab:fda-year-strat} carries a non-trivial pair count even though every other bucket is defined drug-side by year.

\begin{table}[h]
\centering
\caption{Llama-3.2-1B (P4) S2 Recall and AUC, stratified by the older drug's earliest FDA approval year (3-seed mean $\pm$ std on the 800-drug subset). \emph{Drugs} is the per-bucket drug count on the 800-subset (sums to $800$). $\bar n$ is the per-bucket positive pair count averaged across seeds (sums to the S2 positive count).}
\label{tab:fda-year-strat}
\small
\setlength{\tabcolsep}{6pt}
\renewcommand{\arraystretch}{1.1}
\begin{tabular}{@{}l c c cc@{}}
\toprule
\textbf{Approval bucket} & \textbf{Drugs} & \textbf{$\bar n$ pos} & \textbf{Recall} & \textbf{AUC} \\
\midrule
Pre-2000   & 388 & 1240 & $0.687 \pm 0.027$ & $0.768 \pm 0.021$ \\
2000--2009 & 103 &  138 & $0.801 \pm 0.051$ & $0.843 \pm 0.033$ \\
2010--2019 & 127 &  134 & $0.864 \pm 0.129$ & $0.802 \pm 0.079$ \\
2020+      &  62 &   20 & $0.968 \pm 0.055$ & $0.813 \pm 0.084$ \\
Unknown    & 120 &  381 & $0.543 \pm 0.190$ & $0.735 \pm 0.059$ \\
\midrule
ALL        & 800 & 1913 & $0.697 \pm 0.021$ & $0.776 \pm 0.020$ \\
\bottomrule
\end{tabular}
\end{table}

We initially expected, under the pretraining-memorization hypothesis, that older drugs would show higher Recall, since their associated literature accumulates over time.
The observed pattern is the opposite. 
Recall rises monotonically from Pre-2000 ($0.687$) through 2000--2009 ($0.801$) and 2010--2019 ($0.864$) to 2020+ ($0.968$), with Cohen's $d = -6.5$ between Pre-2000 and 2020+ (in the wrong direction for memorization).
Two factors plausibly explain this reverse trend. First, drugs approved more recently appear with more complete KG annotations in DrugBank (more enzymes, transporters, and mediating entities listed), which directly strengthens the channel that drives correct predictions in \cref{sec:modality-results}. 
Second, pairs containing only Pre-2000 drugs are over-represented among the difficult cases that require fully novel mediating paths, since the older subnetwork has been mined more exhaustively in prior interaction databases. 
The Unknown bucket's low Recall ($0.543$) is consistent with this reading. Drugs without parseable approval dates are largely veterinary, withdrawn, or investigational compounds whose KG entries are sparse, removing the mediating signal entirely. 
The observed Recall-year ordering runs counter to the older-drug advantage expected under the pretraining-memorization hypothesis.

\subsubsection{Temporal Cold-Start Split}
\label{app-D.7.2:temporal-split}

\paragraph{Setup.}
We complement the post-hoc stratification with a hard temporal split, where the train/test partition is governed entirely by drug FDA approval year. 
For each cutoff $Y \in \{1985, 1995, 2005, 2010\}$, drugs approved \emph{before} $Y$ form the training pool $G_1$ and drugs approved \emph{in or after} $Y$ form the held-out cold-start pool $G_2$.
The training set is the set of pairs internal to $G_1$ and the test set is the set of pairs internal to $G_2$ (both drugs cold).
The four cutoffs span four FDA regulatory eras, namely Hatch-Waxman (1985, extreme temporal shift), ICH harmonisation (1995, near-balanced split), the targeted-therapy era (2005), and the immune-checkpoint era (2010, smallest shift).
120 of the 800 random-cold-start drugs lack a parseable FDA approval year and are excluded, leaving $|G_1| + |G_2| = 680$ at every cutoff.
Llama-3.2-1B is fine-tuned under P4 (One-Hop KG sequence) separately for each cutoff at three random seeds.

\begin{table}[h]
\centering
\caption{Llama-3.2-1B (P4) on the temporal cold-start splits (3-seed mean$\pm$std). $|G_1| / |G_2|$ is the train / cold-pool drug count after filtering out drugs without a parseable approval year.}
\label{tab:temporal-split}
\small
\setlength{\tabcolsep}{6pt}
\renewcommand{\arraystretch}{1.1}
\begin{tabular}{@{}c c c c c@{}}
\toprule
\textbf{Cutoff} & $|G_1| / |G_2|$ & $n_{\text{test}}$ & \textbf{AUROC} & \textbf{F1} \\
\midrule
1985 & 237 / 443 & 31{,}358 & $0.687 \pm 0.021$ & $0.641 \pm 0.013$ \\
1995 & 335 / 345 & 19{,}106 & $0.717 \pm 0.019$ & $0.671 \pm 0.024$ \\
2005 & 436 / 244 &  9{,}396 & $0.753 \pm 0.020$ & $0.703 \pm 0.015$ \\
2010 & 491 / 189 &  5{,}956 & $0.760 \pm 0.011$ & $0.730 \pm 0.016$ \\
\bottomrule
\end{tabular}
\end{table}

\paragraph{Findings.}
AUROC rises monotonically with the cutoff year, from $0.687$ at 1985 to $0.760$ at 2010, and F1 follows the same ordering.
The 2010 cutoff approaches the random-cold-start S2 AUROC of approximately $0.77$ reported in \cref{sec:cold-start-results}, indicating that the temporal split converges toward the random-drug split in difficulty as the temporal gap between train and test pools shrinks. 
The ranking is consistent with a non-memorization factor. The size of the training pool ($|G_1|$ rises from $237$ to $491$ across the sweep) gives the model more training signal.
The pretraining-memorization hypothesis predicts the opposite ordering, since under that hypothesis test pairs with older drugs should be easier on account of their larger pretraining presence.
The monotonic increase therefore reinforces the conclusion of \cref{app-D.7.1:temporal-posthoc} that pretraining-side memorization is not the primary driver of cold-start performance.

\subsection{Assessing Pretraining Memorization}
\label{app-D.8:pretraining-memorization}

We assess whether direct recall of pretrained DDI facts explains LLM-FT's cold-start performance.
This assessment draws on three behavioral comparisons:

(i) In the approval-year stratification, newer-drug pairs achieve higher Recall than older-drug pairs (\cref{tab:fda-year-strat}), opposite to the expected advantage from older drugs' greater pretraining exposure.
(ii) Without fine-tuning, open-weight LLMs under P4 obtain near-random S2 AUC-ROC (\cref{tab:app-e1-auc}), indicating that pretrained DDI knowledge is not reliably expressed in direct inference.
(iii) Input masking further identifies the evidence supporting fine-tuned predictions: removing drug names has little effect on PK-A, whereas masking KG entities causes a larger decrease (\cref{tab:masking-effects}).
Together, these results suggest that the observed performance is better explained by responsiveness to supplied mediator evidence than by direct recall of drug-pair identities.

\section{Diagnostic Analysis Details}
\label{app-E:diagnostic}

\subsection{Formal KPS Definitions}
\label{app-E.1:diagnostic-metrics}

This appendix gives the operational definitions behind \cref{tab:kps-summary} and the per-method tables in \cref{sec:modality-results}.
All indicators are pair-level absolute prediction-probability differences aggregated by mechanism bucket.
The KPS-derived numbers (\cref{app-E.2:masking-full,app-E.3:kps-full}) are 3-seed means~$\pm$~standard deviation on the 800-drug S2 test set.
The attention-level analyses in \cref{app-E.4:attention} use a single random seed unless otherwise stated, and \cref{app-E.4.4:attn-stability} extends the intervention experiments to three seeds and two architectures.

\paragraph{Drug-Swap Candidate Construction.}
For every positive S2 pair $(u, v)$ with both drugs in $G_2$ (cold), we collect every $u' \in G_2 \setminus \{u\}$ such that (i)~$(u', v)$ is observed in S1 or S2 with $\mathrm{label}(u', v) \neq \mathrm{label}(u, v)$, and (ii)~$u'$ is itself a cold drug. The resulting triples table $\mathcal{C} = \{(u, v, u')\}$ contains $63.5$K--$92.5$K triples per seed (each $(u, v)$ contributes $\sim$20 candidates on average).

\paragraph{KPS-F (Drug-Replacement Sensitivity).}
For each method $m$,
\[
  \mathrm{KPS\text{-}F}_m(b) \;=\; \tfrac{1}{|\mathcal{C}_b|}\sum_{(u,v,u')\in\mathcal{C}_b}\bigl|P_m(u,v) - P_m(u',v)\bigr|,
\]
where $\mathcal{C}_b \subseteq \mathcal{C}$ is the subset whose anchor pair $(u, v)$ falls in mechanism bucket $b\!\in\!\{$PK-A, PK-B, PD-A, PD-B$\}$, and $P_m(\cdot,\cdot)$ is method $m$'s S2-test predicted probability for the positive class. The ALL bucket averages over all positive $(u, v)$. KPS-F is a \emph{per-method-shared} metric. Every baseline and LLM-FT consume the same $\mathcal{C}$, so cross-method comparisons are directly meaningful.

\paragraph{KPS-Channel (Channel-Mask Sensitivity).}
For multi-channel methods (LLM-FT, MKG-FENN, TIGER), we measure how predictions change when one input channel is symmetrically blanked for \emph{both} drugs in the pair. Using the swap-table anchors $(u, v)$:
\[
  \mathrm{KPS\text{-}c}_m(b) \;=\; \tfrac{1}{|\mathcal{T}_b|}\sum_{(u,v)\in\mathcal{T}_b}\bigl|P_m^{\text{base}}(u, v) - P_m^{\text{mask}=c}(u, v)\bigr|,
\]
where $c\!\in\!\{\text{Name},\text{KG},\text{mol}\}$ and $\mathcal{T}_b$ is the bucket-restricted set of $(u, v)$ anchors. The masking operation is method-specific.

\begin{itemize}[leftmargin=*,itemsep=1pt,topsep=2pt]
\item \textbf{LLM-FT.} Substitute prompt placeholders. \mbox{KPS-Name} masks both drug names ($P_{R_1}$ vs.\ $P_{R_0}$ in \cref{tab:masking-conditions}). \mbox{KPS-KG} replaces all KG entity strings with \texttt{[ENTITY]} ($P_{R_2}$ vs.\ $P_{R_0}$).
\item \textbf{MKG-FENN.} Zero the four GNN output embeddings before the FusionLayer. \mbox{KPS-KG} zeros $g_1$ (entity) and $g_3$ (DDI topology) for both $u$ and $v$. \mbox{KPS-mol} zeros $g_2$ (substructure) and $g_4$ (RDKit property).
\item \textbf{TIGER.} Zero the dual-channel embeddings before the fusion FC. \mbox{KPS-KG} zeros $\texttt{drug}_{1,2}\texttt{\_node\_emb}$. \mbox{KPS-mol} zeros $\texttt{mol}_{1,2}\texttt{\_emb}$.
\end{itemize}
For all three methods the mask is symmetric (both $u$ and $v$ are blanked), so KPS-Channel measures the marginal prediction sensitivity to the entire channel rather than a single-side perturbation. Single-channel baselines (DeepDDI, SSI-DDI, DSN-DDI, HDN-DDI, EmerGNN, TextDDI) admit no meaningful channel-mask, so we report only KPS-F for them.

\paragraph{KSAI (Channel-Interaction Asymmetry).}
For LLM-FT, the four R0--R3 mask conditions form a $2\!\times\!2$ factorial (name on/off $\times$ KG on/off). KSAI quantifies whether the KG-channel sensitivity changes when the name channel is also blanked:
\[
  \mathrm{KSAI}_m(b) \;=\; \mathrm{KPS\text{-}KG}_m^{\text{name-masked}}(b) - \mathrm{KPS\text{-}KG}_m^{\text{name-present}}(b),
\]
i.e.,\ the per-pair difference $|P_{R_1} - P_{R_3}| - |P_{R_0} - P_{R_2}|$ averaged within each bucket. Positive KSAI indicates the model becomes \emph{more} sensitive to KG when the name is removed, i.e.,\ the name channel partially compensates for KG when both are present. We report KSAI only for LLM-FT. MKG-FENN and TIGER expose no separable name channel that can be masked independently of the mol or KG channel, so R1 / R3 have no architectural analogue.

\subsection{Masking Experiment: Full Factorial Decomposition}
\label{app-E.2:masking-full}

\begin{figure}[h]
  \centering
  \includegraphics[width=0.85\textwidth]{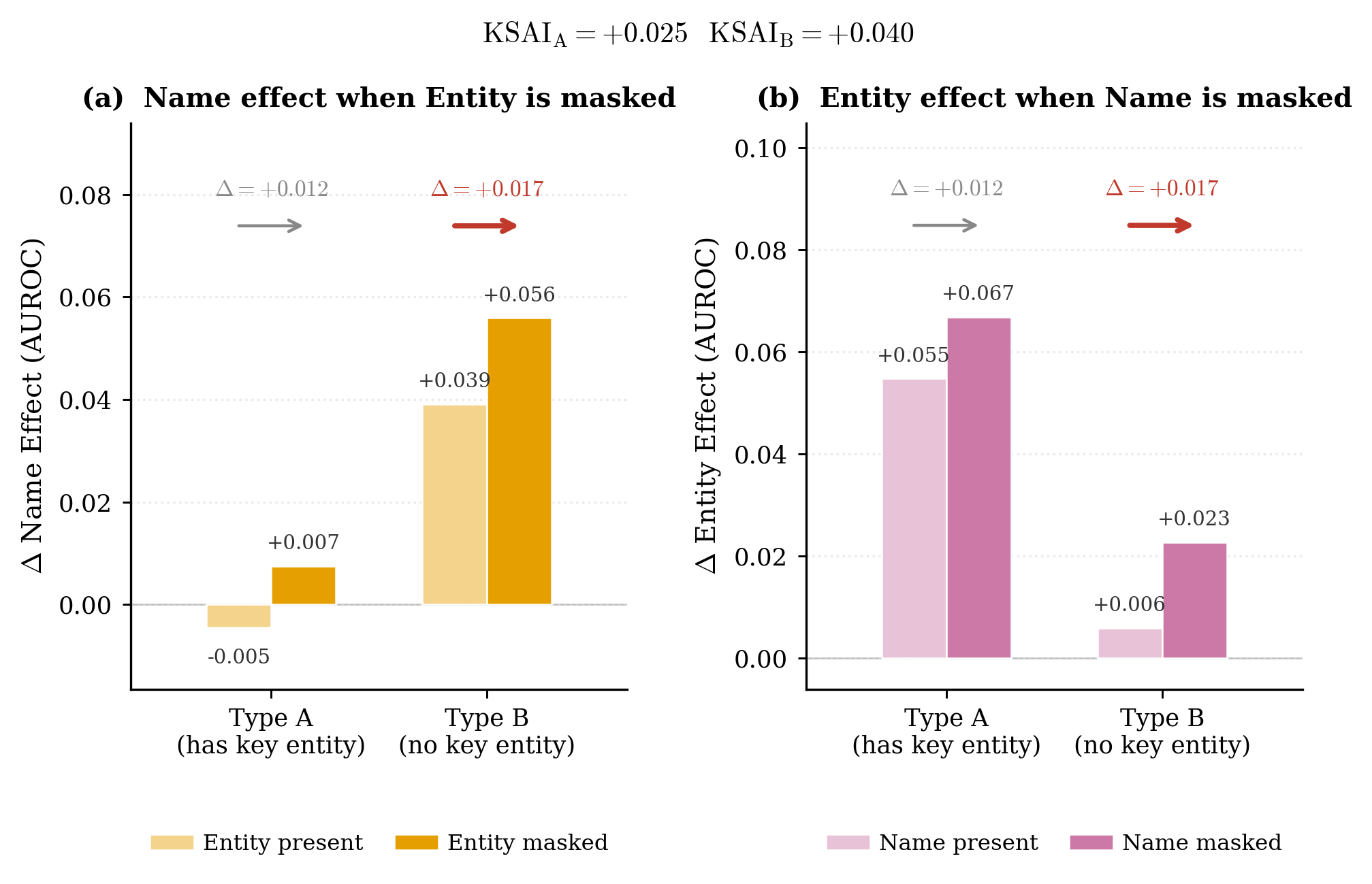}
   \caption{Channel interaction on LLM-FT (800-drug, 3-seed mean, Top-3 KG / R0--R3). Each pair shows one channel's marginal masking effect with the other channel present (light) vs.\ already masked (dark). Type~A. Modest interaction (KSAI averaged over PK-A and PD-A $=0.025$). Type~B. Clear cross-channel \emph{compensation}. Each channel's marginal effect is larger when the other is already masked (KSAI averaged over PK-B and PD-B $=0.040$). Per-bucket KSAI values are reported in \cref{tab:appf-llmft-channels}. Referenced from \cref{sec:modality-results}.}
  \label{fig:channel-interaction}
\end{figure}

This appendix complements \cref{tab:masking-effects} and \cref{fig:channel-interaction} with the full eight-cell factorial under both Top-3 KG and Full KG (no top-$k$ truncation, $\max\_token\_length = 4096$). All numbers are 3-seed means on the 800-drug subset, S2 test set. 
The eight conditions are summarized in \cref{tab:masking-conditions}. Across all conditions the $B_3$ method instruction (the P4 verbatim text in Appendix~\ref{app-D.1:prompts}) is held fixed and only $B_6$ (drug names and key entity strings) is rewritten.
Therefore, the model is never told that masking has occurred and behavioural differences across R0--R7 reflect the controlled contribution of each channel rather than instruction-level priming.

\begin{table}[h]
\centering
\caption{Eight masking conditions used in the \cref{sec:modality-results} factorial. R0--R3 retrieve at most three KG entities per type per drug (Top-3). R4--R7 disable both top-$k$ truncation and the prompt-token cap, so every annotated entity is presented to the model.}
\label{tab:masking-conditions}
\small
\setlength{\tabcolsep}{4pt}
\renewcommand{\arraystretch}{1.1}
\begin{tabular}{@{}clcccc@{}}
\toprule
\textbf{ID} & \textbf{Method directory} & \textbf{KG} & \textbf{Drug name} & \textbf{Key entity} & \textbf{Token cap} \\
\midrule
R0 & One\_Hop\_KG\_Sequence            & Top-3 & real        & real        & 1250 \\
R1 & OHKS\_Mask\_Name                  & Top-3 & \texttt{[DRUG\_*]} & real        & 1250 \\
R2 & OHKS\_Mask\_Entity                & Top-3 & real        & \texttt{[ENTITY]}   & 1250 \\
R3 & OHKS\_Mask\_Name\_Entity          & Top-3 & \texttt{[DRUG\_*]} & \texttt{[ENTITY]}   & 1250 \\
\midrule
R4 & OHKS\_Full                        & Full  & real        & real        & 4096 \\
R5 & OHKS\_Full\_Mask\_Name            & Full  & \texttt{[DRUG\_*]} & real        & 4096 \\
R6 & OHKS\_Full\_Mask\_Entity          & Full  & real        & \texttt{[ENTITY]}   & 4096 \\
R7 & OHKS\_Full\_Mask\_Name\_Entity    & Full  & \texttt{[DRUG\_*]} & \texttt{[ENTITY]}   & 4096 \\
\bottomrule
\end{tabular}
\end{table}

\paragraph{Per-Bucket AUC Under All 8 Conditions.}
\cref{tab:app-f2-full-factorial} reports 3-seed-mean AUC for every (bucket, condition) cell. Values match the main-paper Top-3 columns (R0--R3) and add the four Full-KG columns (R4--R7).

\begin{table}[h]
\centering
\caption{AUC-ROC by mechanism bucket × masking condition. 3-seed mean~$\pm$~std on the 800-drug subset, S2.}
\label{tab:app-f2-full-factorial}
\scriptsize
\setlength{\tabcolsep}{3pt}
\renewcommand{\arraystretch}{1.1}
\resizebox{\textwidth}{!}{%
\begin{tabular}{@{}lcccc cccc@{}}
\toprule
\textbf{Bucket} & \multicolumn{4}{c}{\textbf{Top-3 KG}} & \multicolumn{4}{c}{\textbf{Full KG}} \\
\cmidrule(lr){2-5}\cmidrule(lr){6-9}
 & R0 & R1 & R2 & R3 & R4 & R5 & R6 & R7 \\
\midrule
PK-A & $0.878_{\pm0.067}$ & $0.893_{\pm0.055}$ & $0.798_{\pm0.078}$ & $0.808_{\pm0.074}$ & $0.890_{\pm0.051}$ & $0.902_{\pm0.044}$ & $0.821_{\pm0.060}$ & $0.829_{\pm0.058}$ \\
PK-B & $0.602_{\pm0.079}$ & $0.557_{\pm0.081}$ & $0.606_{\pm0.069}$ & $0.553_{\pm0.060}$ & $0.608_{\pm0.088}$ & $0.564_{\pm0.090}$ & $0.612_{\pm0.083}$ & $0.556_{\pm0.086}$ \\
PD-A & $0.936_{\pm0.006}$ & $0.930_{\pm0.008}$ & $0.907_{\pm0.028}$ & $0.882_{\pm0.041}$ & $0.938_{\pm0.007}$ & $0.935_{\pm0.008}$ & $0.923_{\pm0.012}$ & $0.911_{\pm0.020}$ \\
PD-B & $0.759_{\pm0.043}$ & $0.725_{\pm0.035}$ & $0.743_{\pm0.041}$ & $0.684_{\pm0.026}$ & $0.758_{\pm0.043}$ & $0.729_{\pm0.032}$ & $0.740_{\pm0.041}$ & $0.689_{\pm0.021}$ \\
\midrule
ALL  & $0.776_{\pm0.020}$ & $0.759_{\pm0.014}$ & $0.739_{\pm0.027}$ & $0.708_{\pm0.019}$ & $0.781_{\pm0.021}$ & $0.766_{\pm0.016}$ & $0.749_{\pm0.034}$ & $0.720_{\pm0.026}$ \\
\bottomrule
\end{tabular}%
}
\end{table}

\paragraph{KG Completeness Effect (Top-3 $\rightarrow$ Full).}
Comparing R4 against R0 isolates the joint effect of (i) uncapping KG retrieval and (ii) extending the prompt token budget, with no masking on either condition. The 3-seed-mean uncap effect is small but uniformly positive, with $\Delta\mathrm{AUC} \times 10^2 = +0.67$ on Type~A, $+0.30$ on Type~B, and $+0.59$ overall. Type~A benefits roughly $2\times$ more than Type~B from the additional entities, consistent with the anchor-based paradigm in \cref{sec:modality-results}. A richer entity context provides more candidate anchors when a key mediating entity exists, but offers little to pairs whose interaction must be inferred from name-activated prior knowledge.

\paragraph{Mask Asymmetry Is Preserved Under Full KG.}
The factorial decomposition (R4--R7) reproduces the channel-dominance pattern of R0--R3. Type~A is entity-dominant under Full KG (PK-A entity $-6.8$ / name $+1.2$, PD-A entity $-1.5$ / name $-0.3$, all $\times 10^2$). Type~B is name-dominant (PK-B name $-4.4$ / entity $+0.3$, PD-B name $-2.9$ / entity $-1.8$, all $\times 10^2$). Sign and ranking of every channel effect match the Top-3 column in \cref{tab:masking-effects}. Only PK-A's name effect remains positive under both Top-3 and Full KG, suggesting drug-name information can interfere with the use of explicit KG entities in PK-A prediction. The cross-channel asymmetry under Full KG follows the same Type~B $>$ Type~A ordering observed for KSAI under Top-3 KG. Adding KG entities therefore neither suppresses nor amplifies the masking-driven channel selection observed in the main paper. The A/B channel-dominance pattern is a robust property of LLM-FT independent of the KG-truncation budget.

\subsection{Full KPS Tables}
\label{app-E.3:kps-full}

\paragraph{Per-Bucket KPS-F Across All Nine Methods.}
\cref{tab:appf-kpsf-full} expands \cref{tab:kps-summary} to per-bucket means~$\pm$~standard deviations. 
The bridging-entity selectivity called out in the main text materializes as follows. EmerGNN's PK-A KPS-F ($0.486$) is $1.65\!\times$ its PK-B ($0.294$), and LLM-FT shows a similar PK-A/PK-B ratio of $1.49\!\times$. Every other method falls within $[0.98, 1.23]$, and their A--B gap stays at most $+0.052$, well separated from EmerGNN's $+0.170$ and LLM-FT's $+0.145$.

\begin{table}[h]
\centering
\caption{KPS-F per (method, bucket), 3-seed mean~$\pm$~std. Bold marks the two methods with substantial A--B gaps. Sample counts $|\mathcal{C}_b|$ aggregated across seeds. ALL $\!\approx\!230$K, PK-A $\!\approx\!81$K, PK-B $\!\approx\!47$K, PD-A $\!\approx\!8.4$K, PD-B $\!\approx\!97$K.}
\label{tab:appf-kpsf-full}
\small
\setlength{\tabcolsep}{4pt}
\renewcommand{\arraystretch}{1.05}
\begin{tabular}{@{}lcccccc@{}}
\toprule
\textbf{Method} & \textbf{PK-A} & \textbf{PK-B} & \textbf{PD-A} & \textbf{PD-B} & \textbf{ALL} & \textbf{A--B gap} \\
\midrule
DeepDDI    & $0.333_{\pm0.081}$ & $0.282_{\pm0.065}$ & $0.371_{\pm0.070}$ & $0.318_{\pm0.056}$ & $0.317_{\pm0.066}$ & $+0.052_{\pm0.030}$ \\
SSI-DDI    & $0.166_{\pm0.058}$ & $0.150_{\pm0.060}$ & $0.195_{\pm0.037}$ & $0.166_{\pm0.055}$ & $0.164_{\pm0.056}$ & $+0.022_{\pm0.014}$ \\
DSN-DDI    & $0.420_{\pm0.068}$ & $0.384_{\pm0.074}$ & $0.451_{\pm0.050}$ & $0.416_{\pm0.063}$ & $0.412_{\pm0.065}$ & $+0.035_{\pm0.021}$ \\
HDN-DDI    & $0.164_{\pm0.047}$ & $0.133_{\pm0.042}$ & $0.186_{\pm0.056}$ & $0.153_{\pm0.040}$ & $0.154_{\pm0.043}$ & $+0.032_{\pm0.010}$ \\
\textbf{EmerGNN} & $\boldsymbol{0.486_{\pm0.019}}$ & $\boldsymbol{0.294_{\pm0.012}}$ & $\boldsymbol{0.492_{\pm0.087}}$ & $\boldsymbol{0.343_{\pm0.041}}$ & $\boldsymbol{0.388_{\pm0.016}}$ & $\boldsymbol{+0.170_{\pm0.057}}$ \\
TIGER      & $0.213_{\pm0.067}$ & $0.218_{\pm0.074}$ & $0.237_{\pm0.092}$ & $0.216_{\pm0.078}$ & $0.216_{\pm0.074}$ & $+0.008_{\pm0.018}$ \\
MKG-FENN   & $0.074_{\pm0.027}$ & $0.075_{\pm0.027}$ & $0.064_{\pm0.029}$ & $0.077_{\pm0.034}$ & $0.075_{\pm0.030}$ & $-0.007_{\pm0.003}$ \\
TextDDI    & $0.286_{\pm0.062}$ & $0.287_{\pm0.054}$ & $0.419_{\pm0.055}$ & $0.333_{\pm0.056}$ & $0.310_{\pm0.058}$ & $+0.042_{\pm0.007}$ \\
\textbf{LLM-FT} & $\boldsymbol{0.389_{\pm0.069}}$ & $\boldsymbol{0.261_{\pm0.036}}$ & $\boldsymbol{0.481_{\pm0.025}}$ & $\boldsymbol{0.320_{\pm0.021}}$ & $\boldsymbol{0.336_{\pm0.030}}$ & $\boldsymbol{+0.145_{\pm0.039}}$ \\
\bottomrule
\end{tabular}
\end{table}

\paragraph{Per-Bucket KPS-Channel + KSAI for LLM-FT.}
\cref{tab:appf-llmft-channels} reports the four mask-derived indicators on LLM-FT. The PK-A/PD-A vs.\ PK-B/PD-B asymmetry on KPS-KG (Named) and KPS-Name mirrors the masking dominance pattern of \cref{tab:masking-effects}. PK-A is entity-dominant ($\mathrm{KPS\text{-}KG} = 0.167 > \mathrm{KPS\text{-}Name} = 0.083$), and PD-B is name-dominant ($\mathrm{KPS\text{-}Name} = 0.129 > \mathrm{KPS\text{-}KG} = 0.121$).
KSAI is uniformly positive, indicating the name channel partially compensates for KG presence when both are available. The compensation grows as one moves from Type~A to Type~B (PK-A $0.012$ $\rightarrow$ PD-B $0.046$).

\begin{table}[h]
\centering
\caption{LLM-FT (Llama-3.2-1B, P4) per-bucket channel-mask indicators. KPS-KG and KPS-Name are mask-derived (R0/R1/R2/R3 of \cref{tab:masking-conditions}). KSAI-masking is the $2\!\times\!2$ factorial interaction $|P_{R_1}\!-\!P_{R_3}|\!-\!|P_{R_0}\!-\!P_{R_2}|$. 3-seed mean~$\pm$~std.}
\label{tab:appf-llmft-channels}
\small
\setlength{\tabcolsep}{4pt}
\renewcommand{\arraystretch}{1.05}
\begin{tabular}{@{}lccccc@{}}
\toprule
\textbf{Indicator} & \textbf{PK-A} & \textbf{PK-B} & \textbf{PD-A} & \textbf{PD-B} & \textbf{ALL} \\
\midrule
KPS-Name (KG present)   & $0.083_{\pm0.028}$ & $0.133_{\pm0.033}$ & $0.069_{\pm0.005}$ & $0.129_{\pm0.034}$ & $0.112_{\pm0.028}$ \\
KPS-KG   (Name present) & $0.167_{\pm0.022}$ & $0.106_{\pm0.009}$ & $0.090_{\pm0.036}$ & $0.121_{\pm0.009}$ & $0.134_{\pm0.009}$ \\
KPS-KG   (Name masked)  & $0.179_{\pm0.020}$ & $0.139_{\pm0.006}$ & $0.128_{\pm0.028}$ & $0.168_{\pm0.008}$ & $0.165_{\pm0.007}$ \\
KSAI-masking            & $0.012_{\pm0.010}$ & $0.033_{\pm0.005}$ & $0.038_{\pm0.008}$ & $0.046_{\pm0.004}$ & $0.031_{\pm0.004}$ \\
\bottomrule
\end{tabular}
\end{table}

\paragraph{Per-Bucket KPS-Channel for Multi-Channel Baselines.}
\cref{tab:appf-baseline-channels} reports KPS-mol and KPS-KG for MKG-FENN and TIGER, computed by symmetrically zeroing the relevant embeddings before the fusion layer (Appendix~\ref{app-E.1:diagnostic-metrics}).
Both methods are KG-dominant in absolute terms (KG/mol ratios $1.31\!\times$ for MKG-FENN, $2.11\!\times$ for TIGER). 
Crucially, both are \emph{near-uniform across buckets}. MKG-FENN's KPS-KG range is $[0.294, 0.321]$ ($1.09\!\times$ spread), and TIGER's is $[0.199, 0.217]$ ($1.09\!\times$).
This contrasts sharply with LLM-FT's KPS-KG bucket spread of $1.85\!\times$ (PK-A $0.167$ vs.\ PD-A $0.090$), confirming that LLM-FT dynamically compensates across knowledge sources whereas conventional baselines exhibit near-uniform sensitivity.

\begin{table}[h]
\centering
\caption{Channel-mask KPS for the two multi-channel baselines (MKG-FENN, TIGER) on the 800-drug subset, S2, 3-seed mean~$\pm$~std. Mask is symmetric (both drugs blanked). See Appendix~\ref{app-E.1:diagnostic-metrics} for the per-method embedding-zero protocol.}
\label{tab:appf-baseline-channels}
\small
\setlength{\tabcolsep}{4pt}
\renewcommand{\arraystretch}{1.05}
\begin{tabular}{@{}llccccc@{}}
\toprule
\textbf{Method} & \textbf{Channel} & \textbf{PK-A} & \textbf{PK-B} & \textbf{PD-A} & \textbf{PD-B} & \textbf{ALL} \\
\midrule
\multirow{2}{*}{MKG-FENN}
 & KPS-mol  & $0.234_{\pm0.016}$ & $0.242_{\pm0.019}$ & $0.132_{\pm0.023}$ & $0.246_{\pm0.035}$ & $0.235_{\pm0.011}$ \\
 & KPS-KG   & $0.294_{\pm0.045}$ & $0.317_{\pm0.056}$ & $0.297_{\pm0.067}$ & $0.321_{\pm0.039}$ & $0.309_{\pm0.040}$ \\
\midrule
\multirow{2}{*}{TIGER}
 & KPS-mol  & $0.102_{\pm0.046}$ & $0.096_{\pm0.040}$ & $0.103_{\pm0.058}$ & $0.099_{\pm0.045}$ & $0.099_{\pm0.044}$ \\
 & KPS-KG   & $0.209_{\pm0.068}$ & $0.217_{\pm0.088}$ & $0.199_{\pm0.075}$ & $0.208_{\pm0.075}$ & $0.209_{\pm0.076}$ \\
\bottomrule
\end{tabular}
\end{table}

\subsection{Attention-Level Channel Analysis}
\label{app-E.4:attention}

This appendix provides the full results behind the controlled attention intervention reported in \cref{tab:attn_intervention}. 
Unless otherwise stated, experiments use Llama-3.2-1B with the same LoRA fine-tuning protocol on the S2 test set.
For this model, attention weights are extracted from all 16~layers $\times$ 32~heads using eager attention.
Each input token is classified into one of seven classes (instruction template, drug~A/B name, drug~A/B SMILES, drug~A/B KG entities), and we record the attention from the answer token (last non-padding position) to all input tokens.

\subsubsection{Layer- and Head-Level Functional Specialization}
\label{app-E.4.1:attn-head}

\begin{figure}[h]
\centering
\begin{subfigure}{0.7\textwidth}
  \centering
  \includegraphics[width=\linewidth]{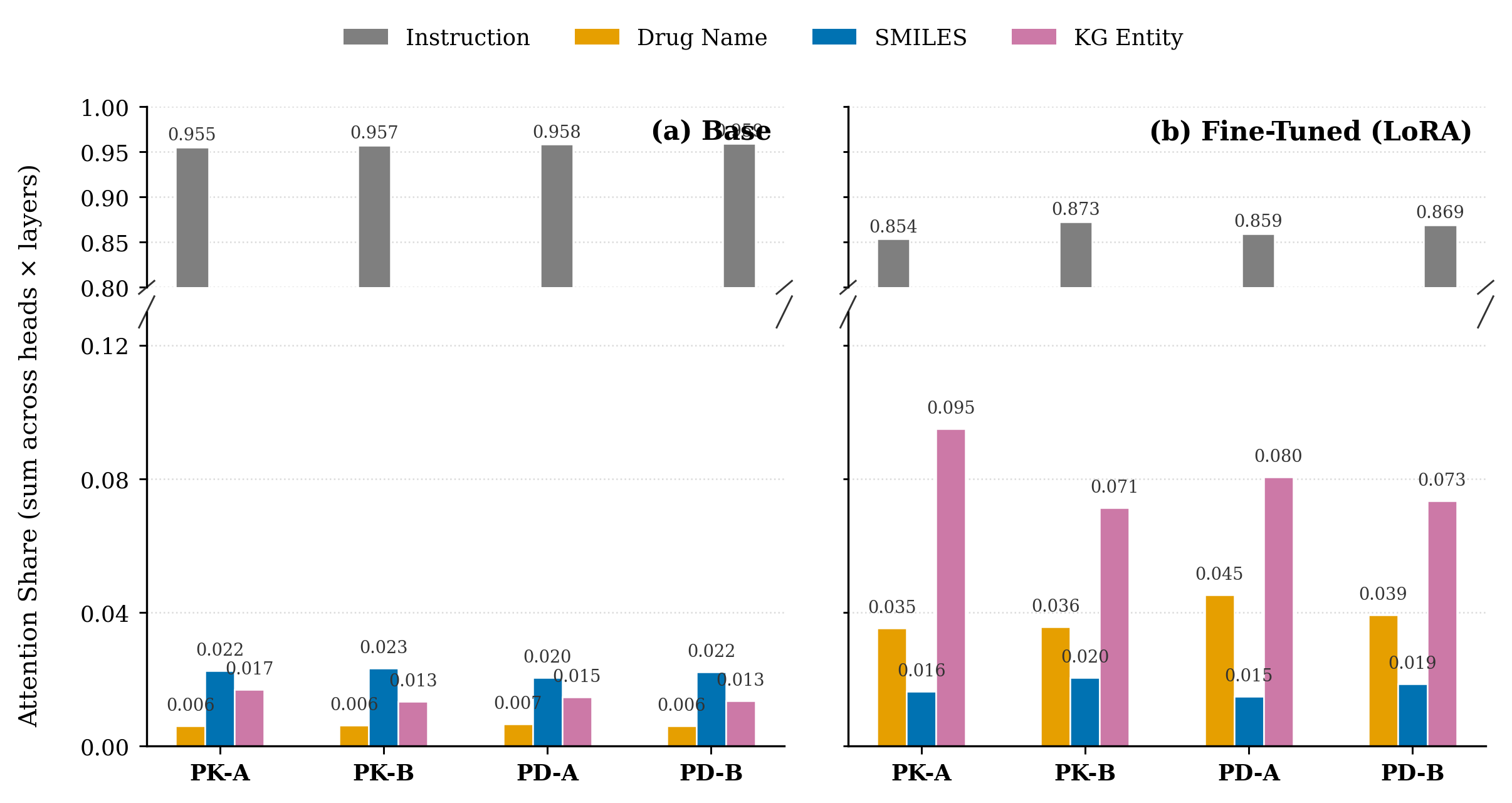}
  \caption{Token-class share, base vs.\ fine-tuned, broken down by mechanism bucket. Y-axis is the fraction of the answer-token attention probability mass falling on each token class, averaged over the bucket's positive pairs and pooled over all layers and heads.}
  \label{fig:attn-app-share}
\end{subfigure}\\[8pt]
\begin{subfigure}{0.7\textwidth}
  \centering
  \includegraphics[width=\linewidth]{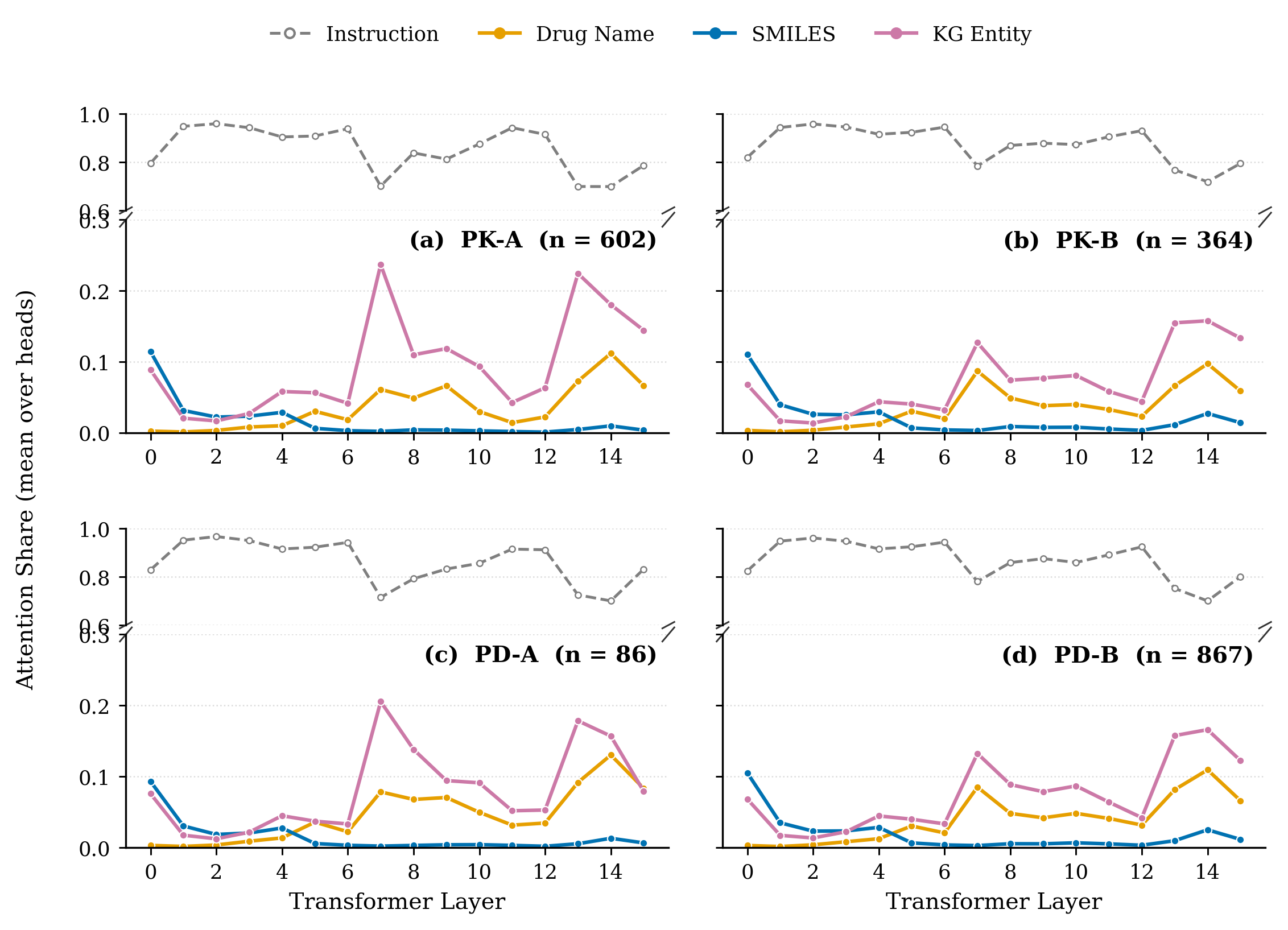}
   \caption{Layer-wise attention trajectory of the fine-tuned model, per mechanism bucket. Y-axis is the answer-token attention probability mass on KG-entity tokens at each transformer layer, averaged across heads and over the bucket's positive pairs.}
  \label{fig:attn-app-layer}
\end{subfigure}\\[8pt]
\begin{subfigure}{0.7\textwidth}
  \centering
  \includegraphics[width=\linewidth]{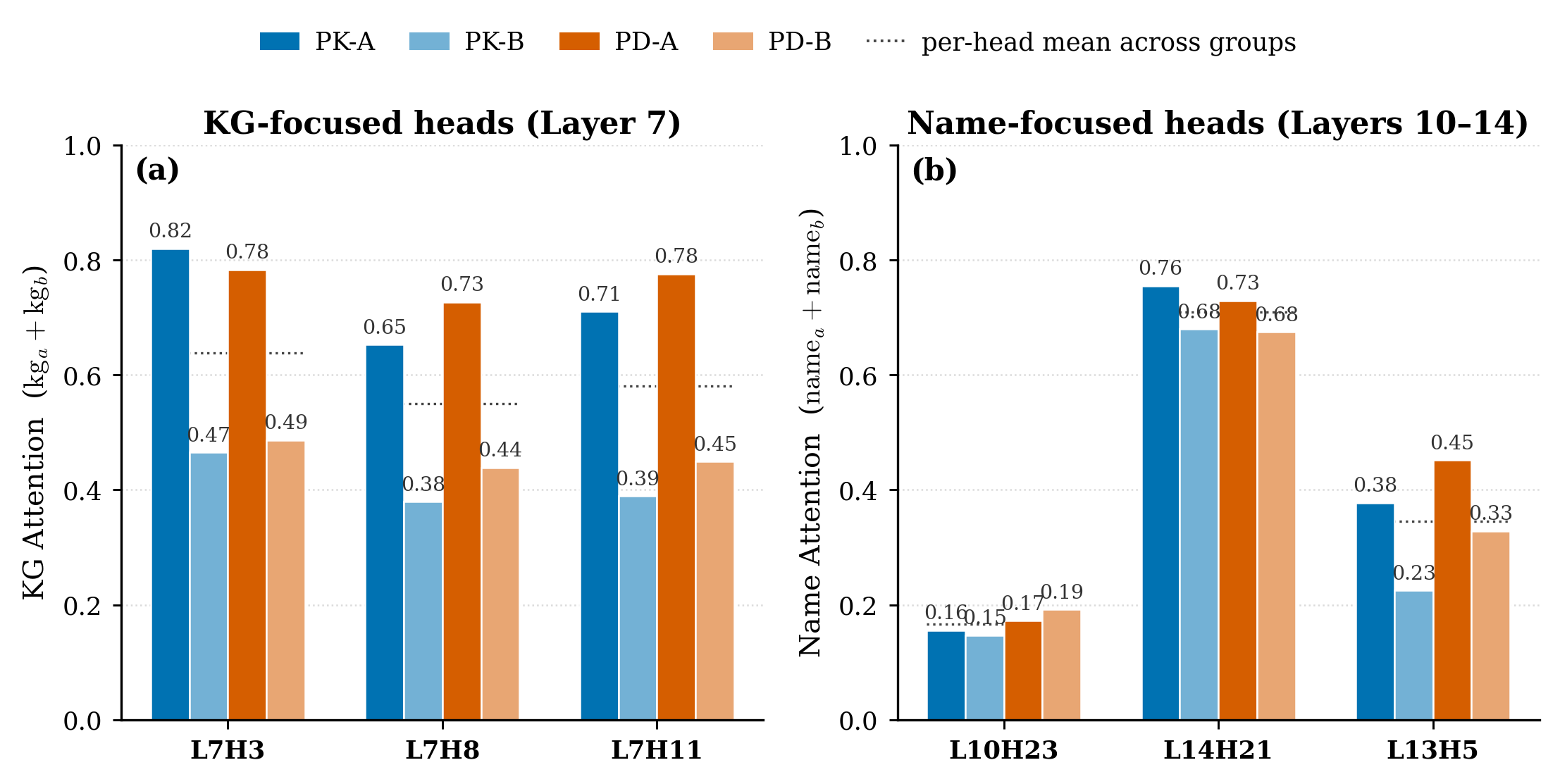}
  \caption{Per-head per-group activations on the six specialised heads identified in \cref{tab:specialised_heads}. Y-axis is the answer-token attention received by each head from each token class, averaged over the mechanism bucket's positive pairs.}
  \label{fig:attn-app-head}
\end{subfigure}
\caption{Fine-tuning redistributes attention from instruction tokens to KG entities and drug names, and concentrates this gain in three KG-attending heads at layer~7 and three name-attending heads at layers~10--15.}
\label{fig:attn_redistribution_app}
\end{figure}

Layer-wise (\cref{fig:attn-app-layer}), KG entity attention rises sharply from layer~7 and peaks at layers~11--13 for PK-A (reaching ${\sim}22\%$ at L12), while drug-name attention rises in the deeper layers (L10--L15) across mechanism buckets. Head-wise (\cref{tab:specialised_heads}), three Layer-7 heads (L7\,H3/H8/H11) allocate ${\sim}51$--$60\%$ of their attention to KG-entity tokens (\textbf{KG heads}), and three deep-layer heads (L14\,H21, L13\,H5, L10\,H23) emerge as \textbf{Name heads}, with L14\,H21 dominantly name-specialised ($70\%$ name attention) and L13\,H5 / L10\,H23 at $17$--$33\%$ (well above the ${\sim}2\%$ base-model baseline). KG heads show a sharp Type~A/B asymmetry, activating much more on Type~A samples (\cref{fig:attn-app-head}, e.g.,\ L7\,H3 KG-attention $0.82$ on PK-A vs.\ $0.49$ on PD-B), while Name heads activate more uniformly across mechanism buckets. This KG-head Type~A selectivity is the architectural mirror of the Type~A entity dominance observed in \cref{sec:modality-results}.

\begin{table}[h]
\centering
\caption{Top specialised attention heads in the fine-tuned Llama-3.2-1B identified by token-class attention fraction.}
\label{tab:specialised_heads}
\small
\begin{tabular}{llccc}
\toprule
\textbf{Function} & \textbf{Head} & \textbf{KG attn} & \textbf{Name attn} & \textbf{Layer} \\
\midrule
KG-focused   & L7\,H3   & \textbf{60.1\%} & 12.5\% & Mid (7) \\
KG-focused   & L7\,H8   & \textbf{50.8\%} & 14.5\% & Mid (7) \\
KG-focused   & L7\,H11  & \textbf{53.5\%} & 17.2\% & Mid (7) \\
\midrule
Name-focused & L14\,H21 & 18.7\% & \textbf{70.4\%} & Deep (14) \\
Name-focused & L13\,H5  & 24.5\% & \textbf{33.0\%} & Deep (13) \\
Name-focused & L10\,H23 & 21.9\% & 17.1\%          & Mid-deep (10) \\
\bottomrule
\end{tabular}
\end{table}

To rule out positional confounds from the causal mask, we compute per-token attention ratios at matched positions (relative position $0.3$--$0.7$). Drug-name tokens receive $4.6\times$ the per-token attention of instruction tokens, while KG-entity tokens receive only $0.6\times$ and SMILES tokens $0.2\times$. The high aggregate KG share is therefore driven by the large number of KG tokens rather than per-token salience, but the ratio still confirms that name and KG tokens are not artifacts of the causal-mask geometry.

\subsubsection{Controlled Attention Intervention: Full Results}
\label{app-E.4.2:attn-intervention}

We selectively suppress attention from the answer token to specific token classes by adding $-10^4$ to the corresponding pre-softmax scores via a 4D attention mask, then re-evaluate the full S2 test set. \cref{tab:intervention_full} reports both absolute AUROC and $\Delta$AUROC under six intervention conditions.

\begin{table}[h]
\centering
\caption{Attention intervention on LLM-FT (S2). The top block reports absolute AUROC, and the bottom block reports $\Delta$AUROC relative to the unintervened baseline. Bold marks the largest effect per group. Suppressing all KG attention collapses PK-A by $-0.219$, while suppressing all Name attention hurts PD-B ($-0.036$) but leaves PK-A unchanged ($+0.000$).}
\label{tab:intervention_full}
\small
\setlength{\tabcolsep}{5pt}
\begin{tabular}{lccccc}
\toprule
\textbf{Condition} & \textbf{ALL} & \textbf{PK-A} & \textbf{PK-B} & \textbf{PD-A} & \textbf{PD-B} \\
\midrule
Baseline               & 0.798 & 0.939 & 0.690 & 0.930 & 0.733 \\
Suppress KG heads (3)  & 0.776 & 0.920 & 0.689 & 0.875 & 0.702 \\
Suppress Name heads (3)& 0.799 & 0.948 & 0.685 & 0.937 & 0.730 \\
Suppress ALL KG        & \textbf{0.673} & \textbf{0.720} & \textbf{0.630} & \textbf{0.739} & 0.652 \\
Suppress ALL Name      & 0.774 & 0.939 & 0.650 & 0.928 & \textbf{0.697} \\
Suppress L7\,H3 only   & 0.786 & 0.945 & 0.690 & 0.898 & 0.704 \\
\midrule
\multicolumn{6}{l}{\textit{$\Delta$AUROC vs.\ baseline}} \\
Suppress KG heads (3)  & $-$0.022 & $-$0.019 & $-$0.001 & $-$0.055 & $-$0.031 \\
Suppress Name heads (3)& $+$0.001 & $+$0.009 & $-$0.005 & $+$0.007 & $-$0.003 \\
Suppress ALL KG        & $-$0.125 & $\boldsymbol{-0.219}$ & $-$0.061 & $-$0.191 & $-$0.081 \\
Suppress ALL Name      & $-$0.024 & $+$0.000 & $-$0.040 & $-$0.002 & $\boldsymbol{-0.036}$ \\
Suppress L7\,H3 only   & $-$0.013 & $+$0.006 & $+$0.000 & $-$0.032 & $-$0.029 \\
\bottomrule
\end{tabular}
\end{table}

\begin{figure}[h]
\centering
\begin{subfigure}{0.9\textwidth}
  \centering
  \includegraphics[width=\linewidth]{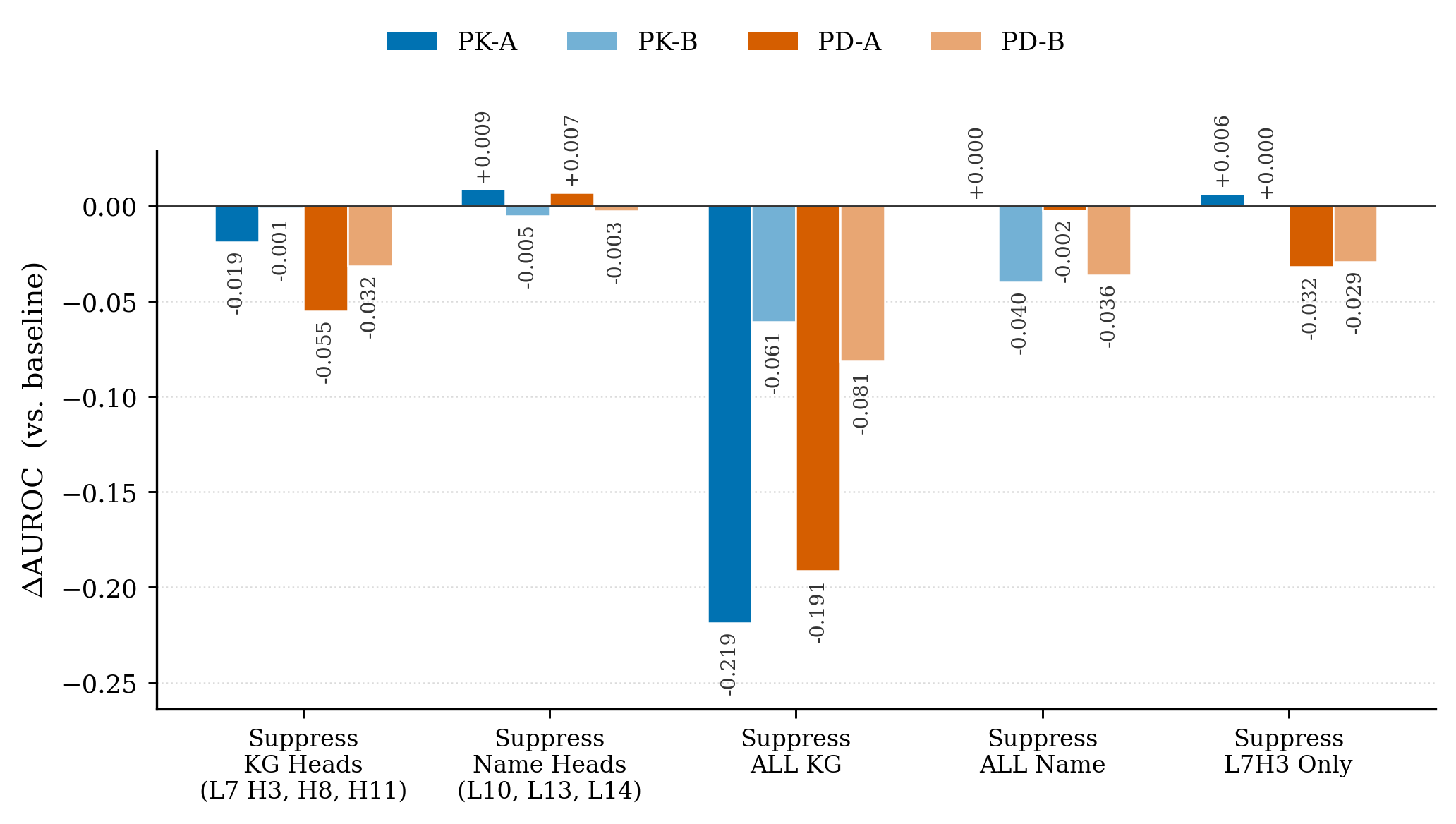}
  \caption{$\Delta$AUROC by mechanism bucket under each intervention condition. Granularity ladder L7\,H3 only $\rightarrow$ 3 KG heads $\rightarrow$ all KG attention shows that the bulk of the PK-A KG dependency is distributed across heads beyond the top-3.}
  \label{fig:intervention-deltaauc}
\end{subfigure}\\[8pt]
\begin{subfigure}{0.9\textwidth}
  \centering
  \includegraphics[width=\linewidth]{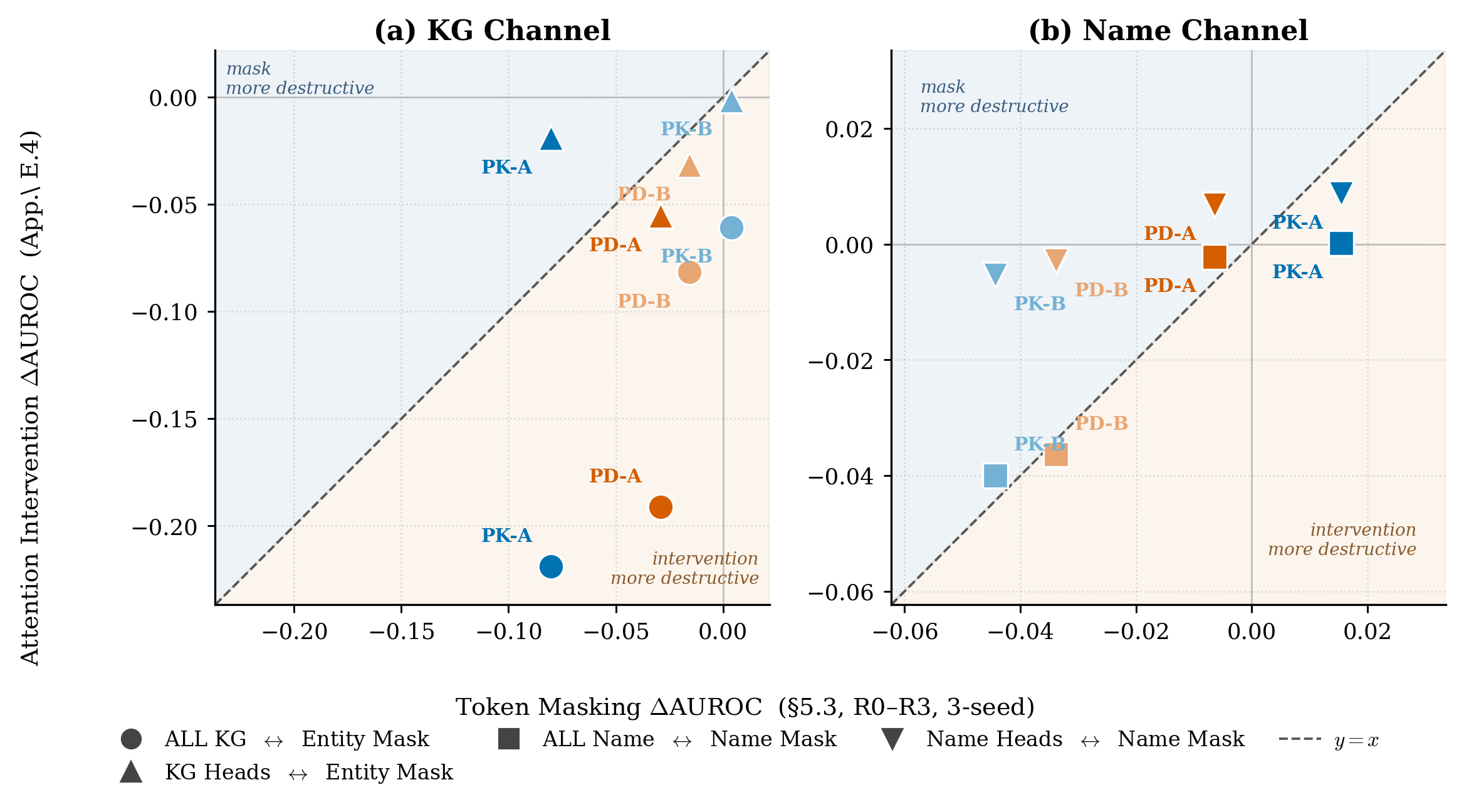}
  \caption{Attention suppression vs.\ input-token masking across all four mechanism buckets. Each marker is one (bucket, perturbation pair). Most KG-related markers sit well below the $y\!=\!x$ diagonal, with the exception of the 3-head suppression on PK-A, indicating attention suppression has a stronger effect than token masking. The Name-related markers (squares) cluster near the origin, indicating both methods agree that name has limited impact on PK-A and PD-A.}
  \label{fig:intervention-vs-masking}
\end{subfigure}
\caption{Controlled attention intervention. (top) $\Delta$AUROC across DDI groups under five intervention conditions plus the unintervened baseline. (bottom) Comparison between attention-level intervention and input-level token masking from \cref{sec:modality-results}.}
\label{fig:intervention_app}
\end{figure}

Beyond the all-KG vs.\ all-Name double dissociation, the table also varies the suppression scope from a single head (L7\,H3) to the three top KG heads to all KG attention. To make this granularity ladder visible across mechanism buckets, and to compare attention-level intervention against input-level token masking from \cref{sec:modality-results}, we plot the per-bucket effects in \cref{fig:intervention_app}.

\paragraph{Granularity.} \cref{fig:intervention-deltaauc} shows that a single head (L7\,H3) produces a negligible drop on PK-A ($+0.006$), and the three top KG heads together produce only a small drop ($-0.019$), accounting for less than $10\%$ of the full all-KG effect ($-0.219$). The single-head and three-head effects are similar in magnitude, indicating functional redundancy within the top KG-head group. The bar height jumps sharply from 3 heads to all KG, far exceeding the gap from one head to three. The same ladder is visible on PD-A and PD-B in the figure, although with smaller absolute magnitudes. The KG-attending function is therefore widely distributed across heads even though it is concentrated at layer~7, and the full all-heads suppression is needed to reproduce the largest PK-A degradation.

\paragraph{Token Masking vs.\ Attention Suppression.} \cref{fig:intervention-vs-masking} compares attention suppression against the input-token masking from \cref{sec:modality-results} across all four mechanism buckets. Each marker is one (bucket, perturbation pair). The two methods probe different surfaces. Token masking rewrites a single key entity or name in the input, whereas attention suppression zeroes out an entire token class across all heads. They therefore agree qualitatively but differ in magnitude. Most KG-related markers sit well below the $y\!=\!x$ diagonal, with the exception of the 3-head suppression on PK-A, and PK-A is the most extreme case (KG attention suppression $-0.219$ vs.\ entity-token masking $-0.080$). Name-related markers (squares) cluster near the origin, indicating both perturbations have limited effect on the name channel for PK-A and PD-A. On PK-A specifically, name-attention suppression leaves the prediction unchanged ($+0.000$) and name-token masking even improves it slightly ($+0.015$), consistent with weak name-channel dependence on PK-A. The two methods therefore agree on the qualitative dissociation that PK-A is KG-driven and name-insensitive, while differing in absolute magnitude in the expected direction.

\subsubsection{Attention Steering: Negative Result}
\label{app-E.4.3:attn-steering}

Motivated by the intervention findings, we test whether explicitly modifying the attention mechanism can improve predictions beyond the LoRA-fine-tuned baseline. We evaluate two categories of methods. \textbf{(A)~Learnable attention modules}, trained with LoRA and base-model parameters frozen, comprise Segment-Aware Attention Bias (SAB, $5{,}632$~params, learnable bias per (layer, head, segment$_q$, segment$_k$)), Per-Head Temperature (PHT, $352$~params), Targeted Head Bias (THB, $96$~params, SAB applied only to the six functional heads of \cref{tab:specialised_heads}), and Instruction Suppression Bias (ISB, $11$~params, one scalar per layer subtracted from attention to instruction-segment keys). \textbf{(B)~Zero-parameter inference-time methods} comprise Instruction Attention Decay (IAD, multiplying instruction-key attention by $\gamma{<}1$), Contrastive Decoding (logits$_\text{FT}-\alpha\cdot$logits$_\text{Base}$), and DoLa (early- vs.\ final-layer logit contrast).

\begin{table}[h]
\centering
\caption{Attention-steering results on S2 AUROC, single random seed on the 800-drug subset. $\Delta$ALL is reported in $\times 10^{2}$ relative to the LoRA-FT baseline ($0.798$, single seed). No method achieves a meaningful improvement.}
\label{tab:steering_full}
\small
\setlength{\tabcolsep}{4pt}
\begin{tabular}{llccccccr}
\toprule
\textbf{Cat.} & \textbf{Method} & \textbf{Best lr} & \textbf{ALL} & \textbf{PK-A} & \textbf{PK-B} & \textbf{PD-A} & \textbf{PD-B} & $\boldsymbol{\Delta}$\textbf{ALL} \\
\midrule
--- & Baseline (LoRA-FT) & --- & 0.798 & 0.939 & 0.690 & 0.930 & 0.733 & --- \\
\midrule
A & SAB & 5e-4 & 0.799 & 0.934 & 0.688 & 0.935 & 0.739 & $+$0.1 \\
A & PHT & 1e-3 & 0.797 & 0.927 & 0.686 & 0.931 & 0.739 & $-$0.2 \\
A & THB & 1e-3 & 0.799 & 0.939 & 0.690 & 0.930 & 0.734 & $+$0.0 \\
A & ISB & 1e-2 & 0.799 & 0.937 & 0.687 & 0.935 & 0.737 & $+$0.1 \\
\midrule
B & IAD ($\gamma{=}0.8$) & --- & 0.796 & 0.939 & 0.688 & 0.925 & 0.730 & $-$0.2 \\
B & CD ($\alpha{=}0.1$)  & --- & 0.798 & 0.939 & 0.690 & 0.931 & 0.733 & $-$0.0 \\
B & DoLa (early=2)        & --- & 0.793 & 0.934 & 0.678 & 0.927 & 0.730 & $-$0.5 \\
\bottomrule
\end{tabular}
\end{table}

The negative result is itself informative. Across both learnable and zero-parameter approaches, no method exceeds noise, indicating that LoRA fine-tuning has already optimised the attention distribution close to its capacity limit. The bottleneck for further improvement lies not in attention allocation but in knowledge capacity. Type~B groups lack shared entities (PK-B) or sufficient pretrained pharmacological coverage (PD-B), and attention redistribution alone cannot overcome these limitations.

\subsubsection{Stability Across Seeds and Architectures}
\label{app-E.4.4:attn-stability}

We repeat the six intervention conditions for Llama-3.2-1B and Qwen-2.5-0.5B over seeds 42, 43, and 44, using the complete S2 test set of each 800-drug subset.
Within each architecture, KG- and name-focused heads are identified once from seed~42 attention and held fixed across the three runs.
This prevents head reselection from contributing to cross-seed variation.
For each condition and group, $\Delta$AUROC is computed relative to the baseline within the same seed before taking the mean and sample standard deviation across seeds.

The fixed KG heads in Llama-3.2-1B are L7\,H3, L7\,H8, and L7\,H11, with L10\,H23, L14\,H21, and L13\,H5 as the name heads.
For Qwen-2.5-0.5B (24~layers $\times$ 14~heads), the corresponding sets are L12\,H2, L14\,H8, and L12\,H4 for KG attention, and L17\,H0, L16\,H11, and L12\,H9 for name attention.
Single-head suppression targets L7\,H3 in Llama and L12\,H2 in Qwen, and the all-KG and all-name conditions apply to every layer and head.

\begin{table}[t]
\centering
\caption{Attention-intervention stability on S2 of the 800-drug subset (3-seed mean$\pm$std). For each model, the upper block reports absolute AUROC and the lower block reports within-seed $\Delta$AUROC relative to the unintervened baseline.}
\label{tab:attn_stability}
\small
\setlength{\tabcolsep}{4pt}
\renewcommand{\arraystretch}{1.1}
\begin{tabular}{@{}lccccc@{}}
\toprule
\textbf{Condition}
& \textbf{ALL}
& \textbf{PK-A}
& \textbf{PK-B}
& \textbf{PD-A}
& \textbf{PD-B} \\
\midrule
\multicolumn{6}{l}{\textbf{Llama-3.2-1B: AUROC}} \\
Baseline
& $0.776_{\pm0.020}$ & $0.878_{\pm0.068}$
& $0.602_{\pm0.078}$ & $0.949_{\pm0.017}$
& $0.759_{\pm0.043}$ \\
Suppress KG heads (3)
& $0.754_{\pm0.019}$ & $0.857_{\pm0.066}$
& $0.609_{\pm0.069}$ & $0.880_{\pm0.051}$
& $0.726_{\pm0.044}$ \\
Suppress Name heads (3)
& $0.778_{\pm0.018}$ & $0.889_{\pm0.063}$
& $0.596_{\pm0.079}$ & $0.957_{\pm0.017}$
& $0.758_{\pm0.041}$ \\
Suppress ALL KG
& $0.661_{\pm0.011}$ & $0.670_{\pm0.049}$
& $0.586_{\pm0.039}$ & $0.758_{\pm0.018}$
& $0.685_{\pm0.036}$ \\
Suppress ALL Name
& $0.759_{\pm0.015}$ & $0.894_{\pm0.054}$
& $0.556_{\pm0.082}$ & $0.950_{\pm0.019}$
& $0.725_{\pm0.033}$ \\
Suppress L7\,H3 only
& $0.764_{\pm0.020}$ & $0.885_{\pm0.063}$
& $0.606_{\pm0.075}$ & $0.905_{\pm0.036}$
& $0.724_{\pm0.044}$ \\
\midrule
\multicolumn{6}{l}{\textit{$\Delta$AUROC vs.\ baseline}} \\
Suppress KG heads (3)
& $-0.022_{\pm0.003}$ & $-0.021_{\pm0.005}$
& $+0.007_{\pm0.009}$ & $-0.069_{\pm0.048}$
& $-0.033_{\pm0.002}$ \\
Suppress Name heads (3)
& $+0.002_{\pm0.002}$ & $+0.011_{\pm0.006}$
& $-0.007_{\pm0.003}$ & $+0.008_{\pm0.005}$
& $-0.001_{\pm0.003}$ \\
Suppress ALL KG
& $-0.115_{\pm0.010}$ & $-0.208_{\pm0.022}$
& $-0.017_{\pm0.039}$ & $-0.191_{\pm0.003}$
& $-0.074_{\pm0.017}$ \\
Suppress ALL Name
& $-0.017_{\pm0.008}$ & $+0.016_{\pm0.015}$
& $-0.047_{\pm0.006}$ & $+0.001_{\pm0.003}$
& $-0.034_{\pm0.014}$ \\
Suppress L7\,H3 only
& $-0.012_{\pm0.004}$ & $+0.007_{\pm0.008}$
& $+0.003_{\pm0.003}$ & $-0.044_{\pm0.031}$
& $-0.035_{\pm0.006}$ \\
\midrule
\multicolumn{6}{l}{\textbf{Qwen-2.5-0.5B: AUROC}} \\
Baseline
& $0.763_{\pm0.018}$ & $0.859_{\pm0.067}$
& $0.624_{\pm0.023}$ & $0.944_{\pm0.038}$
& $0.738_{\pm0.037}$ \\
Suppress KG heads (3)
& $0.733_{\pm0.025}$ & $0.807_{\pm0.075}$
& $0.625_{\pm0.012}$ & $0.890_{\pm0.078}$
& $0.713_{\pm0.054}$ \\
Suppress Name heads (3)
& $0.743_{\pm0.023}$ & $0.887_{\pm0.053}$
& $0.547_{\pm0.086}$ & $0.938_{\pm0.046}$
& $0.698_{\pm0.058}$ \\
Suppress ALL KG
& $0.635_{\pm0.017}$ & $0.649_{\pm0.051}$
& $0.586_{\pm0.009}$ & $0.722_{\pm0.051}$
& $0.643_{\pm0.011}$ \\
Suppress ALL Name
& $0.744_{\pm0.023}$ & $0.890_{\pm0.052}$
& $0.553_{\pm0.084}$ & $0.938_{\pm0.047}$
& $0.697_{\pm0.065}$ \\
Suppress L12\,H2 only
& $0.750_{\pm0.017}$ & $0.846_{\pm0.064}$
& $0.625_{\pm0.015}$ & $0.909_{\pm0.071}$
& $0.720_{\pm0.049}$ \\
\midrule
\multicolumn{6}{l}{\textit{$\Delta$AUROC vs.\ baseline}} \\
Suppress KG heads (3)
& $-0.031_{\pm0.014}$ & $-0.052_{\pm0.012}$
& $+0.001_{\pm0.022}$ & $-0.054_{\pm0.044}$
& $-0.024_{\pm0.018}$ \\
Suppress Name heads (3)
& $-0.021_{\pm0.009}$ & $+0.028_{\pm0.028}$
& $-0.077_{\pm0.064}$ & $-0.006_{\pm0.012}$
& $-0.039_{\pm0.023}$ \\
Suppress ALL KG
& $-0.128_{\pm0.006}$ & $-0.210_{\pm0.016}$
& $-0.038_{\pm0.032}$ & $-0.222_{\pm0.050}$
& $-0.095_{\pm0.029}$ \\
Suppress ALL Name
& $-0.019_{\pm0.011}$ & $+0.030_{\pm0.032}$
& $-0.071_{\pm0.063}$ & $-0.006_{\pm0.010}$
& $-0.041_{\pm0.030}$ \\
Suppress L12\,H2 only
& $-0.013_{\pm0.008}$ & $-0.014_{\pm0.006}$
& $+0.002_{\pm0.010}$ & $-0.035_{\pm0.034}$
& $-0.018_{\pm0.013}$ \\
\bottomrule
\end{tabular}
\end{table}

For Llama-3.2-1B, cross-seed variation in overall $\Delta$AUROC is small relative to the effect of suppressing all KG attention (\cref{tab:attn_stability}).
Across architectures, the 95\% $t$-intervals for the mean PK-A change under this intervention overlap and exclude zero, both for Llama and Qwen.
The PK-A decrease also exceeds the PD-B decrease in all six model--seed combinations.
Together, these results show that stronger PK-A sensitivity to KG-attention suppression persists across the tested seeds and architectures.

\subsection{Embedding Geometry Evidence for Mechanism and KG Utilization}
\label{app-E.5:embedding-geometry}

This appendix provides direct embedding-level evidence supporting two design choices of \textsc{ColdDDI}: (i) the PK/PD mechanism distinction is geometrically encoded in model representations, and (ii) KG access does not imply KG utilization, as demonstrated by opposite KG-injection effects across methods with similar input access.

\paragraph{Setup.}
We compare TIGER (mol+KG hybrid baseline) and LLM-FT (LoRA-tuned Llama-3.2-1B under P4) on the S2 split's $1{,}919$ positive pairs. Each method has two embedding conditions, namely \textit{Without KG} (KG channel zeroed for TIGER, entity tokens masked for LLM-FT) and \textit{With KG} (full forward pass). 

\paragraph{Projection.}
For each method, the pooled (no-KG, with-KG) embeddings are projected onto two axes. The horizontal axis is a Fisher LDA direction between pooled positive and negative centroids. The vertical axis is the first principal component of the residual after the discriminant component is removed, orthogonal to the LDA direction by construction. Both axes are $z$-scored on the pooled distribution, so the two rows of a column share a common visual frame. For visual clarity, each panel shows a randomly sampled subset of the $1{,}919$ positives, with the same subset reused across all four panels for direct point-by-point comparison.

\paragraph{Decision Boundary.}
The dashed line in each panel is the 2D linear approximation of the model's high-dimensional decision boundary. For each column we fit a logistic regression on the pooled $(\text{no-KG}, \text{with-KG})$ predictions of that method, with the model's predicted label (not the ground-truth label) as the target. The same line is therefore shared between the two rows of a column, so points crossing the line between rows correspond directly to a flip in the model's prediction. Residual disagreements (full-opacity dots on the negative side and half-opacity dots on the positive side) reflect head structure that this 2D plane cannot capture.

\begin{figure}[t]
  \centering
  \includegraphics[width=0.85\linewidth]{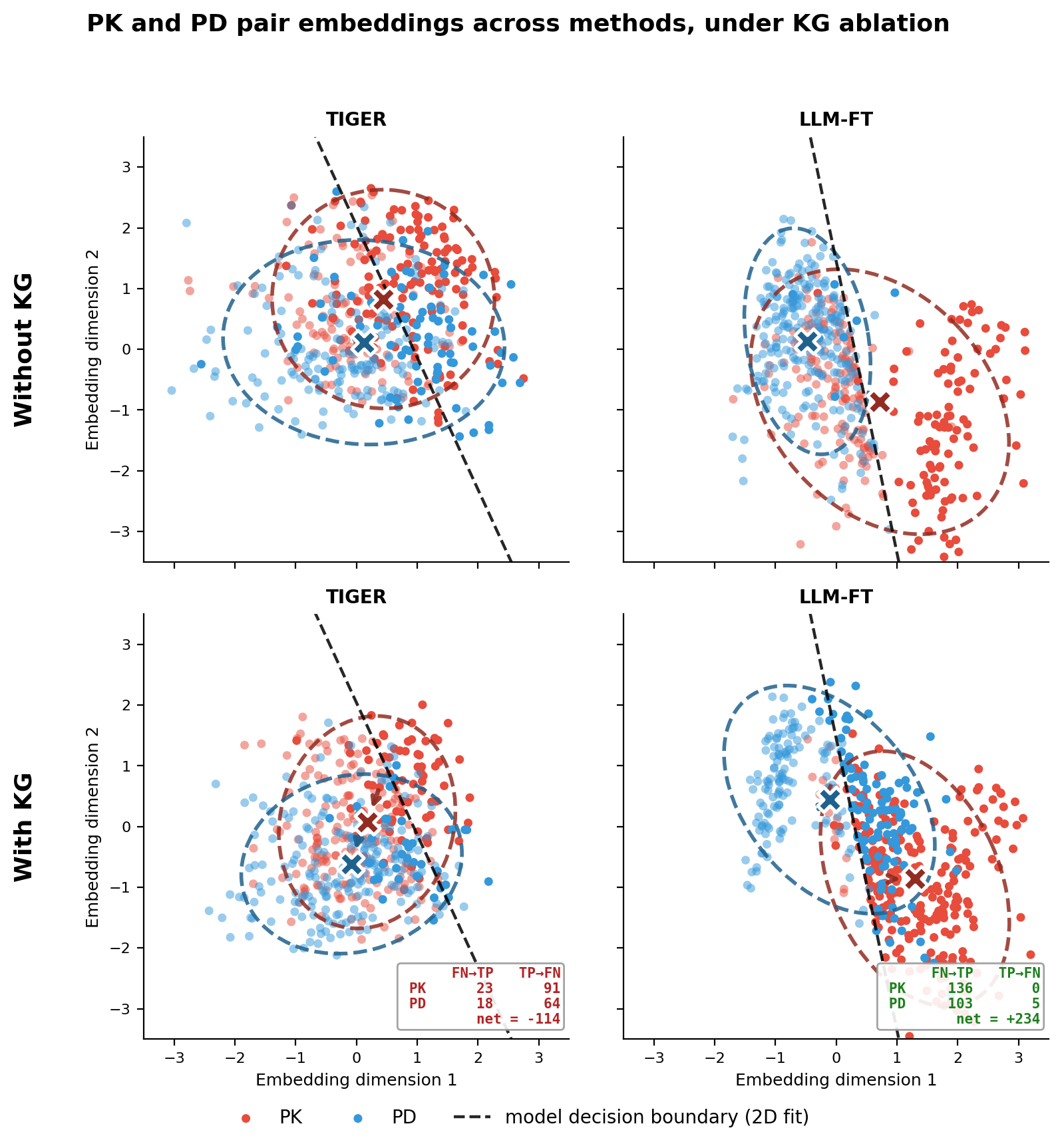}
  \caption{S2 pair embeddings projected onto each method's Fisher LDA axis and residual PC1. Rows. Without/with one-hop KG. Columns. TIGER, LLM-FT. Red/blue dots. PK/PD positives, full opacity if predicted correctly and half opacity otherwise. Dashed line. 2D logistic regression on the model's predicted label, fitted on pooled (no-KG, with-KG) pairs and shared between the two rows of a column. Dashed ellipses. $2\sigma$ Gaussian confidence regions per class on the visible subset. $\times$ markers. Per-class centroids on the visible subset. The KG-induced centroid shift can be read by comparing the $\times$ markers between the two rows of a column. Bottom-right inset (bottom row). FN$\to$TP and TP$\to$FN counts split by PK/PD plus signed net total. \textbf{Net flip}. $-114$ (TIGER, KG hurts), $+234$ (LLM-FT, KG helps).}
  \label{fig-sec1:fn-flip}
\end{figure}

\paragraph{Two Narrative Claims Supported by \cref{fig-sec1:fn-flip}.}

\textit{(1) PK/PD stratification is geometrically real.} The PK and PD centroids ($\times$ markers) are visibly separated within each method's representation, with LLM-FT showing a clearer PK/PD split than TIGER on the with-KG row. PK/PD is therefore not just a label distinction but a structural property that stronger representations capture more clearly.

\textit{(2) KG access $\neq$ KG utilization.} Comparing the $\times$ centroid markers between the two rows of each column reveals the KG-induced centroid shift. For TIGER, KG injection drives the centroids toward the model-negative side of the decision boundary, with a net flip of $-114$ (TP$\to$FN exceeding FN$\to$TP). For LLM-FT, the same KG injection drives the centroids toward the model-positive side, with a net flip of $+234$. Two methods with the same KG inputs therefore route the KG signal in opposite directions, providing solid visual evidence that KG access does not imply KG utilization (also evidence for cross-method KPS-F gap, reported in \cref{tab:kps-summary}).

\end{document}